%% file: ejs_arxiv.tex
\documentclass{article}
\RequirePackage[OT1]{fontenc}
\RequirePackage{amsthm,amsmath}
\RequirePackage[numbers]{natbib}
\RequirePackage[colorlinks,citecolor=blue,urlcolor=blue]{hyperref}
\usepackage{amssymb}
\usepackage{xspace}
\usepackage[top=1in, bottom=1in, left=1.3in, right=1.3in]{geometry}
\usepackage{alltt}
\usepackage{lmodern}
\usepackage{graphicx} % more modern
\newtheorem{thm}{Theorem}[section]
\date{}
\input{header}
\begin{document}
\def\argmin{\operatornamewithlimits{arg\,min}}

%\begin{frontmatter}
\title{Nonparametric Link Prediction in Large Scale Dynamic Networks}
%\runtitle{Nonparametric Link Prediction}
%\thankstext{T1}{Footnote to the title with the `thankstext' command.}

%\begin{aug}
\author{ Purnamrita Sarkar\\
\texttt{psarkar@eecs.berkeley.edu}\\
U. C. Berkeley
\\
\\
Deepayan Chakrabarti\footnote{This work was partly done when the author was at Yahoo! Research}\\
\texttt{deepay@fb.com}\\
Facebook
\\
\\
Michael I. Jordan\\
\texttt{jordan@cs.berkeley.edu}\\
University of California, Berkeley
}
\maketitle

\begin{abstract}
We propose a nonparametric approach to link prediction in large-scale dynamic
networks.  Our model uses graph-based features of pairs of nodes as well as
those of their \emph{local neighborhoods} to predict whether those nodes
will be linked at each time step.  The model allows for different types of
evolution in different parts of the graph (e.g, growing or shrinking communities).
We focus on large-scale graphs and present an implementation of our model that
makes use of locality-sensitive hashing to allow it to be scaled to
large problems.  Experiments with simulated data as well as five real-world
dynamic graphs show that we outperform the state of the art, especially when
sharp fluctuations or nonlinearities are present.  We also establish
theoretical properties of our estimator, in particular consistency and
weak convergence, the latter making use of an elaboration of Stein's method
for dependency graphs.
\end{abstract}

%
%\begin{keyword}[class=AMS]
%\kwd[Primary ]{62G08}
%\kwd[; secondary ]{91D30}
%\end{keyword}

%\begin{keyword}
%\kwd{link prediction}
%\kwd{dynamic networks}
%\kwd{nonparametric}
%\end{keyword}
%\tableofcontents
%\end{frontmatter}

\input{intro}
\input{proposed}

%%
\input{hashing}

\input{exp}
\input{consistency}

\input{stein-general-v3}

\input{dist-conv}
%%
\input{related}
%%\vspace{-1em}
\input{conc}

\input{appendix}
\section*{Acknowledgements}
We are grateful to Peter Bickel for helpful discussions on this topic.
\bibliographystyle{plainnat}
\bibliography{purna}
\end{document}

%% file: header.tex
\newcommand{\denselist}{\itemsep 0pt\partopsep 0pt}
\newcommand{\reminder}[1]{{}}
\providecommand{\mypara}[1]{\par \noindent \textsc{#1.}}

\newcommand{\ind}{\ensuremath\mathbf{1}\xspace}
\newcommand{\var}{\mathrm{var}}
\newcommand{\g}{\dcf} % As in Masry
\newcommand{\f}{f} %Masry f
\newcommand{\h}{h} %Masry h

\newcommand{\nn}{\ensuremath{n}\xspace}
\newcommand{\convprob}{\ensuremath{\stackrel{P}{\longrightarrow}}}

\newcommand{\dist}{\mathrm{dist}\xspace}

\newcommand{\dcYBASE}[3]{\ensuremath{Y_{#1}(#2,#3)}\xspace}
\newcommand{\dcY}{\dcYBASE{t+1}{i}{j}}
\newcommand{\dcYp}{\dcYBASE{t'+1}{i'}{j'}}

\newcommand{\dcf}{\ensuremath{g}}
\newcommand{\dchatf}{\ensuremath{\est{\g}}\xspace}

\newcommand{\dcD}{\ensuremath{D}\xspace}

\newcommand{\dcK}{\ensuremath{\Gamma}\xspace}
\newcommand{\dcKone}{\ensuremath{K}\xspace}
\newcommand{\dcS}{\ensuremath{S}\xspace}
\newcommand{\dcfeat}{\ensuremath{\psi\sub{\nt}(i,j)}\xspace}
\newcommand{\dcfeatsmall}{\ensuremath{\psi_t(i,j)}\xspace}
\newcommand{\dcfeatp}{\ensuremath{\psi_{t'}(i',j')}\xspace}

\newcommand{\dcgBASE}[3]{\ensuremath{\mbox{s}_{#1}\left(#2,#3\right)}\xspace}
\newcommand{\dcg}{\dcgBASE{t}{i}{j}}
\newcommand{\dcgp}{\dcgBASE{t'}{i'}{j'}}
\newcommand{\dcgq}{\dcgBASE{T}{q}{q'}}

\newcommand{\dchBASE}[2]{\ensuremath{d_{#1}\left(#2\right)}\xspace}
\newcommand{\dch}{\dchBASE{t}{i}}
\newcommand{\dchp}{\dchBASE{t'}{i'}}
\newcommand{\dchq}{Q}
\newcommand{\dcN}[1]{\ensuremath{N_{#1}(i)}\xspace}

\newcommand{\dcnBASE}[3]{\ensuremath{\eta_{#1,#2}\left(#3\right)}\xspace}
\newcommand{\dcnplusBASE}[3]{\ensuremath{\eta_{#1,#2}^+\left(#3\right)}\xspace}
\newcommand{\dcnJUSTFUNC}{\ensuremath{\eta_{.}(s)}\xspace}

\newcommand{\dcnplusJUSTFUNC}{\ensuremath{\eta^+_{.}(s)}\xspace}
\newcommand{\dcn}{\dcnBASE{i}{t}{s}}
\newcommand{\dcnplus}{\dcnplusBASE{i}{t}{s}}
\newcommand{\dcno}{\dcnBASE{i}{t+1}{s}}
\newcommand{\dcnpluso}{\dcnplusBASE{i}{t+1}{s}}
\newcommand{\dcnp}{\dcnBASE{i'}{t'}{s}}
\newcommand{\dcnplusp}{\dcnplusBASE{i'}{t'}{s}}

\newcommand{\vecN}[1]{\ensuremath{\vec{N}_{#1}}\xspace}
\newcommand{\cov}{\mbox{cov}}

\newcommand{\expll}{\textsf{LL}\xspace}
\newcommand{\expcn}{\textsf{CN}\xspace}
\newcommand{\expaa}{\textsf{AA}\xspace}
\newcommand{\expkz}{\textsf{Katz}\xspace}
\newcommand{\expcnall}{\textsf{CN-all}\xspace}
\newcommand{\expaaall}{\textsf{AA-all}\xspace}
\newcommand{\expkzall}{\textsf{Katz-all}\xspace}

\newcommand{\expnp}{\textsf{NNI}\xspace}
\newcommand{\seasonal}{\textsf{Seasonal}\xspace}
\newcommand{\stationary}{\textsf{Stationary}\xspace}

\newtheorem{lem}[thm]{Lemma}

\newtheorem{proposition}[thm]{Proposition}
\theoremstyle{remark}
\newtheorem{remark}[thm]{Remark}
\newcommand{\et}{\ensuremath{\mathcal{E}_{T_1}}\xspace}
\newcommand{\ec}{\xspace\ensuremath{S_C}\xspace}
\newcommand{\mc}{\xspace\ensuremath{M}\xspace}

\newenvironment{myquote}{\list{}{\leftmargin=5em\rightmargin=0em}\item[]}{\endlist}
\newcommand{\txtred}[1]{\textcolor{red}{{}}}
\newcommand{\abs}[1]{\ensuremath{\left|#1\right|}\xspace}

\newcommand{\nt}{\ensuremath{T}\xspace}

\newcommand{\nfatm}{\ensuremath{F_{m}}\xspace}
\newcommand{\nfltk}{\ensuremath{F_{\leq k}}\xspace}
\newcommand{\nfgtk}{\ensuremath{F_{> k}}\xspace}
\newcommand{\nltk}{\ensuremath{N_{\leq k}}\xspace}

\newcommand{\natm}{\ensuremath{N_{m}}\xspace}
\newcommand{\ngtk}{\ensuremath{N_{> k}}\xspace}

\newcommand{\sfatm}{\ensuremath{|F_{m}|}\xspace}
\newcommand{\sfltk}{\ensuremath{|F_{\leq k}|}\xspace}
\newcommand{\sltk}{\ensuremath{|N_{\leq k}|}\xspace}

\newcommand{\satm}{\ensuremath{|N_{m}|}\xspace}
\newcommand{\sgtk}{\ensuremath{|N_{> k}|}\xspace}

\newcommand{\cone}{\ensuremath{c_1}}
\newcommand{\ctwo}{\ensuremath{c_2}}
\newcommand{\cthree}{\ensuremath{c_3}}
\newcommand{\cfour}{\ensuremath{c_4}}
\newcommand{\cfive}{\ensuremath{c_5}}

\newtheorem{theorem}{Theorem}
\newtheorem{assumption}[theorem]{Assumption}
\newcommand{\sub}[1]{\text{{\fontsize{10}{.1}\selectfont $_{#1}$}}}
\newcommand{\cn}{\ensuremath{\gamma\sub{\nt}}\xspace}
\newcommand{\bt}{\ensuremath{b\sub{\nt}}\xspace}
\newcommand{\bti}{\ensuremath{\zeta\sub{\nt}}\xspace} %Bandwidth for the inner kernel
\newcommand{\Kb}{K_{b_{\scriptscriptstyle{\nt}}}} %Kernel
\newcommand{\Bt}{\ensuremath{B_{\nt}}\xspace}
\newcommand{\est}[1]{\widehat{#1}\sub{\nt}} %estimate of \g,\h,\f

\newcommand{\qit}{\ensuremath{q_{it}}\xspace}

\newcommand{\sjtr}{\ensuremath{u_{tr}}\xspace}
\newcommand{\sjtR}{\ensuremath{u_{tR_t}}\xspace}

%% file: intro.tex
\vspace{-.1in}
\section{Introduction}
%\vspace{-.1in}
Many real-world problem domains generate data in the form of graphs or
networks.  Examples include social networks (e.g., Facebook), recommendation
services (e.g., Netflix or Last.fm), biochemical networks, citation graphs
and market analysis.  The inferential problem in these settings is often
one of \emph{link prediction}.  This problem can be formulated in a
static setting where one assumes that a fixed but unknown graph is
partially observed, and one wishes to assess whether a pair of nodes that
are not known to be linked are in fact linked, given an observed linkage
pattern among other nodes.  Many real-world graphs are often best modeled,
however, as dynamic entities, where links can arise and disappear over time.
In the dynamic setting the link prediction problem involves assessing
whether two nodes will be linked at time $t$ given the linkage patterns
at all previous times.

Real-world graphs of current interest are often very large, involving
many hundreds of thousands or millions of nodes.  The dynamic setting
involves sequences of such graphs.  Given the large-scale nature of these
data structures, inferential methodology that may be feasible on smaller
graphs of hundreds of nodes, such as Markov random fields and other
graphical models, are generally infeasible for real-world link
prediction problems, and practical approaches to such problems generally
involve simple heuristics, such as estimating a probability of a link
being present as a simple function of the last time a pair of nodes
formed a link, or the number of common neighbors between a pair of
nodes~\citep{arimaHuang,kleinberg04,SarkarChenDubrawski2008,TylendaTimeaware}.
While these heuristics do respect the computational imperative, and
are often useful in practice, there has been little in the way of
statistical analysis to provide a sound foundation for their use
and to assess the quality of the inferences that they provide.
This is particularly true in the dynamic setting, where link
prediction is often approached by specifying various measures of
connectivity in a static graph and extending these measures in an
ad hoc manner to sequences of graphs.

In this paper, we develop a nonparametric methodology for link prediction
in large-scale dynamic networks.  Our methodology is a relatively simple
kernel-based approach, one that aims to retain the virtues of the simple
heuristic methods, both in their favorable computational scaling and in
the relatively weak assumptions that they appear to make on the graph
generation process.  As compared to existing heuristic approaches,
however, our kernel-based approach allows us to provide a formal inferential
treatment of link prediction---we establish consistency and weak convergence
of our estimator.  On the computational front, while a naive implementation
of a kernel method would have poor scaling (due to the need to compare
query points to every point in a training set), we show that our kernel-based
approach is amenable to locality sensitive hashing (LSH)~\citep{Motwani},
which provides a fast and scalable implementation of the estimator.

Our approach is in the spirit of the nonparametric autoregressive time
series models~\citep{Masry_nonparametric}.  In these models the evolution
of a sequence $x_t$ of continuous univariate random variables is modeled
by taking the conditional expectation of $x_t$ to be a function of a moving
window $(x_{t-1}, \ldots, x_{t-p})$, and estimating this function via kernel
regression.  It is also possible to consider multivariate extensions of such
models.  While it would be possible in principle to apply such models to
our problem by encoding graphs as vectors, in practice the large-scale
graphs that are our focus would generate high-dimensional vector
representations that would be fatal to naive kernel regression.
Instead, we think of the graphs as providing a ``spatial'' dimension
that is orthogonal to the time axis.  In addition to imposing the
conditional independence assumption implicit in the use of a moving
window, we make the additional assumption that the linkage behavior of
any node $i$ is independent of the rest of the graph given its ``local
neighborhood''; in effect, local neighborhoods are to the spatial
dimension what moving windows are to the time dimension.

Thus we model the out-edges of $i$ at time $t$ as a function of the local
neighborhood of $i$ over a moving window of time, resulting in a much more
tractable problem. As a byproduct, this also allows for different evolutions
for different regions to exist in the same graph; e.g., regions of slow versus fast
change in links, assortative versus disassortative regions (where high-degree
nodes are more/less likely to connect to other high-degree nodes), densifying
versus sparsifying regions, and so on.

As a brief summary, our contributions are as follows:\\
(1) {\em Nonparametric problem formulation:} We offer, to our
knowledge, the first nonparametric model for link prediction in dynamic networks.
The model is powerful enough to accommodate different regions with different dynamics,
which is not accommodated in existing heuristic approaches. It also allows covariates
to be incorporated (such as demographic data about a node).

(2) {\em Consistency and weak convergence of the estimator:} We prove consistency
of our estimator using notions of strong mixing in Markov chains.  To establish
weak convergence we show how to adapt Stein's method to our setting, going beyond
the dependency graph formulation of Stein's method~\citep{Rinott:1996:MCL:230170.230181}
to allow long-range weak dependence instead of marginal independence.

(3) {\em Fast implementation via LSH:} Nonparametric methods such as kernel
regression require computing kernel similarities between a query and all members
of the training set. A naive implementation would lead to computation linear in
the training set size, which is generally infeasible for large-scale networks.
In order to mitigate this issue, we adapt the locality sensitive hashing algorithm
of~\citet{Motwani} to our particular kernel function.

(4) {\em Empirical improvements over previous methods:}
We demonstrate the empirical effectiveness of our method on link prediction tasks
on both simulated and real networks.  On graphs with nonlinear linkage patterns
(e.g., seasonal trends), we outperform all of the state-of-the-art heuristic
measures for static and dynamic graphs.  This result is obtained in particular
on a real-world sensor network graph.  On other real-world datasets with
smoother and simpler evolution, we perform as well as the best competitor.
Finally, we compare our LSH-based kernel regression to exact kernel regression,
and show that the LSH-based approach yields almost identical accuracy at a
fraction of the computational cost.

The rest of the paper is organized as follows. We present the model and
the estimator in Section~\ref{sec:proposed}.  Our LSH implementation is
described in Section~\ref{sec:lsh}.  Section~\ref{sec:exp} provides
an experimental evaluation of our method.  We provide an analysis of
consistency in~\ref{sec:consistency}.  In Section~\ref{sec:stein-general-v3}
we discuss our adaptation of Stein's method which we use to establish
weak convergence of our estimator in Section~\ref{sec:dist-conv}.
We provide a discussion of related work in Section~\ref{sec:related}
and we present our conclusions in Section~\ref{sec:conc}.

\reminder{We can also predict link deletions; PUT in conclusions}

%% file: proposed.tex
\section{The Model and the Estimator}
\label{sec:proposed}
%\vspace{-.1in}

We begin by introducing some notation. Consider a sequence of
directed graphs, $\mathcal{G} = \{G_1, G_2, \ldots, G_{t}\}$.
Define the indicator $\dcYBASE{t}{i}{j}$ which equals $1$ if
the edge $i\rightarrow j$ exists at time $t$, and $0$ otherwise.
Let \dcN{t} denote the \emph{local neighborhood} of node $i$ in $G_t$;
in our experiments, we define it to be the set of nodes within
two hops of $i$ and all edges between the nodes in that set.
Note that the neighborhoods of nearby nodes can overlap.
Let $\vecN{t,p}(i) = \{\dcN{t},\ldots, \dcN{t-p+1}\}$;
this represents the local neighborhood of $i$ along both spatial
and temporal dimensions.
%\textcolor{red}{Deepay: we should add how common neighbor can be made a function of the neighborhood.}

\subsection{The Model}
Our model is as follows:
%\small{
%\vspace{-.5em}
\begin{align*}
\dcY | \mathcal{G} & \sim \mbox{Bernoulli}(\dcf(\dcfeatsmall)) \\
\dcfeatsmall & = \{\dcg, \dch\},
\end{align*}
%\vspace{-1em}
%}
where $0\leq \dcf(\cdot) \leq 1$ is a function of
two sets of features: those specific to the {\em pair} of nodes
$(i,j)$ under consideration---$\{\dcg\}$---and those for the local
neighborhood of the endpoint $i$---$\{\dch\}$.  We require that both
of these feature sets be functions of $\vecN{t,p}(i)$.
%specific to the pair $(i,j)$, and \dch for the neighborhood of $i$; both \dcg
%and \dch are functions of $\vecN{t,p}(i)$.
Thus, \dcY is assumed to be independent of $\mathcal{G}$ given
$\vecN{t,p}(i)$, limiting the dimensionality of the problem.
Note that two pairs of nodes $(i,j)$ and $(i',j')$ that
are close to each other in terms of graph distance are likely to have
overlapping neighborhoods, and hence a higher probability of sharing
neighborhood-specific features. Thus, link prediction probabilities
for pairs of nodes from the same region are likely to be similar,
as desired.

To make this statement precise, we will need to impose smoothness
properties on $\g$.  We will show that given appropriate assumptions
of smoothness (Assumption~\ref{assumption:smooth} in
Section~\ref{sec:consistency}), nonparametric
kernel estimators have desirable consistency properties.

Assume that the pair-specific features \dcg come from a finite set $\dcS$;
if not, they are discretized into such a set. For example, one may let
$\dcg$ record the number of common neighbors between $i$ and $j$ and
the last time a link appeared between these nodes ($\mbox{lastlink}$);
note that both are functions of $\vecN{t,p}(i)$.
Let $\dch = \{\dcn, \dcnplus; \forall s\in S\}$, where
\dcn are the number of node pairs in $\dcN{t-1}$ with feature vector $s$,
and \dcnplus the number of such pairs which were also linked by an
edge in the next timestep $t$.
%$$\dcn = \sum\limits_{i',j'\in\dcN{t-1}} I\{\dcgBASE{t}{i'}{j'} =
%s\} \quad \dcnplus = \sum\limits_{i',j'\in\dcN{t-1}} I\{\dcgBASE{t}{i'}{j'} = s\}\cdot
%I\{\dcYBASE{t}{i'}{j'}=1\}$$
In a nutshell, \dch tells us the chances of an edge being created in $t$  given
its features in $t-1$, averaged over the whole neighborhood
$\dcN{t-1}$---in other words, it captures the {\em change} of the
neighborhood around $i$ over one timestep.
%The intuition behind this
%choice of \dch is that past evolution patterns of a neighborhood are
%the most important characteristics in predicting future evolution.

One can think of \dch as a contingency table indexed by the features $s$.
Contingency tables are widely referred to as ``datacubes'' in the database
community, and we will adopt this terminology, refering to \dch as a datacube,
and a feature vector $s$ as the ``cell'' $s$ in the datacube with contents
$(\dcn, \dcnplus)$. Finiteness of $\dcS$ is necessary to ensure that
datacubes are finite-dimensional, which allows us to index them and
quickly find nearest-neighbor datacubes.

\subsection{The Estimator}
Our estimator of the function $\dcf(\cdot)$ at time $\nt$ is:
\begin{equation}
\label{eq:first-definition}
\tilde{g}\sub{\nt}(\dcfeat) = \frac{\sum\limits_{i',j',t'}\dcK(\dcfeat,
\dcfeatp)\cdot\dcYp}{\sum\limits_{i',j',t'}\dcK(\dcfeat,
\dcfeatp)},%\label{eq:hatf}
\end{equation}
where we factor the kernel function $\dcK(\dcfeat, \dcfeatp)$ into
neighborhood-specific and pair-specific  parts: $\dcKone(\dch, \dchp)\cdot
\xi(\dcg,\dcgp)$. Let $\dist(s,s')$ denote the $L_1$ distance between
features $s$ and $s'$, and let $n(s)$ denote the set of features at
$L_1$ distance $1$ from feature $s$.
We define $\xi(\dcg,\dcgp)$ as
{\small
\begin{align}
\label{eq:innerkernel}
%\xi(\dcg,\dcgp):=\frac{I\{\dcgp=\dcg\}+\bti\!\!\!\! \sum\limits_{s\in n(\dcgp)}\!\!\!\!\!I\{s=\dcg\}}{1+\bti |n(\dcgp)|},
\xi(\dcg,\dcgp):=\frac{I\{\dcgp=\dcg\}+\bti I\{\dist(\dcg,\dcgp)=1\}}{1+\bti |n(\dcg)|},
\end{align}
}
where $\bti$ is a bandwidth parameter which we will require to be
$O(\nt^{-(1/2+\epsilon)})$
for some $\epsilon>0$ in order to obtain consistency and distributional convergence.
%While the discrete component of the similarity function may seem strange, we show in equation~\eqref{eq:hatf2} that this decomposition effectively reduces our estimator to a kernel regression problem of estimating $E[\dcnplus|\dcn]$ with kernel function $\dcKone$ (equation~\ref{eq:tv}).
$\dcKone(\dch,\dchp)$ is a discrete analog of a continuous kernel function
(similar functions can be found in~\citet{Aitchison_Aitken} and \citet{Wang_Ryzin}).
As is the case with continuous kernel functions,  it has the property that as the bandwidth parameter
$\bt\rightarrow 0$, it is equal to one if and only if $\dch=\dchp$, and zero
otherwise. Similarly $\xi(\dcg,\dcgp)$ has the property that as $\bti\rightarrow 0$,
it approaches $I\{\dcgp=\dcg\}$.  This inner kernel function can also be extended
to features at $L_1$ distance two and so forth, while weighing those terms by
powers of $\bti$.
%Now $\dcK(\dcfeat, \dcfeatp)$ is defined as:
%{\small
%\begin{equation}
%%\hspace{-2em}\dcK(\dcfeat, \dcfeatp) = \dcKone(\dch, \dchp)\cdot
%%I\{\dcg=\dcgp\}\nonumber%\label{eq:K}
% \dcKone(\dch, \dchp)\cdot
%I\{\dcg=\dcgp\}\nonumber%\label{eq:K}
%\end{equation}
%}
%In other words, the similarity measure $\dcK(\cdot)$ computes the
%similarity between the two neighborhood evolutions (i.e., the
%datacubes),
%but only for pairs $(i', j')$ at time $t'$ that had exactly the same features as
%the query pair $(i, j)$ at $t$ (i.e., pairs belonging to the cell
%$s=\dcg$).
%Combining Equations~\ref{eq:hatf} and~\ref{eq:K}, we can get a

Plugging in the definition of the kernel in Equation~\ref{eq:first-definition},
we obtain the following interpretation of the estimator:
%The estimator $\dchatf(\dcfeat)$ is given by:
%{\small
\begin{equation}\label{eq:hatf2}
\begin{array}{l}
%\hspace{-1em}\dfrac{\sum\limits_{i',t'} \dcKone\left(\dch,
%\dchp\right)\cdot \sum\limits_{j'} \left[\xi(\dcg,\dcgp) \cdot \dcYp\right]}{\sum\limits_{i',t'}
%\dcKone\left(\dch, \dchp\right)\cdot \sum\limits_{j'}\xi(\dcg,\dcgp)}\\
\hspace{-1em}\tilde{g}\sub{\nt}(\dcfeat)= \frac{\sum\limits_{i',t'} \dcKone\left(\dch, \dchp\right)
\left(\dcnplusBASE{i'}{t'+1}{\dcg}+\bti\sum\limits_{s\in n(\dcg)}\eta^+_{i',t'+1}(s)\right)}{\sum\limits_{i',t'} \dcKone\left(\dch,
\dchp\right)\left(\dcnBASE{i'}{t'+1}{\dcg}+\bti\sum\limits_{s\in n(\dcg)}\eta_{i',t'+1}(s)\right)}.
\end{array}
\end{equation}
%}
Useful intuition can be obtained by considering the case $\bti=0$.
Here, given the query pair $(i, j)$ at time $t$, we look
inside cells for the query feature $s=\dcg$ in all neighborhood datacubes,
compute the average \dcnplusp and \dcnp in these cells after accounting for the
similarities of the datacubes to the query neighborhood datacube, and
use their quotient as the estimate of linkage probability.
Letting $\bti>0$ provides an estimator that deals more effectively
with sparsity by computing weighted averages of $\dcnplusp$ and
$\dcnp$ over features $s$ that are ``close'' to $\dcg$.

%and compute the linkage probability
%$\dcnplusp/\dcnp$ in these cells;
%these probabilities are then averaged by the similarities of the
%datacubes to the query neighborhood datacube.
Thus, the probability estimates are derived from historical instances where
(a) the feature vector of the historical node pair matches the query,
and (b) the local neighborhood is similar as well.
%Intuitively,
%given the query pair $(i, j)$ at time $t$, we find all historical
%pairs matching $\dcfeat$, look at their {\em neighborhoods} to collect
%all pairs with the same features as \dcfeat, then compute the
%probability of link formation in the next timestep among these pairs.
%These probabilities are weighted by the closeness of the
%neighborhoods to the query neighborhood.

Now, we need a measure of the similarity between neighborhoods, with
the goal of treating two neighborhoods as similar if they have similar
probabilities of generating links between node pairs with feature vector
$s$, for any $s\in S$.  To this end we could simply compare point estimates
$\dcnplusJUSTFUNC/\dcnJUSTFUNC$, but we also wish to account for the variance
in these estimates.  We achieve this by defining a similarity measure that
has a Bayesian flavor:
\begin{eqnarray}
\dcKone(\dch, \dchp) &=& e^{-\dcD\left(\dch, \dchp\right)/b_T} \quad
(0<\bt<1)\label{eq:tv}\\
\dcD(\dch, \dchp)&=&\sum\limits_{s\in S} \mbox{TV}(X,Y)\nonumber\\
X&\sim& \mathcal{B}\left(\dcnplus,\dcn-\dcnplus\right) \nonumber\\
 Y&\sim& \mathcal{B}\left(\dcnplusp,\dcnp-\dcnplusp\right),\nonumber
\end{eqnarray}
%where, \dcD(\dch, \dchp) equals:
%\begin{align*}
%\hspace{-2em}\sum\limits_{s\in S} \mbox{TV}\left(\mathcal{B}\left(\dcnplus,\dcn-\dcnplus\right),
%	\mathcal{B}\left(\dcnplusp,\dcnp-\dcnplusp\right)\right)
%\end{align*}
%\begin{align*}
%\sum\limits_{s\in S} \mbox{TV}\left(\mathcal{N}\left(\hat{p}_s,
%	\frac{\hat{p}_s(1-\hat{p}_s)}{\dcn}\right),
%	\mathcal{N}\left(\hat{p}'_s,
%	\frac{\hat{p}'_s(1-\hat{p}'_s)}{\dcnp}\right)\right)
%\end{align*}
%%\label{eq:tv}
%\\\nonumber
%\dcKone(\dch, \dchp) & = & \dcD\left(\sum\limits_{s\in S}
%\mbox{TV}\left(\mathcal{N}\left(\hat{p}_s,
%\frac{\hat{p}_s(1-\hat{p}_s)}{\dcn}\right),
%\mathcal{N}\left(\hat{p}'_s, \frac{\hat{p}'_s(1-\hat{p}'_s)}{\dcnp}\right)\right)\right)\label{eq:tv}\\
%\dcD(x) & = & b^x (0<b<1),
%\qquad \hat{p}_s=\frac{\dcnplus}{\dcn}, \qquad
%\hat{p}'_s=\frac{\dcnplusp}{\dcnp}\nonumber
%\end{eqnarray}
where $\mbox{TV}(X,Y)$ denotes the total variation distance between the distributions of
$X$ and $Y$, ${\cal B}$ is the beta distribution and $\bt\in(0,1)$ is a bandwidth parameter.
We will require $\bt = O(\nt^{-(1/2+\theta)})$ for some  $\theta>0$ to obtain appropriate
rates when we study the consistency and distributional convergence of our estimator.

%such that $\dcKone(\dch, \dchp)\rightarrow 0$ as $b\rightarrow 0$
%unless $\dcD\left(\dch, \dchp\right)=0$.

%\mypara{Consistency}

%We chose features which were shown to be useful in prior empirical
%work~\cite{TylendaTimeaware,kleinberg04}, we set $\dcg =
%\{\mbox{cn}_t(i,j), \ell\ell_t(i,j)\}$, where $\mbox{cn}_t(i,j)$
%represents the number of common neighbors of $i$ and $j$ at time $t$,
%and $\ell\ell_t(i,j)$ the number of timesteps elapsed since the edge
%$i\rightarrow j$ last occurred in $\vecG{t,p}$.  While we only use two
%graph-based features of the pair $i,j$, it is possible to use more
%features, and also include features composed from `meta-data' of the
%entities.  For example, one could use the age, or gender of entities
%in a social network, etc.  Features based on node and edge labels can
%also be used here.  All feature values are binned logarithmically in
%order to combat sparsity in the tails of the feature distributions.

\mypara{Remarks}
To better understand our choice of estimator, consider by way of
contrast a simple estimator that computes the fraction of pairs for
which the feature $\mbox{lastlink}$ was equal to $k$ at time $t'$ and
which formed an edge at time $t'+1$ (for $k=1, 2, \ldots$).
This approach suffers from two
key problems that make it perform poorly on real-world graphs.
First, it does not allow for local variations in the link-formation
fractions, as would be expected for communities evolving differently
within the same graph.  We address this problem by maintaining a separate
datacube for each local neighborhood.  The second, more subtle, problem
is the implicit assumption of stationarity---a node's link-formation
probabilities are assumed to be time-invariant functions of the datacube
features.  This assumption does not allow for seasonal changes in linkage
patterns, or for a transition from slow to fast growth, etc.
Our model addresses this issue by finding historical neighborhoods
from some previous time $t'$ with datacubes similar to the query datacube,
and uses their evolution from $t'$ to $t'+1$ to predict link formation in
the next time step for the current neighborhood. This helps us
learn nonlinear trends.% of link formation.

Our estimator also has the virtue that it combats sparsity by aggregating
data across similarly-evolving communities even if they are separated by graph
distance and time.  That said, sparsity remains a serious issue, and we
provide a further discussion of sparsity in the following section.

Finally, note that we build the datacube so as to encode the recent change
of a neighborhood, and not just the distribution of features in the
neighborhood.  Thus, for example, two neighborhoods may have the same datacube
if the fraction of $\mbox{lastlink}=1$ node pairs that formed an edge
in the next timestep is the
same in both neighborhoods, and not if they both merely had the same
number of $\mbox{lastlink}=1$ pairs. Thus, it is the change in link structure that
drives the estimation of linkage probabilities. Moreover, two neighboring
nodes may end up having very similar datacubes, and will end up forming
links in a similar way, whereas very different datacubes will reflect the
variations in link formation patterns among different communities.

%Another question about the model choice can be: why use graph neighborhoods?
%One could simply build one datacube for every edge, and learn using a
%time-series of datacubes for a given pair how likely a link formation
%is in the next timestep. For example we may find that small lastlink
%and high common neighbors is a strong indicator of future
%connectivity. Unfortunately, this approach is too local, not to
%mention expensive, and is bound to suffer from the lack of data in
%case of sparse graphs. A second idea is to build one datacube for each
%graph from $t\rightarrow t+1$ by aggregating over all pairs of nodes
%in a graph. This approach suffers from being too global, and is also
%expensive. A simple way to bridge the gap between these to extremes
%is to build datacubes from
%local neighborhoods around a node, and learn using these datacubes.

\subsection{Sparsity}
For sparse graphs, or short time series, two practical problems can
arise.  First, a node $i$ can have zero degree and hence an empty
neighborhood.  To cope with this issue, we define the neighborhood of
node $i$ as the union of two-hop neighborhoods over the last $p$ timesteps.
Second, and more problematically, the $\dcnJUSTFUNC$ and $\dcnplusJUSTFUNC$
values obtained from kernel regression can be small, yielding an estimated
linkage probability $\dcnplusJUSTFUNC/\dcnJUSTFUNC$ that is unreliable
numerically.

We offer a threefold solution to this problem, the first element of
which is already present in our estimator.  (a) The inner kernel $\xi$
(Equation~\ref{eq:innerkernel}) combines $\dcnJUSTFUNC$ and $\dcnplusJUSTFUNC$
with a weighted average of the corresponding values for any $s'$ that
are ``close'' to $s$, the weights encoding the similarity between $s'$ and $s$.
(b) In determining a final ranking, instead of using $\dcnplusJUSTFUNC/\dcnJUSTFUNC$
directly, we use the lower end of the $95\%$ Wilson score interval~\citep{Wilson:1927}.
The node pairs that are ranked highest according to this ``Wilson score''
are those that have high estimated linkage probability $\dcnplusJUSTFUNC/\dcnJUSTFUNC$
\emph{and} $\dcnJUSTFUNC$ is high (implying a reliable estimate).
(c) We use a ``backoff'' smoothing procedure for the Wilson scores,
in which the raw scores are smoothed against the scores obtained
from a ``prior'' datacube, which is the average of all historical
datacubes.  The degree of smoothing depends on $\dcnJUSTFUNC$.
This can be thought of as a simple hierarchical model, where the
lower level (set of individual datacubes) smooths its estimates
using the higher level (the prior datacube).

%% file: hashing.tex
%\vspace{-1em}
\section{Fast search using LSH}
%\vspace{-1em}
\label{sec:lsh}

% total variation = hamming
% doing LSH
% picking k

A naive implementation of the nonparametric estimator in
Equation~\eqref{eq:hatf2} computes kernel similarity between the query datacube and
all $n$ datacubes for each of the $T$ timesteps for each prediction,
which can be infeasibly slow for large graphs.  To obtain a more
computationally tractable estimator, we consider only the top-$r$
closest neighborhoods (in terms of the largest kernel similarities).
The value of $r$ is a parameter of the algorithm; for our experiments
we use $r=20$.  What is needed to make this practical is a fast
method (one that runs in sublinear time) to quickly find the top-$r$
closest neighborhoods.

We achieve this by using locality sensitive hashing (LSH)~\citep{Motwani}.
Hashing is often used in databases for fast ``table-lookups'' or retrieving
matching items from a large database. The key component is a hash function
that maps a given ``key'' or object to a certain hash value. In order to
search for a particular key, we compute the hash value and do a table lookup
with this value.
%An ideal hash function leads to few collisions (cases
%where two different keys map to the same value).  Rather recently
The concept of ``locality sensitive'' hashing refers
to hash functions having the property that, with high probability, two
``similar'' data items are hashed to the same value. This facilitates
approximate nearest neighbor search, and is suitable for high-dimensional
spaces, where traditional nearest neighbor search techniques are often
infeasible.

The standard LSH method operates on bit sequences, and maps sequences with
small Hamming distance to the same hash bucket. In our setting, we must hash
datacubes, and use the total variation distance metric.  We make use
of the fact that total variation distance between discrete distributions
is half the $L_1$ distance between the corresponding
probability mass functions. %We discretize the continuous distributions
If we could approximate the probability distributions in each datacube cell
with bit sequences, then the $L_1$ distance would just be the Hamming
distance between these sequences, making our setting amenable to the
use of standard LSH.  We achieve this with three steps:
\begin{description}
\item[Conversion to bit sequence] The key idea is to approximate the
linkage probability distribution by discretization.  We first discretize
the range $[0, 1]$ (since we deal with probabilities) into $B_1$ buckets.
For each bucket we compute the probability mass $p$ falling inside it.
This $p$ is encoded using $B_2$ bits by setting the first $\lfloor p
B_2\rfloor$ bits to 1, and the others to 0. In this way the entire
distribution (i.e., one cell) is represented by $B_1 B_2$ bits.
As a result the entire datacube can now be stored in $|S| B_1 B_2$ bits.
However, in all our experiments, datacubes were very sparse with only
$M\ll |S|$ cells ever being non-empty (usually, 10-50); thus,
we use only $M B_1 B_2$ bits in practice.
The Hamming distance between two pairs of $MB_1B_2$ bit vectors yields
the total variation distance between datacubes (modulo a constant
factor).
\item[Distances via LSH] We create a hash function by picking a uniformly
random sample of $k$ bits out of $M B_1 B_2$. For each hash function,
a hash table is created to store all datacubes whose hashes are
identical in these $k$ bits.  We use $\ell$ such hash functions.
A query datacube is first hashed using each of these $\ell$
functions. Then we create a {\em candidate set} containing
$O(\max(\ell,\text{r}))$ of distinct datacubes sharing any of
these $\ell$ hashes.  The total variation distance of these candidates
to the query datacube is computed explicitly, yielding the closest
matching historical datacubes.
\item[Picking $k$] The number of bits $k$ is crucial in balancing accuracy
versus query time: while a large $k$ hashes all datacubes to their own hash
bucket, returning a few or no matches to the query, a small $k$ bunches many
datacubes into the same bucket, decreasing the probability of finding the
`true' near neighbors.  In the spirit of~\citet{Motwani},
we do a binary search to find the $k$ for which
the average hash-bucket size over a query workload is just enough to
provide the desired top-$20$ matches.
We evaluate the accuracy of this approach in Section~\ref{sec:exp}.
\end{description}
We conclude this section with two additional points. First, we never create
the entire bit representation of $M B_1 B_2$ bits explicitly; only the
hashes need to be computed, taking $O(k\ell)$ time. Second, the main cost
in the algorithm is in creating the hash table, which needs to be done once
as a preprocessing step.  Query processing is extremely fast and sublinear,
since the candidate set is much smaller than the size of the training set.

%% file: exp.tex
\section{Experiments}
\label{sec:exp}
We start  by introducing several baseline algorithms, and our evaluation metric.
These baselines were picked carefully from previous work as being those that
have yielded state-of-the-art performance in a range of link prediction tasks.
In our first set of experiments we use simulated data to compare the performance
of our algorithm to these baselines, focusing on situations involving seasonality
in link formation.  Second, we study the performance of our algorithm and the
baselines on several real-world graphs: a sensor network, two co-authorship
graphs, and a graph of Facebook employees.  Finally, we investigate the
computational scaling of our approach, comparing the improvement in runtime
of the LSH-based algorithm to an exact algorithm, and investigating the effect
of the LSH bit-size $k$ on accuracy.

\subsection{Baselines and metrics}
We compare our nonparametric network inference algorithm (\expnp) to the
following baselines which, although quite naive, have proved difficult
to beat in practice~\citep{kleinberg04,TylendaTimeaware}:
%\noindent
\begin{description}
\denselist
\item \expll: ranks pairs using ascending order of \textit{last time
of linkage}~\citep{TylendaTimeaware}.
\item \expcn (last timestep): ranks pairs using descending order of
the number of \textit{common neighbors}~\citep{kleinberg04}.
\item \expaa (last timestep): ranks pairs using descending order of
the \textit{Adamic-Adar} score~\citep{AdaAda03}, a weighted variant
of common neighbors which it has been shown to outperform~\citep{kleinberg04}.
\item \expkz (last timestep): extends \expcn to paths with length
greater than two, but with longer paths getting exponentially smaller
weights~\citep{Katz:1953}.
\item \expcnall, \expaaall, \expkzall: \expcn, \expaa, and \expkz
computed on \textit{the union of all graphs until the last timestep}.
\end{description}
For \expnp, we only predict on pairs which are in the
neighborhood (generated by the union of two-hop neighborhoods of the last
$p$ timesteps) of each other.
We deliberately used a simple feature set for \expnp, setting
$\dcg = \{\mbox{cn}_t(i,j), \ell\ell_t(i,j)\}$ (i.e., common neighbors
and last-link) and not using any external ``meta-data'' (e.g., stock
sectors, university affiliations, etc.).
All feature values were binned logarithmically in order to
combat sparsity in the tails of the feature distributions.
Strictly speaking, our feature $\ell_t(i,j)$ should be capped at $p$.
However, since the heuristic \expll uses no such capping, for
fairness, we used the uncapped ``last time a link appeared'' as
the feature $\ell_t(i,j)$ for the pairs we predict on. The bandwidth
$\bt$ was picked by cross-validation.

For any graph sequence $(G_1, \ldots, G_T)$, we test link prediction
accuracy on $G_T$ for a subset $S_{>0}$ of nodes with non-zero degree
in $G_T$.  Each algorithm is provided training data up to and including timestep
$T-1$, and must output, for each node $i\in S_{>0}$, a ranked list of
nodes in descending order of probability of linking with $i$ in $G_T$.
For purposes of efficiency, we only require a ranking on the nodes
that have ever been within two hops of $i$ (call these the candidate
pairs); all algorithms under consideration predict the absence of a
link for nodes outside this subset. We compute the AUC score for
predicted scores for all candidate pairs against their actual edges
formed in $G_{T}$.
%for each $i\in S_{>0}$, we compute the AUC score for the
%predicted ranking against the actual edges formed in $G_T$, and we
%report the mean AUC over $S_{>0}$.

\subsection{Simulations}
In this section we compare \expnp to the baseline algorithms using
simulated data, focusing on seasonal patterns as an example of the
kind of nonlinear behavior that may be difficult to capture with the
heuristic methods.  We simulated a model of Hoff et al~\citep{HoffSlides}
that posits an independently drawn ``feature vector'' for each node.
Time moves over a repeating sequence of seasons, with a different set
of features being ``active'' in each. Nodes with these features are more
likely to be linked in that season, though noisy links also exist.
The user features also change smoothly over time, to reflect changing
user preferences.

\paragraph{Model Specifications}
We generate features $u_{i,t}\in \mathbb{R}^6$ for node $i$ at time $t$. Node pair $\{i,j\}$ has a link if $u_{i,t}^TL_t u_{j,t}$ exceeds one, where $L_t$ is a matrix governing feature interactions. We now formally define $u_{i,t}$ and $L_t$.

For every node we generate a two features $a_i, b_i\sim \mathcal{N}(\mathbf{0}_6, I_{6\times 6})$. The six features are  divided into three blocks each of size two.
Now, for the $t^{th}$ timestep, the features of node $i$ are given by
$u_{i,t}=(c_t a_i+(1-c_t)b_i)/\sqrt{c_t^2+(1-c_t)^2}$, where  $c_t=\frac{T-t}{T-1}$. The normalization ensures identical variance of features at any timestep. For $t=3i+j$, the feature interaction matrix $L_t$ is generated as follows:
\begin{align*}
B_{k,\ell}&=\mu&\mbox{For $k,\ell\in\{2j+1,2j+2\}$,}\\
L_{t}&=B+\sigma\frac{R+R^T}{2}&\mbox{Where $R\sim N(0,1)^{k\times k}$,}
\end{align*}
where $\mu$ represents the signal and $\sigma$ represents the noise.
%In particular, for node $i$ we generate feature vectors $a_i\in \mathbb{R}^6$, and $b_i\in \mathbb{R}^6$, where the features are divided into three blocks each of size two.
%Now, for the $t^{th}$ timestep, the features of node $i$ are given by
%$u_{i,t}=(c_t a_i+(1-c_t)b_i)/\sqrt{c_t^2+(1-c_t)^2}$, where  $c_t=\frac{T-t}{T-1}$. For $t=3i+j$, we construct a $k\times k$ matrix $B$, whose $j^{th}$ diagonal block corresponding to the features in block $j$ equals $\mu$. Now $L_t$ is set to $B$ plus a symmetric matrix whose entries are generated from $N(0,\sigma)$.
% Node pair $\{i,j\}$ has a link if $u_{i,t}^TL_t u_{j,t}$ exceeds one.
%
% MJ We also investigated a stationary data-generating process, where...

\begin{figure}[ht]
\hspace*{-.2in}
	\centering
\begin{minipage}[t]{.45\textwidth}
%	\subtable[Caption of subfigure 1]{
\vspace{0pt}
\includegraphics[height=60mm]{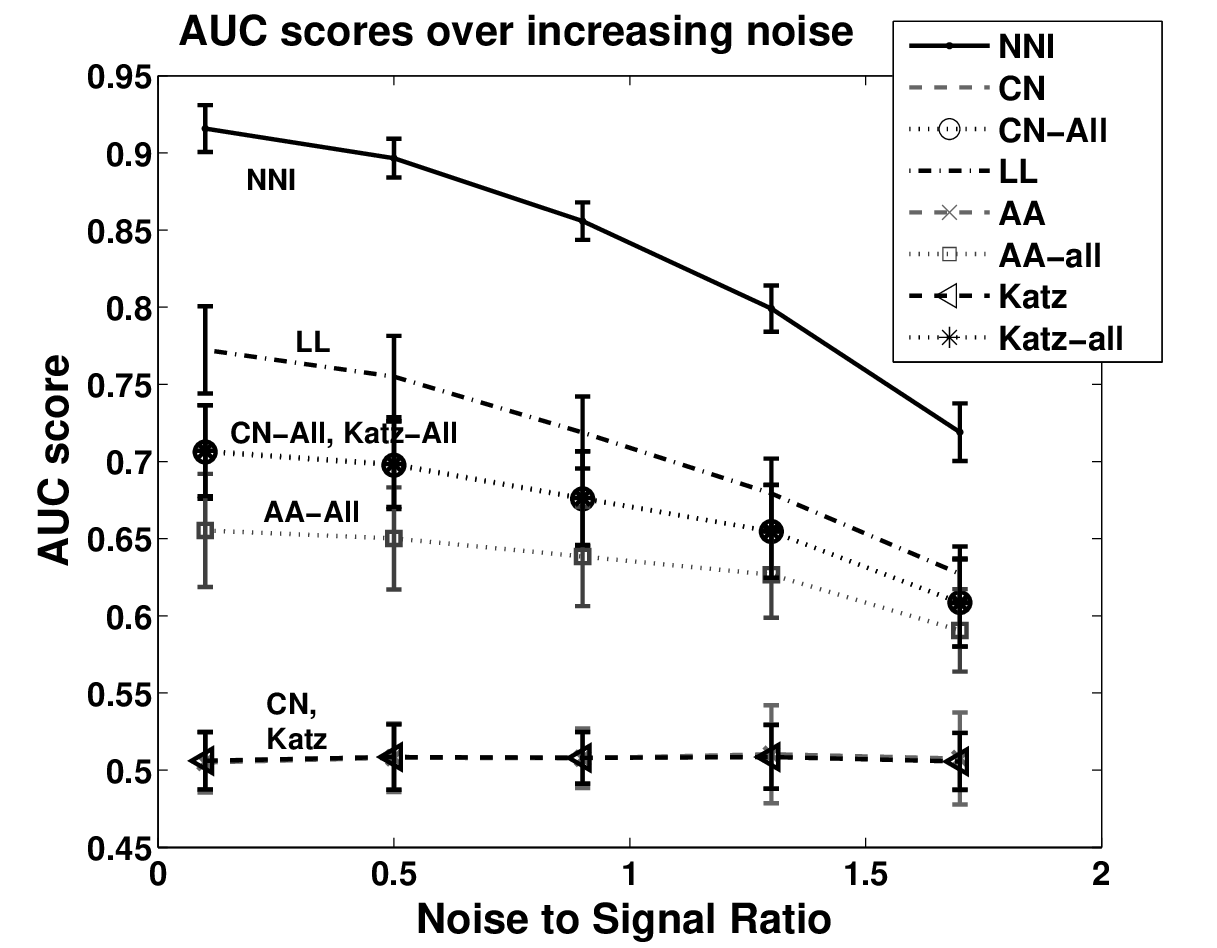}
\vspace{-2em}
\caption{Simulated graphs: Effect of noise.}
\label{fig:incr-noise}
\end{minipage}
\hfill
%\hspace{.8in}
%	\subtable[Caption of subfigure 3]{%{\small
\begin{minipage}[t]{.4\textwidth}
\vspace{2em}
{\small
\begin{tabular}{|c|cc|}\hline
% & \expnp & \expll & \expcn &\expaa& \expkz &\expcnall&\expaaall&\expcnall\\\hline
& \seasonal & \stationary \\\hline
\expnp&  $\mathbf{.91\pm .01}$&$\mathbf{0.99\pm .005}$\\\hline
\expll&  $.77\pm .03$ & $0.97\pm .006$\\\hline
\expcn&  $.51\pm .02$ & $0.97\pm .01$\\
\expaa&  $.51\pm .02$ & $0.95\pm .02$\\
\expkz&  $.50\pm .02$ & $0.97\pm .01$\\\hline
\expcnall& $.71\pm .03$ & $0.86\pm .03$\\
\expaaall& $.65\pm .04$ & $0.71\pm .04$\\
\expkzall& $.71\pm .03$ & $0.87\pm .03$\\\hline
\end{tabular}
}
\vspace{5em}
\caption{Average AUC for $T=20$ timesteps.}
\label{tbl:simul}
\end{minipage}
\vspace{-0em}
%\label{tbl:realworld}
	%}
%	\caption[Optional caption for list of figures]{Figure~\ref{fig:sensor-net} AUC scores for a \textbf{periodic} sensor network. Table~\ref{tbl:realworld}Avg.\ AUC in real world \textbf{Stationary} graphs}
%	\label{fig:subfigureExample}
	\end{figure}

%We use these graphs to show the following:
%effect of increasing noise on all our baselines, comparison between
%the exact search algorithm and our hashing scheme, and lastly the
%importance of choosing the right number of bits for hashing.
%\begin{figure}[htb]
%\includegraphics[height=65mm]{incr-noise-rocfull.eps}
%\caption{Simulated graphs: Effect of noise}
%\label{fig:incr-noise}
%\end{figure}
We generated $100$-node graphs over $20$ timesteps using $3$ seasons,
and plotted AUC averaged over $10$ random runs for several
noise-to-signal ratios (Fig.~\ref{fig:incr-noise}).
%In figure~\ref{fig:incr-noise} we see that
\expnp consistently outperformed all other baselines by a large margin.
Clearly, seasonal graphs have nonlinear linkage patterns: the best predictor
of links at time $T$ are the links at times $T-3$, $T-6$, etc., and
\expnp is able to learn this pattern.  By contrast, \expcn,
\expaa, and \expkz are biased towards predicting links between pairs
which are linked (or have short paths connecting them) at the previous
timestep $T-1$; this implicit smoothness assumption makes them perform
poorly; indeed, they behaved essentially as poorly as a random predictor
(an AUC of 0.5).

Baselines \expll, \expcnall, \expaaall and \expkzall use information
from the union of all graphs until time $T-1$. Since the off-seasonal
noise edges are not sufficiently large to form communities, most of
the new edges come from communities of nodes created in season.
This is why \expcnall, \expaaall and
\expkzall outperform their ``last-timestep'' counterparts. As for
\expll, since links are more likely to come from the last seasons, it
performed well, although poorly compared to \expnp. Also note that the
changing user features forces the community structures to change
slowly over time; in our experiments, \expcnall performed worse than
it would were there was no change in the user features, since the
communities stayed the same.

Table~\ref{tbl:simul} summarizes the average AUC scores for graphs
with seasonality, and also presents results for stationary data.
In both cases, the noise was set to the smallest value in
Fig.~\ref{fig:incr-noise}.  For the stationary data, links formed in
the last few timesteps of the training data are good predictors of future
links, and so \expll, \expcn, \expaa and \expkz all performed very well.
Interestingly, \expcnall, \expaaall and \expkzall were worse than their
``last time-step'' variants, presumably owing to the slow movement of the
user features.  As for \expnp, it performed slightly better than all other
methods for the stationary data, in addition to showing substantial
improvements over the other methods for the seasonal networks.

\subsection{Real-world graphs}
\begin{figure}[ht]
\hspace*{-.4in}
	\centering
\begin{minipage}[t]{.45\textwidth}
%	\subtable[Caption of subfigure 1]{
\vspace{0pt}
\includegraphics[height=75mm, angle=90]{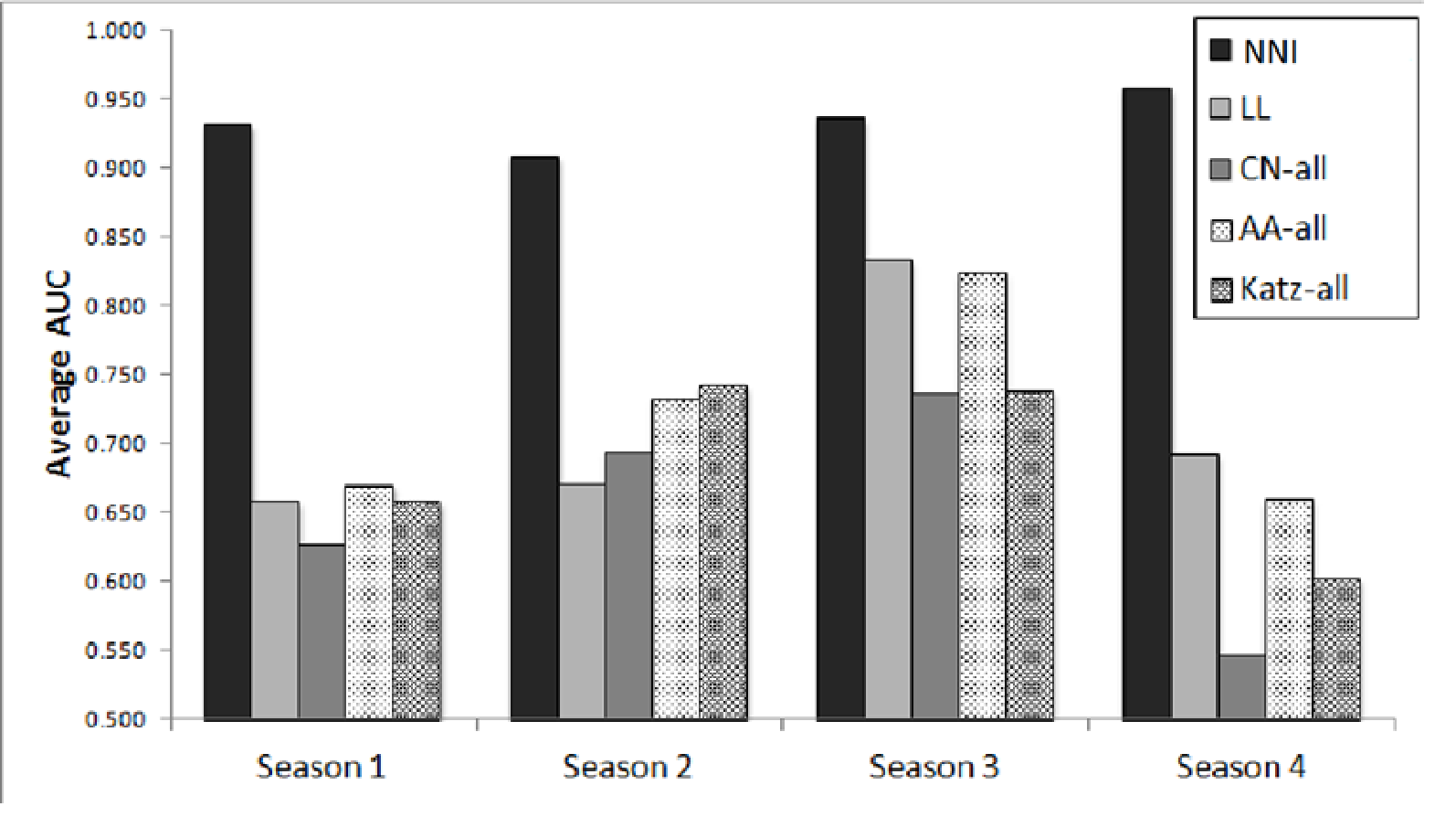}
\vspace{-2em}
\caption{\label{fig:sensor-net}AUC scores for a \textbf{periodic} sensor network}
\end{minipage}
\hfill
%\hspace{.8in}
%	\subtable[Caption of subfigure 3]{%{\small
\begin{minipage}[t]{.4\textwidth}
\vspace{0pt}
\begin{tabular}{|c|c@{\hspace{.25em}}c@{\hspace{.25em}}c@{\hspace{.25em}}c@{\hspace{.25em}}|}\hline
% & \expnp & \expll & \expcn &\expaa& \expkz &\expcnall&\expaaall&\expcnall\\\hline
& NIPS & HepTh & Citeseer & Facebook\\\hline
\expnp&  $\mathbf{.87}$ & $\mathbf{.89}$ & $\mathbf{.89}$ & $.82$\\\hline
\expll&  $.84$ & $.87$ & $\mathbf{.90}$ & $.81$\\\hline
\expcn&  $.74$ & $.76$ & $.69$ & $.70$\\
\expaa&  $.84$ & $.87$ & $\mathbf{.90}$ & $.71$\\
\expkz&  $.75$ & $.83$&$.83$&$.78$\\\hline
\expcnall& $.56$ & $.62$&$.70$&$\mathbf{.87}$\\
\expaaall& $.77$ & $.83$&$.83$&$\mathbf{.89}$\\
\expkzall& $.67$ & $.71$&$.81$&$\mathbf{.89}$\\\hline
\end{tabular}\label{tbl:realworld}
\vspace{3em}
\caption{Average AUC for co-authorship and Facebook graphs.}
\end{minipage}
	%}
%	\caption[Optional caption for list of figures]{Figure~\ref{fig:sensor-net} AUC scores for a \textbf{periodic} sensor network. Table~\ref{tbl:realworld}Avg.\ AUC in real world \textbf{Stationary} graphs}
%	\label{fig:subfigureExample}
	\end{figure}

We begin by presenting results on a $24$-node sensor network where
each edge represents the successful transmission of a
message\footnote{\url{http://www.select.cs.cmu.edu/data}}.
We considered up to $82$ consecutive measurements. These networks
exhibit clear periodicity; in particular, a different set of sensors turn on
and communicate during four different periods.  Fig.~\ref{fig:sensor-net}
shows our results for these four periods averaged over several cycles.
The maximum standard deviation, averaged over the periods, was $.07$.
We do not show results for \expcn, \expaa and \expkz, as they all performed
no better than a random predictor. \expnp significantly outperformed
the baselines, confirming the results from the simulation experiments
for seasonal graphs.

We also present results on three dynamic co-authorship graphs:
the Physics ``HepTh'' community ($\nn=14,737$ nodes, $e=31,189$ total
edges, and $T=8$ timesteps), NIPS ($\nn=2,865$, $e=5,247$, $T=9$), and
authors of papers on Citeseer ($\nn=20,912$, $e=45,672$, $T=11$) with
``machine learning'' in their abstracts. Each timestep considers $1-2$
years of papers (so that the median degree at any timestep is at least $1$).
%Finally, we also considered a dynamic stock-correlation network; in
%this case the nodes are a subset of stocks in the S\&P500, and two
%stocks are linked if the correlation of their daily returns over a
%two-month window exceeds 0.8 ($\nn=424$, $e=41,699$, $T=49$).
Finally we also considered a dynamic undirected network of Facebook employees over several weeks, where the nodes represent employees and edges are formed if one employee mentions another in a post. The network contains above five thousand nodes, and above $100,000$ edges in total.
%\begin{narrow}{-3in}{0in}
%	\end{narrow}

%\begin{narrow}{-.5in}{0in}
%\begin{figure}\label{fig:sensor-net}
%\includegraphics[height=45mm]{CarlosSensorResults.eps}
%\vspace{-.1in}
%\caption{AUC scores for a \textbf{periodic} sensor network}
%\end{figure}
%\end{narrow}
%\begin{narrow}{-.1in}{-.2in}
%\begin{table}[!t]
%\vspace{-.5em}
%\begin{center}
%{\small
%\begin{tabular}{|c|c|c|c|c|}\hline
%% & \expnp & \expll & \expcn &\expaa& \expkz &\expcnall&\expaaall&\expcnall\\\hline
%& NIPS & HepTh & Citeseer & S\&P500\\\hline
%\expnp&  $\mathbf{.87}$ & $\mathbf{.89}$ & $\mathbf{.89}$ & $.73$\\\hline
%\expll&  $.84$ & $.87$ & $\mathbf{.90}$ & $.70$\\\hline
%\expcn&  $.74$ & $.76$ & $.69$ & $.72$\\
%\expaa&  $.84$ & $.87$ & $\mathbf{.90}$ & $.70$\\
%\expkz&  $.75$ & $.83$&$.83$&$.76$\\\hline
%\expcnall& $.56$ & $.62$&$.70$&$\mathbf{.79}$\\
%\expaaall& $.77$ & $.83$&$.83$&$.76$\\
%\expkzall& $.67$ & $.71$&$.81$&$\mathbf{.79}$\\\hline
%\end{tabular}
%\caption{Average AUC for real-world graphs}
%\label{tbl:realworld}
%}
%\end{center}
%\vspace{-1.5em}
%\end{table}
%\end{narrow}

%\vspace{-1em}
Table~\ref{tbl:realworld} shows the average AUC for all algorithms
for the co-authorship graphs and the Facebook graph.  For the
co-authorship graphs, we do not expect to see seasonal variation,
and we expect a relatively simple model to be effective; authors
will tend to keep working with a similar set of co-authors over
time.  For such graphs, \citet{TylendaTimeaware} have shown that
\expll is the best heuristic, and we replicate that result here.
Our kernel-based approach, \expnp, also performs well on these
graphs, slightly outperforming \expll.  For the Facebook graph,
employees in the same research group tend to post more messages mentioning each other, and hence algorithms working on all edges seen so far should intuitively pick up this community structure. This is indeed reflected in the AUC scores. \expcnall, \expaaall and \expkzall perform the best. These algorithms outperform \expnp, primarily because
they count paths through edges that exist in different timesteps,
which is not allowed in our model.

In summary, for graphs having a seasonal trend, \expnp is the
best method by a large margin.  For the co-authorship graphs,
\expnp remains the best algorithm, although \expll is also
effective.  For the correlation graph, \expkzall is the best
algorithm, but its performance is quite poor on the co-authorship
graphs and the seasonal graphs.  Overall, the performance of
\expnp dominates that of the other algorithms.

\subsection{Evaluation of LSH}
We have found the use of LSH to be essential in our experimental
work.  In this section we provide quantitative support for this
assertion.

\begin{figure}
\begin{tabular}{c@{}c}
% \hspace{-2.5em}
 \includegraphics[height=50mm]{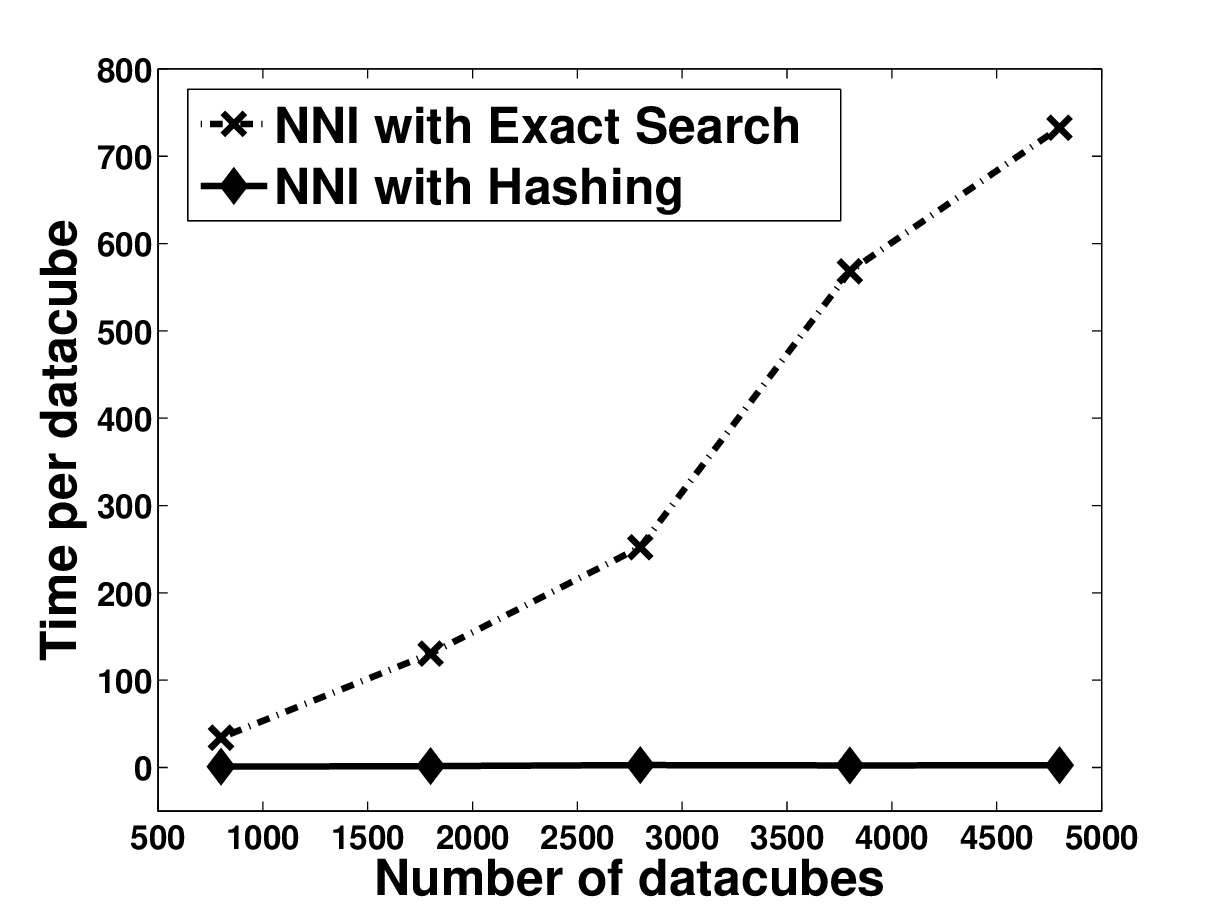} &
%  \hspace{-2.5em}
\includegraphics[height=50mm]{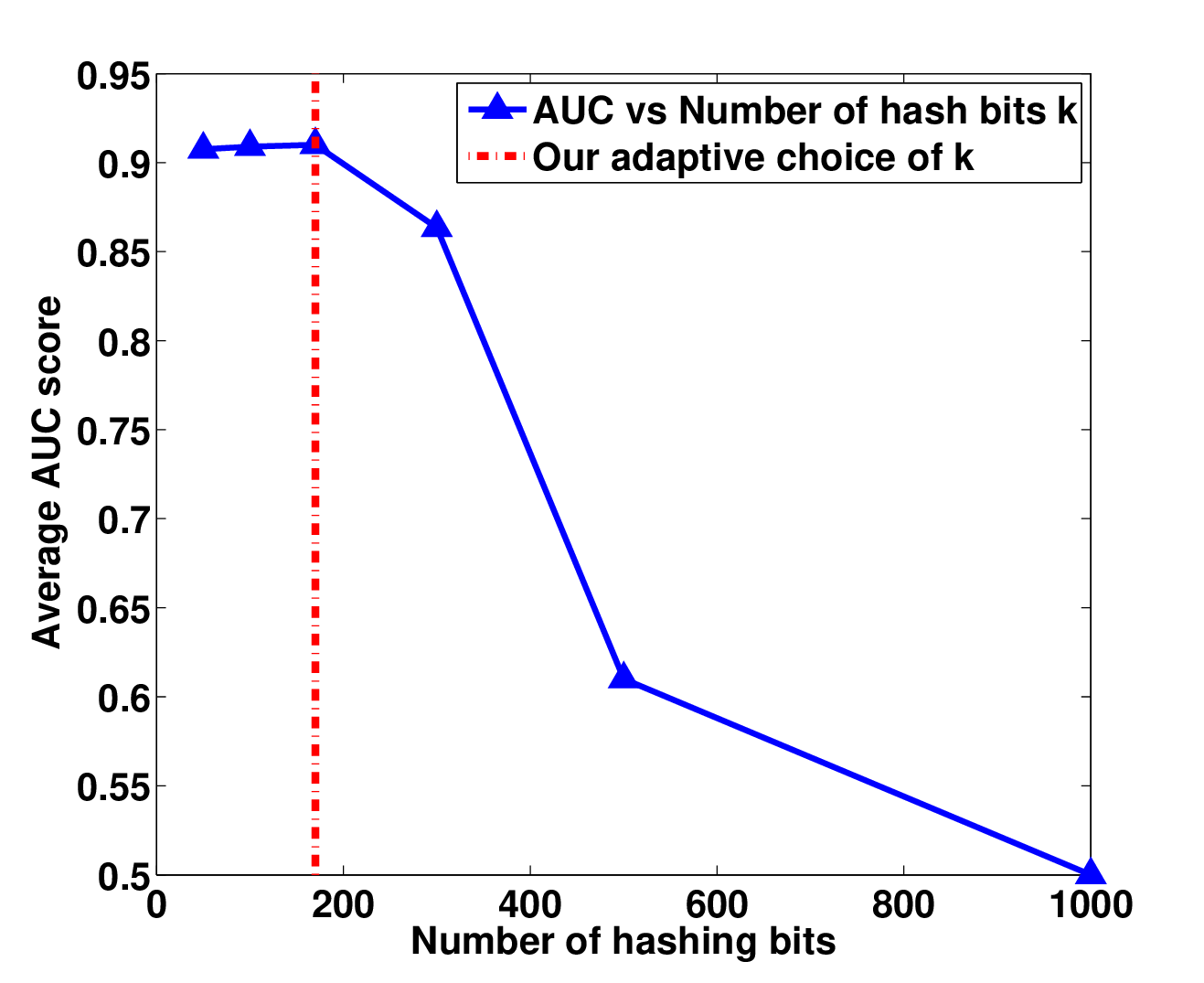}\\
 \hspace{-3em}(a) Time vs. \#-datacubes  & (b) AUC vs. hash bitsize $k$
%& (our adaptive algorithm picks $k=170$)
\end{tabular}
\vspace{-1em}
\caption{Time and accuracy using LSH.}
%(a) LSH requires far less time than exact search. \textbf{LSH has AUC within 0.4\% of Exact}.
%(b) The AUC score vs. increasing number of hashbits. Our adaptive algorithm picks $k=170$}%
\label{fig:sensitivity}%
\vspace{-2em}
\end{figure}

\mypara{Exact search vs.\ LSH}
In Fig.~\ref{fig:sensitivity}(a) we plot the time taken to perform
top-$20$ nearest neighbor search for a query datacube using simulated
data.  We fixed the number of nodes at $100$, and increased the number
of timesteps. As expected, the exact search time increases linearly
with the total number of datacubes, whereas LSH searches in nearly
constant time. Also, the AUC score of \expnp with LSH is within 0.4\%
of that of the exact algorithm on average, implying minimal loss of
accuracy from LSH.

In our experiments with real-world graphs, the query time per datacube
using LSH was quite small: $0.3$s for Citeseer, $0.4$s for NIPS,
$0.6$s for HepTh, and $1.9$s for Facebook.  Exact search was infeasible
for these large-scale graphs.

%\mypara{Choice of the bandwidth parameter**********}
%The choice of the bandwidth parameter is crucial in kernel regression.
%In table~\ref{table:h} we show accuracy averaged over 10 runs for
%increasing values of the bandwidth parameter. We also switch off the
%smoothing using the prior datacube for these experiments.  We see that
%for very small $b$, since all the kernel similarity values are
%	vanishingly small, the AUC scores become worse. Interestingly the
%	AUC scores seem to be stable for a large range of bandwidth
%	parameters, and we pick the bandwidth in this range.
%\begin{narrow}{0in}{-1in}
%\end{narrow}
%\begin{figure}\hspace{-.3in}
%%\centering
%\parbox{1.5in}{\includegraphics[height=35mm]{hashing-VS-exact.png}}%
%\qquad
%\begin{minipage}{1.2in}%
%\includegraphics[height=35mm]{acc_vs_k.png}
%\end{minipage}\\
%\centering
%(a) \hspace{1.3in} (b)
%% (a) Time vs. \#-datacubes  \hspace{1.3in}(b) AUC vs. hash bitsize $k$
%\caption{(a) LSH requires far less time than exact search. \textbf{LSH has AUC within 0.4\% of Exact}.
%(b) The AUC score vs. increasing number of hashbits. Our adaptive algorithm picks $k=170$}%
%\label{fig:sensitivity}%
%\end{figure}

\mypara{Number of Bits in Hashing}
Fig.~\ref{fig:sensitivity}(b) shows the
effectiveness of our adaptive scheme to select the number of hash bits
(Section~\ref{sec:lsh}). For these experiments, we
turned off the smoothing based on the prior datacube.
As $k$ increases, the accuracy goes down to $50\%$, as a
result of the fact that \expnp fails to find any matches of the query
datacube. Our adaptive scheme finds $k\sim 170$, which yields
the highest accuracy.  Note also that larger $k$ translates to
fewer entries per hash bucket and hence faster searches, and thus
our adaptive choice of $k$ yields the fastest runtime performance
as well.

%% file: consistency.tex
\section{Consistency of Kernel Estimator}
\label{sec:consistency}

In this section we study the consistency of the estimator $\tilde{\g}$
defined in Eq.~\eqref{eq:hatf2}.  Recall that our model is:
%\begin{narrow}{-.3in}{-.15in}
\begin{align}\hspace{-1.5em}
%\y_{T+1}(i,j) = \ber(\g\bb{\st_{G_T}(i,j),\st(\n^T_i)})
\dcY | \mathcal{G} \sim \mbox{Bernoulli}(g(\psi_t(i,j))),%\mbox{, }
\label{eq:Y}
\end{align}
%\end{narrow}
where $\dcfeat$ equals $\{\dcg, \dch\}$.
Assume that all graphs have $\nn$ nodes (\nn is finite).
%the number of nodes in a graph at all timesteps is $\nn$.
For a fixed node $q\in\{1,\dots,n\}$, let $Q$ represent the query datacube $d_T(q)$.
We want to study the consistency of predictions for timestep $T+1$.

%Denote by
%{\small
%\begin{align*}
%\tilde{h}\sub{\nt}(s,\dchq):=\frac{\sum\limits_{i,t} \dcKone\left(\dch, Q\right)\cdot
%\left(\dcnplusBASE{i}{t+1}{\dcg}+\bti\sum\limits_{s\in n(\dch)}\eta^+_{i,t+1}(s)\right)}{\nn(T-p)(1+\bti|n(\dch)|)}
%\end{align*}
%%}
%%{\small
%\begin{align*}
%\tilde{f}\sub{\nt}(s,\dchq):=\frac{\sum\limits_{i,t} \dcKone\left(\dch, Q\right)\cdot
%\left(\dcnBASE{i}{t+1}{\dcg}+\bti\sum\limits_{s\in n(\dch)}\eta_{i,t+1}(s)\right)}{\nn(T-p)(1+\bti|n(\dch)|)}
%\end{align*}
%}
%and
%\begin{align*}
%\tilde{g}\sub{\nt}(s,\dchq)=\frac{\tilde{h}\sub{\nt}(s,\dchq)}{\tilde{f}\sub{\nt}(s,\dchq)}
%\end{align*}

Rather than studying $\tilde{\g}$ directly, it proves to be simpler to study
a slightly different estimator which we show (in Lemma~\ref{lem:est-g-dom})
to be asymptotically equivalent to $\tilde{\g}$.  Define $\est{g}(s,\dchq),
\est{\h}(s, \dchq)$ and $\est{\f}(s, \dchq)$ as follows:
\begin{align}
\dchatf(s, \dchq) = & \frac{\est{h}(s,
\dchq)}{\est{f}(s, \dchq)}\qquad\mbox{(where $s=\dcgq$)}\label{eq:est-def}\\
\est{\h}(s, \dchq) =&
\frac{1}{\nn(T-p)}\sum\limits_{t=p}^{T-1}\sum\limits_{i=1}^\nn
\Kb(\dch,\dchq)\dcnpluso\nonumber\\
\est{\f}(s, \dchq) = &
\frac{1}{\nn(T-p)}\sum\limits_{t=p}^{T-1}\sum\limits_{i=1}^\nn
\Kb(\dch,\dchq)\dcno. \nonumber
\end{align}

\begin{lem}
\label{lem:est-g-dom}
%\textcolor{red}{Deepay: CHECK this lemma.}
Define $\tilde{g}\sub{\nt}(.)$ as in Equation~\ref{eq:first-definition}, and $\dchatf(.)$ as in Equation~\ref{eq:est-def}. We have:
$$|\tilde{g}\sub{\nt}(s,Q)-\est{g}(s,Q)|=O(\bti)$$
\end{lem}
\begin{proof}
Recall that $n(s)$ denotes the set of features at $L_1$ distance
$1$ from $s$.
Let $k:=|n(s)|$.
We have:
%{\small
\begin{align*}
\tilde{g}\sub{\nt}(s,Q)&=\frac{\est{\h}(s, \dchq)+C_\nt}{\est{\f}(s, \dchq)+D_\nt},
\end{align*}
%}
where by virtue of the finiteness of number of features, $\eta$ and $\eta^+$, we have:
%{\small
\begin{align*}
C_\nt:=\bti\sum_{i,t}\Kb(\dch,\dchq)\sum\limits_{s'\in n(s)}\!\!\!\!\!\eta^+_{it+1}(s')=O(\bti).
\end{align*}
%}
Similarly, $D_\nt=O(\bti)$.
Also, note that both $C_\nt$ and $D_\nt$ are non-negative. Thus we have:
%{\small
\begin{align*}
\left|\tilde{g}\sub{\nt}(s,Q)-\hat{g}\sub{\nt}(s,Q)\right|&=\left|\frac{C_\nt\est{\f}(s,Q)-D_\nt\est{\h}(s,Q)}{(\est{\f}(s,Q)+D_\nt)\est{\f}(s,Q)}\right|=O(\bti),
\end{align*}
where the last step follows because both $\est{\h}$ and $\est{\f}$ are
bounded and $\est{\f}$ tends to some positive constant with
probability tending to one as $\nt\rightarrow\infty$ (as shown in
Theorem~\ref{thm:consistency}).
%}
%The result follows because $\est{g}(s,Q)$ is bounded.
\end{proof}

The estimator \dchatf is defined only when $\est{\f}>0$, which holds with probability
tending to one as will be shown in the next theorem. The kernel
was defined earlier as $\Kb(\dch,\dchq)=e^{-\dcD(\dch,\dchq)/\bt}$, where
the bandwidth $\bt$ tends to $0$ as $T\rightarrow\infty$, and $\dcD(\cdot)$
is the distance function defined in Eq.~\eqref{eq:tv}.  This has %is similar
%to other discrete kernels~\cite{Aitchison_Aitken}, and has
the following property:
%\footnote{This property follows from the finiteness of the graphs, and hence finiteness of possible datacubes.}:
%\footnote{This
%can be shown from our definition of $\dcD(.)$ and
%assumption (A1) discussed later.}:
%{\small
%\vspace{-1em}
%{\small
\begin{align}
\lim_{b_T\rightarrow 0}\Kb(\dch,\dchq)=\begin{cases}
1 & \text{if $\dch=\dchq$}\\
0 & \text{otherwise}.
\end{cases}
\label{eq:bandwidth}
\end{align}
%}
%\vspace{-1em}

From now on, we will drop the arguments $s$ and $\dchq$ and instead write $\g$, $\est{\g}$, $\est{\f}$ and $\est{\h}$ for simplicity.
Our graph evolution model is Markovian; assuming each ``state'' to represent $p+1$ consecutive graphs, the next graph (and hence the next
state) is a function only of the current state. The state space is
also finite, since each graph has bounded size. Thus, the state space $\mathcal{S}$
may be partitioned into a set of transient states and $\bigcup_i C_i$, where $C_i$ is an irreducible closed
communication class, and there exists at least one $C_i$~\citep{grimmett}.

%Assume that there is only one aperiodic communication class.
The Markov chain must eventually enter one of the (finitely many) communication classes. We will denote the time of entering some communication class by  $T_1$, and the event by $\mathcal{E}_{T_1}$.   We remind the reader that using simple arguments for finite state space Markov chains, it can be shown that the tail probability of $T_1$ decays geometrically (see \citep{grimmett}), leading to the finiteness of the first and second moments.
Also let \ec denote the event $S_\nt\in C$, where $S_t$ denotes the state of the Markov chain at time $t$.
Thus $\et\bigcap\ec$ is the event that the chain enters class $C$ at time $T_1$ and remains there henceforth.
%\vspace{-em}
\begin{thm}[Consistency]
\label{thm:consistency}
%If $E[\est{\f}]>0$, then
Let $\bt=o(1)$ as $T\rightarrow\infty$. For two fixed nodes $q,q'\in\{1,\dots,n\}$, $\est{\g}(s(q,q'),d_T(q))$ is well-defined with probability tending to one as $T\rightarrow\infty$. Also, $\est{\g}(s(q,q'),d_T(q))$ is a consistent estimator of $\g(s(q,q'),d_T(q))$, i.e., $\est{\g}(s(q,q'),d_T(q))\convprob\g(s(q,q'),d_T(q))$ as $T\rightarrow\infty$.
%\vspace{-1em}
\end{thm}
\begin{proof}
First, note that our query datacube is obtained at time $T$, and we are interested in the asymptotic behavior of the chain as $T\rightarrow\infty$. Since our Markov chain has a finite state space, the query datacube belongs to some closed communication class $C$ with probability tending to one. Thus, as $T\rightarrow\infty$, the estimator's distribution is governed by that communication class. We prove our result in two parts; first we show that the convergence statement holds conditioned on \ec, for any communication class $C$; i.e., $P(|\est{\g}-\g|\geq \epsilon|\ec) \rightarrow 0$ as $T\rightarrow\infty$. Next, we have
%{\small
\begin{align*}
&P(|\est{\g}-\g|\geq \epsilon)\leq \sum_C P(|\est{\g}-\g|\geq \epsilon|\ec)P(\ec)+P(T_1>\nt),
\end{align*}
%}
which implies $\limsup\limits_{\nt\rightarrow\infty}P(|\est{\g}-\g|\geq \epsilon) =  0$,
given the tail bound on $T_1$ and the fact that the first term is a sum over a finite
number of terms, each converging to zero as $T\rightarrow\infty$.
In what follows, we will give a proof of statistical consistency
conditioned on \ec for any communication class $C$.

%
%The proof is in two parts. First we prove that $\est{\g}\convprob
%\g$ as $T\rightarrow\infty$, conditioned on \ec. Since we have a finite number of communication classes, averaging over the probability of failing to converge for each class also converges to zero as $T\rightarrow \infty$, thus proving the result unconditionally.

Define $\Bt(s,\dchq,C)=E[\est{\h}|\ec]/E[\est{\f}|\ec]-g$. We have:
%{\small
\begin{align}
\label{eq:est-g-bias}
%&\mbox{For communication class $C$, such that $P(S_T\in C)\stackrel{T\rightarrow\infty}{\rightarrow} 1$}\\
& \est{\g}-g=\left.([\est{\h}-g\est{\f}]-E[\est{\h}-g\est{\f}|\ec])\right/\est{\f}+\Bt E[\est{\f}|\ec]/\est{\f}.
\end{align}
%}
Lemma~\ref{lem:ourexp} shows that $E[\est{\f}|\ec]\rightarrow R_c$, $R_c$ being a positive deterministic function of class $C$. Thus, \Bt is asymptotically
well defined. Also Lemma~\ref{lem:ourvar} shows that $\var(\est{\f}|\ec)$ tends to $0$ as $T\rightarrow \infty$. This along with Lemma~\ref{lem:ourexp}  shows that, conditioned on \ec, $\est{\f}\stackrel{P}{\rightarrow} R_c $, thus also proving that $\est{\g}$ is asymptotically well defined for $C$.

Next, we will define the following:
%{\small
\begin{eqnarray}
\est{\h}(t)&:=&\frac{1}{\nn}\sum\limits_{i=1}^\nn
\Kb(\dch,\dchq)\dcnpluso,\nonumber\\
\est{\f}(t)&:=&\frac{1}{\nn}\sum\limits_{i=1}^\nn \Kb(\dch,\dchq)\dcno\label{eq:htft}.
\end{eqnarray}
%}
Note that $\est{\h}$ and $\est{\f}$ (Equation~\ref{eq:est-def}) equals $\sum_t \est{\h}(t)/(\nt-p)$ and $\sum_t \est{\f}(t)/(T-p)$ respectively.
 Also let
% {\small
 \begin{align}\label{eq:qt}
 q_t:=\est{\h}(t)-E[\est{\h}(t)|\ec]-\g(\est{\f}(t)-E[\est{\f}(t)|\ec]).
 \end{align}
% }
 Thus $q_t$ is a bounded deterministic function of the state at time $t$.  In Lemma~\ref{lem:ourvar} we prove that $\var(\sum_t q_t/\sqrt{T}|\ec)\rightarrow \sigma_c$ for some non-negative constant $\sigma_c$, as $T\rightarrow\infty$. Thus we have, $\var(\sum_t q_t/T|\ec)\rightarrow 0$, as $T\rightarrow \infty$. Since $E[q_t|\ec]=0$, we have $\sum_t q_t/T\sim([\est{\h}-\g\est{\f}]-E[\est{\h}-\g\est{\f}|\ec])\stackrel{qm}{\rightarrow}0$ conditioned on \ec.

Since convergence in quadratic mean implies convergence in probability, we have:
%{\small
\begin{equation*}
(\est{\f},[\est{\h}-\g\est{\f}]-E[\est{\h}-\g\est{\f}|\ec])\stackrel{P}{\rightarrow} (R_c,0) \quad\mbox{conditioned on \ec}.
\end{equation*}
%}
Using the continuous mapping theorem on $f(X,Y)=Y/X$ and the fact that
$\Bt=o(1)$ (Lemma~\ref{lem:bias}) we have that, for any $C$ such that
 $S_T\in C$, $\est{\g}\stackrel{P}{\rightarrow} g$.
\end{proof}

\input{conv_expectation}
The following smoothness condition on $g$ is introduced to ensure appropriate rates of convergence of the bias terms $\Bt$.

\begin{assumption}
\label{assumption:smooth}
The function $g$ satisfies the following smoothness condition with respect
to the distance metric $D$: $|g(s,\dch)-g(s,d_{t'}(j))|=O(D(\dch,d_{t'}(j)))$.
\end{assumption}
\begin{lem}
\label{lem:bias}
Define {\small$\Bt(s,Q,C)=(E[\est{\h}(s,Q)|\ec]-gE[\est{\f}(s,Q)|\ec])/E[\est{\f}(s,Q)|\ec]$}. If Assumption~\ref{assumption:smooth} holds, then we have $\Bt=O(\bt)$. Since $\bt\rightarrow 0$ as $T\rightarrow \infty$, this implies $\Bt=o(1)$.
\end{lem}
\begin{proof}[Proof Sketch]
For $t\in [p,T-2]$, $i\in [1,N]$ and $s=\dcgq$,
the numerator of $\Bt$ is an average of the terms:
%{\small
\begin{align*}
A_t:=E\left[\Kb(\dch,\dchq)\dcnpluso|\ec\right]-E\left[\Kb(\dch,\dchq)\dcno|\ec\right]\g(s,\dchq).
%E\left[\Kb(\dch,\dchq)\dcnpluso\right]-E\left[\Kb(\dch,\dchq)\dcno\right]\g(s,\dchq)\\\text{for
%} t\in p:T-2;\ \ i\in 1:N; s=\dcgq
\end{align*}
%$$E[\Kb(\n^t_i,\n_q)\np{t+1}{ij}]-E[\Kb(\n^t_i,\n_q)\nt{t+1}{ij}]\g(\n_q,\st_q)\ \ \text{for } t\in 1:T-2,\ \ i\in 1:N$$
%}
%Taking expectations w.r.t. \dch, and denoting
%$\Kb(\dch,\dchq)$ by $\gamma$, the first term becomes:
%\begin{eqnarray}
%E\left[\gamma \dcnpluso|\ec\right] & = &
%E\left[\gamma E\left[\dcnpluso | \dch,\ec\right]|\ec\right]
%  =  E\left[\gamma \dcno\g(s, \dch)|\ec\right]\nonumber
%%\nonumber\\
%% & = & E\left[\Kb(\dch,\dchq) \dcno\cdot \g(s, \dch)\right]\nonumber
%\end{eqnarray}
%
%Now note that $E\left[\dcnpluso | \dch,\ec\right]=E[E[\dcnpluso|\dch,\et,\ec]|\ec]$. Conditioning on $\et$ makes $\dcnpluso$ conditionally independent of
%\ec given $\dch$ if $t>T_1$. Also, for $t\geq T_1$, $E\left[\dcnpluso | \dch,\et,\ec\right] =\dcno\cdot \g(s, \dch)$, as
%can be seen by summing Eq.~\ref{eq:Y} over all pairs $(i, j)$ in a
%neighborhood with identical $\dcg$, and then taking expectations\footnote{Note that the conditioning on \et is crucial here.}.
%This along with the fact that $\gamma\dcnpluso$ is bounded leads to:
%\begin{align*}
%E[\dcnpluso|\dch,\et,\ec]&\leq \dcno g(s,\dch) \ind[T_1\leq t] + c\ind[T_1> t] \\
%&\leq \dcno g(s,\dch)+c\ind[T_1> t]
%\end{align*}
Using a further conditioning step on \et, we can show that the numerator of \Bt can be upper bounded as:
%{\small
\begin{align*}
 |\sum_t A_t/T| &\leq \sum_t  |E[\Kb(\dch,\dchq) \dcno (g(s,\dch)-g(s,\dchq))|\ec]|/T+o(1).
\end{align*}
%}
We now analyze each term in the average; i.e., terms of the form:
%{\small
\begin{align*}
E\left[\Kb(\dch,\dchq) \dcno\cdot \left(\g(s, \dch) - \g(s,
\dchq)\right)|\ec\right].
\end{align*}
%}
%}
This expectation is computed over all possible configurations of the
neighborhoods $\dcN{t}$ and $\dcN{t+1}$.
Since our neighborhood sizes are bounded (because $n$ is bounded),  the expectation
is a sum over a finite number of terms.

We now use the smoothness assumption on $\g$. Using $\left|\g(s, \dch) - \g(s,
\dchq)\right|=O(\dcD(\dch,\dchq))$ and that $\dcno$ is finite for all $T$ and Lemma~\ref{lem:ourexp},  we have:
%{\small
\begin{align*}
\Bt= O \left(E[\dcD(\dch,\dchq)e^{-\dcD(\dch,\dchq)/\bt}|\ec]\right) =O(\bt),
\end{align*}
%}
which holds because for non-negative $x$, we have $xe^{-x/\bt}\leq \bt/e$.
%\vspace{-1em}
%We now let $b\rightarrow 0$. Since the expectation is over a finite
%sum, so we can take the limit inside the expectation.
%Since $\lim_{b\rightarrow 0} \Kb(\dch, \dchq)$ is zero unless
%$\dch=\dchq$ (Eq.~\ref{eq:bandwidth}),
%%. For any $i,t$ the expectation is over a finite number of terms, and so we can exchange limit and expectation.
%we find that
%$E\left[\Kb(\dch,\dchq) \dcno\cdot \left(\g(s, \dch) - \g(s,
%\dchq)\right)\right]\rightarrow 0$, for any $i$ and $t$.  So the numerator of $B$
%goes to zero asymptotically. Hence, the estimator $\est\g$ is
%unbiased.
\end{proof}
\input{variance}

%% file: conv_expectation.tex
The proof of the following lemma is deferred to the Appendix.

\begin{lem}
%Denote  $\eta$ value of the query datacube $\dchq$ for feature $s$ by $\eq$.
As $T\rightarrow \infty$, for some $R_c>0$ (a deterministic function of class $C$),
\vspace{-.1in}
\begin{align*}
%E[\est{h}(s,\dchq)] \rightarrow g(s,\dchq) R_c, \quad
E[\est{f}(s,\dchq)|\et,\ec] \rightarrow R_c & \mbox{,} & E[\est{f}(s,\dchq)|\ec] \rightarrow R_c.
\end{align*}
\label{lem:ourexp}
\end{lem}
\vspace{-2em}

%% file: variance.tex
\reminder{Assumption of finiteness of graphs}
%Define the strong
%mixing coefficients $\alpha(k)\doteq \sup_{\dch,\dchp}
%\{|P(A\cap B)-P(A)P(B)|:A\in \mathcal{F}(\dch), B\in\mathcal{F}(\dchp), \dist(\{i,t-p+1,\ldots,t+1\},\{i',t'-p+1,\ldots,t'+1\})\geq k\}$,
%where $\mathcal{F}(\dch)$ is the
%sigma algebra generated by the random variables in the datacube for
%$i,t$.
We now show that the variance of $\est{\f}$ and $\est{\h}$ converge to zero.
In order to upper bound the growth of variance terms, we make use of strong
mixing.  For a Markov chain
$S_{t}$, define the strong mixing coefficients $\alpha(k)\doteq
\sup_{|t-t'|\geq k} \{|P(A\cap B)-P(A)P(B)|:A\in \mathcal{F}_{\leq t},
B\in\mathcal{F}_{\geq t'}\}$, where $\mathcal{F}_{\leq t}$ and
$\mathcal{F}_{\geq t'}$ are the sigma algebras generated by events in
$\bigcup_{i\leq t}S_i$ and $\bigcup_{i\geq t'}S_i$ respectively.
Intuitively, small values of $\alpha(k)$ imply that states that are
$k$ apart in the Markov chain are almost independent. For bounded $A$
and $B$, this also limits their covariance: $|\cov(A,B)| \leq c
\alpha(k)$ for some constant $c$~\citep{durrett}. Instead of proving that the variance of $\est{\h}$ or $\est{\f}$ converges to zero, we will prove that the variance divided by \nt converges to a non-negative constant. This is a stronger result that we will find useful in proving weak convergence in section~\ref{sec:dist-conv}.

We introduce some notation that will be used in stating the next few results.
Let $q_t$ denote a bounded deterministic function of the state of a finite state space Markov chain at time $t$.  Also define $U_{\nt}:=\sum_t q_t/\sqrt{T}$.
Recall that our Markov chain will eventually hit one of the finitely many closed communication classes. Earlier we used \ec to define the event $\{S_\nt\in C\}$, by $T_1$ the time of entering some communication class, and the event by \et.
%Let us denote the time of entering some communication class by  $T_1$, and the event by $\mathcal{E}_{T_1}$.
We will denote the event of entering class $C$ at time $T_1$ by $\mathcal{E}_{T_1}\bigcap\ec$. If $C$ is aperiodic, then once inside $C$, the Markov chain gets arbitrarily close to the stationary distribution of $C$ after some constant time \mc; we state this more formally in the following lemma, whose proof is deferred to the Appendix.% (Lemma~\ref{lem:diff-exp}).

\begin{lem}
\label{lem:diff-exp}
Consider an irreducible and aperiodic finite state Markov chain with probability transition matrix $P$, initial distribution $\pi_0$ and stationary distribution $\pi$.
Let $X_t$ be a random variable (with finite support) that is conditionally independent of all other states, given the state at time $t$. The expectation of $X_t$ under the distribution at time $t$ is denoted by $E[X_t|\pi_0]$. Let $\mu$ denote the expectation of $X_\infty$
(i.e., the expectation with respect to $\pi$).
 There exists a constant $\lambda\in(0,1)$, and a constant $M$ such that, for all $t>M$, $\max_{x\in \mathcal{S}} \sum\limits_{y\in \mathcal{S}}|P(x,y)-\pi(y)|=O(\lambda^t)$, and
 $|E[X_t|\pi_0]-\mu|=O(\lambda^t)$.
\end{lem}

Our estimators are weighted sums of $1,\dots,\nt$ variables; for
$T_1\leq T$,  we will break this sum up into three parts, indexed by
$1,\dots,T_1-1$, followed by $T_1,\dots,T_1+\mc-1$, and finally $T_1+\mc,\dots,T$, where $\mc$ is a constant. For $T_1>T$, we will use the fact that $T_1$ has bounded first and second moments. Since we are interested in the behavior of the sum unconditionally, our analysis will consist of two steps of nested conditioning, the outer one obtained by conditioning on \ec, which in turn is obtained by analyzing the sum conditioned on $\et\bigcap\ec$.
%First we will prove results on asymptotic behavior of the variance conditioned on $\mathcal{C}$.
For ease of exposition we will assume $C$ to be aperiodic. The more general case of cyclo-stationarity, which is similar in principle, is discussed in remark~\ref{remark:variance-cyclo}.

%\begin{proposition}
%\label{lem:qm-conv-uncond}
%%Let $q_t$ denote a bounded deterministic function of the state of a finite state space Markov chain at time $t$.  Also define $U_\nt:=\sum_t q_t/\sqrt{T}$.
%%Let \qit be a bounded function of $\dcno$,$\dcnpluso$ and $\dch$.
%Let $U_\nt$ be defined as above. $\var(U_\nt)
%\rightarrow \sigma$ as $T \rightarrow \infty$, where $\sigma$ is a non-negative constant.
%\end{proposition}
%\begin{proof}
%Using lemma~\ref{lem:qm-conv} we see that
%\end{proof}

\begin{lem}
\label{lem:qm-conv}
%Let $q_t$ denote a bounded deterministic function of the state of a finite state space Markov chain at time $t$.  Also define $U_\nt:=\sum_t q_t/\sqrt{T}$.
%Let \qit be a bounded function of $\dcno$,$\dcnpluso$ and $\dch$.
%Let $U_\nt$ and $\mathcal{C}$ be defined as above.
$\var(U_\nt|\ec)
\rightarrow \sigma_c$ as $T \rightarrow \infty$, for some constant $\sigma_c\geq 0$.
\end{lem}
\begin{proof}
We have $\var(U_\nt|\ec)=E[\var(U_\nt|\et,\ec)|\ec]+\var(E[U_\nt|\et,\ec]|\ec)$.
%The variance of $U_\nt$ conditioned on \ec can be broken into two parts, one being the expectation of the conditional variance , and the other being the variance of the conditional expectation (both conditioned on \et).
We prove that the first part converges to a non-negative constant $\sigma_c$ (a deterministic function of $C$) (Lemma ~\ref{lem:cond-var}), and the second is asymptotically $o(1)$ (Lemma~\ref{lem:cond-exp}).
%First assume that this class is aperiodic.
\end{proof}

\begin{lem}
\label{lem:cond-var}
 For any finite integer $k$, we have
 \begin{align}
 \var(\sum_{t\geq T_1+\mc} q_t|\mathcal{E}_{T_1},T_1=k,\ec)/T&\rightarrow\sigma_c\qquad\mbox{for some $\sigma_c\geq 0$}\label{eq-var-endpart}\\
 \var(\sum_t q_t|\mathcal{E}_{T_1},T_1=k,\ec)/T&\rightarrow\sigma_c\qquad\mbox{for some $\sigma_c\geq 0$}. \label{eq-var-tot}
 \end{align}
%$\var(U_\nt|\mathcal{E}_{T_1})=O(T_1^2/T)+\dfrac{1}{T}\sum\limits_{\ell=0}^{d-1}\sum\limits_{r=0}^{R_t-1}c_{T_1+rd+\ell}$.
For a Markov chain with a finite state space, we also have $E[\var(U_\nt|\mathcal{E}_{T_1},\ec)|\ec]\rightarrow \sigma_c$ for some $\sigma_c\geq 0$.
\end{lem}
\begin{proof}[Proof Sketch]
For ease of exposition, for the proof sketch we assume there is only one communication class, which is aperiodic.
%The proof for a cyclic communication class, while similar in principle, is more involved and is deferred to the Appendix.
Recall that $T_1$ is the time to hit the communication class. Once inside the communication class,
irreducibility and aperiodicity implies geometric
ergodicity (Lemma~\ref{lem:diff-exp}), which implies absolute regularity which in turn implies strong mixing with
exponential decay~\citep{Bradley2005}:
$\alpha(k)\sim e^{-\beta k}$ for some $\beta>0$. %Let us define $q_t:=1/n\sum_i \qit$; thus we are interested in $\var(\sum_t q_t|\et,\ec)/T$.
We can prove that for finite $T_1$, $\var(\sum_t q_t|\mathcal{E}_{T_1},T_1=k,\ec)/T=\var(\sum_{t\geq T_1+\mc} q_t|\mathcal{E}_{T_1},T_1=k,\ec)/T+o(1)$.
So we focus on proving Equation~\ref{eq-var-endpart}. Denote $\sum_{t\geq T_1+\mc} q_t$ by $P$.
%
%
%This variance can be broken into three parts, $\var(\sum\limits_{t<T_1}q_t|\et,\ec)$, $\var(\sum\limits_{t\geq T_1}q_t|\et,\ec)$ and finally $\cov(\sum\limits_{t<T_1}q_t,\sum\limits_{t'\geq T_1}q_{t'}|\et,\ec)/T$. The first part is $O(T_1^2/T)$.
%
%
% For the second part, we will break up the sum into two parts; denote $P_1:=\sum\limits_{T_1\leq t< T_1+\mc}q_t$, and $P_2:=\sum\limits_{t\geq T_1+\mc}q_t$.
% $$\var(\sum\limits_{t\geq T_1}q_t|\et,\ec)=\var(P_1|\et,\ec)+\var(P_2|\et,\ec)+\cov(P_1,P_2|\et,\ec)$$
%We have, $\var(P_1|\et,\ec)=O(\mc^2)$.

Recall that for our Markov chain, $S_t$ involves $p+1$ graphs ($G_{t-p+1},\dots,G_{t+1}$). Since $p_t$ is a function of $S_t$, it also depends on $p+1$ graphs. Hence, the distance $\dist(t,t')$ between two sigma-algebras $\mathcal{F}_{\leq t}$ and $\mathcal{F}_{> t'}$ is defined as $\max(t'-t-(p+1),0)$.
Now we can write $\var(P|\et,\ec)$ as
$$\var(P|\et,\ec)=2\sum\limits_{t\geq T_1+\mc} \sum\limits_{\dist(t,t')=0}^{T-t} \cov(q_t, q_{t'}|\et,\ec).$$  Since the number of states at distance $0$ is $O(p+1)$, and at distance $\geq 1$ is $O(1)$, for constants $\{c_k, k\geq 0\}$ we have,
%{\small
 \begin{align*}
 \sum_{\dist(t,t')=0}^{T-t} |\cov(q_t, q_{t'}|\et,\ec)|\leq \sum_{k=0}^{\infty} c_k \alpha(k)= O(\sum_k e^{-\beta k}) =O(1).
 \end{align*}
 %}
 %Thus for $t,t'\geq T_1$, conditioned on $\et,\ec$, $\sum_{t\geq T_1} \sum_{|t-t'|=0}^{T} |\cov(q_t, q_{t'}|\et,\ec)|$  is upper bounded by $\sum_{t\geq T_1}\sum_{k=0}^{\infty} c\alpha(k)\leq \sum_k ce^{-\beta k}$, which in turn is upper bounded by some positive constant.
 This shows that the above sum converges to some constant $a_t$. %$\sum\limits_{t'=t}^{T} \cov(\qit, \qjtp|\et,\ec)\rightarrow a_t$, where $a_t$ is some constant.
Since $t\geq T_1+\mc$, the chain will get arbitrarily close to stationarity, and $a_t\rightarrow \sigma_c$ for some constant $\sigma_c$. Thus $\var(P|\et,\ec)/T$ is asymptotically equivalent to $\sum_t a_t/T$, which also converges to $\sigma_c$ as $T\rightarrow\infty$. Since, for all $T$, $\var(P|\et,\ec)/T$ is non-negative, $\sigma_c$ is also non-negative. This proves Equation~\ref{eq-var-endpart}.
%Using the Cauchy Schwartz inequality, $$\cov(P_1,P_2|\et,\ec)/T=O(\sqrt{(\var(P_1|\et,\ec)/T)(\var(P_2|\et,\ec)/T)})=o(1).$$
%Thus the second part converges to $\sigma_c$ as $T\rightarrow\infty$ for some non-negative constant $\sigma_c$.
%
%Another use of the Cauchy Schwartz argument from before, %We also can upper bound the third part by $$\sqrt{\left.\var(\sum_{t<T_1}q_t|\et,\ec)\right/T\left.\var(\sum_{t\geq T_1}q_t|\et,\ec)\right/T}$$.
%along with the convergence result on the second part lets us upper bound the third part by $O(T_1/\sqrt{T})$.
%
Thus Equation~\ref{eq-var-tot} is proved, and also, since $T_1$ has finite first and second moments for a finite state space Markov chain,  $E[\var(\sum_t q_t|\et,\ec)|\ec]/T$ converges to $\sigma_c$, as $\nt\rightarrow\infty$.
\end{proof}

It remains to analyze $\var(E[U_\nt|\et,\ec]|\ec)$ in the variance decomposition.
Using Lemma~\ref{lem:diff-exp}
we can prove that $|E[U_\nt-\mu_c|\ec,\et]|$ approaches zero at a geometric rate as \nt$\rightarrow\infty$, where $\mu_c$ denotes the
expectation of $q_t$ under the stationary distribution in communication class $C$. This implies the following lemma, which is proved in the Appendix.
\begin{lem}
\label{lem:cond-exp}
%Let $R_t = \lfloor (T-t)/d \rfloor$, and also let $E_{t}$ denote the event $\{S_t\in C\}$.
$\var(E[U_\nt|\et,\ec]|\ec)=o(1)$.
\end{lem}

\begin{lem}
 $\var(\est{\h}|\ec)$ and $\var(\est{\f}|\ec)$
tend to $0$ as $T\rightarrow\infty$.
\label{lem:ourvar}
\end{lem}
%\end{narrow}
%\vspace{-2em}
\begin{proof}
The result follows by applying Lemma~\ref{lem:qm-conv} with $q_t(.)$ equal to\\
$\sum_i \Kb(\dch,\dchq)\dcnpluso/n$ and
$\sum_i\Kb(\dch,\dchq)\dcno/n$ respectively.
\end{proof}

\begin{remark}
\label{remark:variance-cyclo}
Recall that Lemma~\ref{lem:cond-var} was obtained under the assumption that $C$ is aperiodic. The case of periodic $C$ implies cyclo-stationarity; i.e., the chain $S_{t+kd}$ approaches stationarity as $k\rightarrow\infty$. Hence, for periodic $C$ (with period $d$) we consider $\mathcal{M}'$, which is a Markov chain where each transition corresponds to $d$ transitions of the original chain.
Now, $\mathcal{M}'$ is irreducible and aperiodic (since $C$ was irreducible and had period $d$). A state $S'_{t}$ in $\mathcal{M}'$ started at $S_1$ simply corresponds to the old state $S_{td+1}$ in $\mathcal{M}$.
Now, $1/\sqrt{\nt}\sum_{t=1}^\nt q_t$ can be written as $1/\sqrt{\nt}\sum_{t=1}^{\lfloor\nt/d\rfloor}q'_t+o_P(1)$, where $q'_i:=\sum\limits_{j=id+1}^{(i+1)d} q_j$ is the sum of $d$ consecutive random variables.
Since, $q'_t$ is independent of all other $q'$s conditioned on $S'_t,S'_{t+1}$, we have: %$\forall t_1>t+1$,  $q'_t\ci q'_{t_1}| S_{t+1}$.%, and $\forall t_1<t$  $ q'_t\ci q'_{t_1}| \mathcal{F}_{\geq t}$.
%$q'_t$ is independent of all other $q'$'s given $S_t$ and $S_{t+d}$.
%Hence, for $q'_i:=\sum\limits_{j=id+1}^{(i+1)d} q_i$,
%Hence we have the following:
% \begin{align}
%\cov(q'_t,q'_{t+k}) %\cov(q_t,q_{t+kd})=E[q_tq_{t+kd}]-E[q_t]E[q_{t+kd}]\\
%&=E[E[q'_tq'_{t+k}|\mathcal{F'}_{\leq t+1},\mathcal{F}'_{> t+k}]]-E[q'_t|\mathcal{F}'_{\leq t+1}]E[q'_{t+k}|\mathcal{F}'_{> t+k}]\label{eq-cyclo-cov}\\
%&=E[E[q'_t|\mathcal{F}'_{\leq t+1}]E[q'_{t+kd}|\mathcal{F}'_{> t+k}]]-E[q'_t|\mathcal{F}'_{\leq t+1}]E[q'_{t+kd}|\mathcal{F}'_{> t+k}]\nonumber\\
%%&=E[f(\mathcal{F}'_{\leq (t+1)d}),g(\mathcal{F}'_{\geq (t+k)d})]-E[f(S_t,S_{t+d}),g(\mathcal{F}'_{\geq (t+k)d})]\\
%&=\cov(E[q'_t|\mathcal{F}'_{\leq t+1}],E[q'_{t+kd}|\mathcal{F}'_{> t+k}]) \nonumber\\
%&= O(\alpha(k-1))\nonumber
% \end{align}
 \begin{align}
&\cov(q'_t,q'_{t+k}) %\cov(q_t,q_{t+kd})=E[q_tq_{t+kd}]-E[q_t]E[q_{t+kd}]\\
=E[E[q'_tq'_{t+k}|S'_{t+1},S'_{t+k}]]-E[q'_t]E[q'_{t+k}]\label{eq-cyclo-cov}\\
&=E[E[q'_t|S'_{t+1}] E[q'_{t+k}|S'_{t+k}]]-E[E[q'_t|S'_{t+1}]]E[E[q'_{t+k}|S'_{t+k}]]\nonumber\\
%&=E[f(S'_{t+1}) g(S'_{t+k})]-E[f(S'_{t+1})]E[ g(S'_{t+k})]\nonumber\\
%&=E[E[q'_t f(S'_{t+k})|S'_{t+k}] ]-E[q'_t]E[f(S'_{t+k})]\nonumber\\
%&=E[q'_t f(S'_{t+k})]-E[q'_t]E[f(S'_{t+k})]\nonumber\\
%&=E[f(\mathcal{F}'_{\leq (t+1)d}),g(\mathcal{F}'_{\geq (t+k)d})]-E[f(S_t,S_{t+d}),g(\mathcal{F}'_{\geq (t+k)d})]\\
&=\cov(E[q'_t|S'_{t+1}],E[q'_{t+k}|S'_{t+k}]) = O(\alpha(k-1)).\nonumber
 \end{align}
The last step uses the fact that the $q'_t$ are bounded. Now, $E[\var(1/\sqrt{\nt}\sum_{t=1}^{\lfloor\nt/d\rfloor}q'_t|\et,\ec)]$ can again be shown to converge to some non-negative constant using a slight modification of the argument in Lemma~\ref{lem:cond-var}. The $o_P(1)$ remainder of $1/\sqrt{T}\sum_{t=1}^T q_t$ can be shown to be negligible via a simple application of the Cauchy-Schwartz inequality. A detailed proof of Lemma~\ref{lem:cond-var} using this idea can be found in the Appendix.

As for $E[U_\nt|\et,\ec]$ in the cyclic case, we simply have to apply Lemma~\ref{lem:diff-exp} for each of the $d$ cyclic classes. For the $i^{th}$ cyclic class, $q_{T_1+kd+i}$ is independent of all states (in that cyclic class) given $S_{T_1+kd+i}$. Hence there exists $\mc_i$, and $\lambda_i\in (0,1)$, such that for all $k$ with $kd+i>\mc_i$, $|E[q_{T_1+kd+i}|\et,\ec]-\mu_i|=O(\lambda_i^{k})$, thus proving Lemma~\ref{lem:diff-exp} for a periodic $C$. This again proves Lemma~\ref{lem:cond-exp} for the case where $C$ is periodic.
\end{remark}

%% file: stein-general-v3.tex
\section{Stein's Method for Graphical Data}
\label{sec:stein-general-v3}
%\section{Sum of random variables with local dependence}
Our estimators, and indeed many kernel estimators, involve
weighted sums of dependent variables.  While their distributional
convergence can be studied using existing results on ergodic Markov chains,
we take a different approach, based on an adaptation of Stein's
method to the setting of graphs.

We begin with a brief introduction to Stein's method.  The method
reposes on the following key lemma~\citep{chen2010normal}, which
provides a characterization of the normal distribution:
\begin{lem}[Stein's Lemma]
\label{lem:stein-orig}
 If $W$ has a standard normal distribution, then
 {\small
 \begin{align}
 Ef'(W)=E[Wf(W)],
 \label{eq-stein-orig}
 \end{align}
 }
for all absolutely continuous functions $f:\mathbb{R}\rightarrow\mathbb{R}$ with $E|f'(Z)|<\infty$. Conversely, if Equation~\ref{eq-stein-orig} holds for all bounded, continuous and piecewise continuously differentiable functions $f$ with $E|f'(Z)|<\infty$, then $W$ has a standard normal distribution.
\end{lem}

Recall that the Wasserstein distance between a mean zero, unit variance random
variable $W$ and a standard normal variate $Z$ is defined as $\sup_{h\in\mathcal{H}}|E h(X)- E h(Z)|$, where $\mathcal{H}:=\{h: |h(x)-h(y)|\leq |x-y|\}$. Weak convergence of $W$ to $Z$ can
be established by showing that the Wasserstein distance converges to zero. Now,
Stein's Lemma~(\ref{lem:stein-orig}) shows that $W\stackrel{d}{=}Z$ if $|Ef'(W)-E[Wf(W)]|$ equals zero for appropriate choices of $f$. This key observation leads to the Stein Equation:
\begin{align}
f'(W)-Wf(W)=h(W)-E[h(Z)].
\end{align}
It can be shown that the solution to the Stein Equation, for $h\in\mathcal{H}$, satisfies  $\|f\|\leq 2$, $\|f'\|\leq 2$, $\|f''\|\leq \sqrt{2/\pi}$~\citep{chen2010normal}. Thus, instead of dealing with $E[h(W)]-E[h(Z)]$ we need to show that  $|E[f'(W)-Wf(W)]|$ is small (where $f$ satisfies the aforementioned conditions); this is an easier quantity to analyze.

The existing application of Stein's method to sums of weakly dependent random variables
has focused on marginal-independence structures that can be captured by a
bounded-degree dependency graph~\citep{Rinott:1996:MCL:230170.230181}.
In this section, we relax the requirement of marginal independence by
allowing arbitrary dependency structures among the summed variables
as long as certain conditions on strong mixing coefficients $\alpha(k)$ hold.
(See also \citet{SteinStrong07} for a similar approach to ours for chain-structured
dependencies; he obtains a slightly tighter bound than ours at the expense of
a more complex proof.)

Our approach proceeds by bounding the Wasserstein distance between the
(appropriately scaled and centered) sum $W$ of the dependent variables and
a standard normal variate $Z$ in terms of $\alpha(k)$ and the degree
of dependence of the random variables. We then show that this bound
tends to zero for our estimators, demonstrating convergence to a
normal distribution and yielding a rate of convergence as a by-product.
We note that although we use this to prove normal convergence for
a cyclo-stationary Markov chain, it can potentially be used for more
general dependence structures, as long as suitable strong mixing
properties are available.

%First we will present our result on
We let $\nt$ denote the total number of variables in our model.
Let $Y_i,\{i=1,\dots \nt\}$ be bounded, ($|Y_i|\leq B$), mean-zero
random variables. Let $\sigma\sub{\nt}^2$ denote the variance of $\sum\limits_i^\nt Y_i$; assume
$0<\sigma\sub{\nt}<\infty$ for all $\nt$. Define $X_i=Y_i/\sigma\sub{\nt}$, where $|X_i|\leq B/\sigma\sub{\nt}$.  Let
$W:=\sum\limits_i^\nt X_i$, and $\cn=\nt/\sigma\sub{\nt}$. %We want to show that $W$ converges in distribution to a standard normal.
%In order to upper bound the Wasserstein distance between $W$ and
%$Z\stackrel{d}{=}\mathcal{N}(0,1)$, we will bound $|E[f'(W)-Wf(W)]|$,
%for any twice differentiable function $f$ satisfying $\|f\|$, $\|f'\|\leq 2$, $\|f''\|\leq \sqrt{2/\pi}$.
We will assume that the index set underlying the random variables $\{X_i\}$
is endowed with a distance metric, $\dist(i,j)$.  This can be the geodesic
distance if the variables are connected via a graph structure or the absolute
difference in time indices in a time series model, etc.  Let $\natm(i)$ denote the set of nodes at distance $m$ from node $i$; similarly let $N_{\leq k}(i)$ and $\ngtk(i)$ respectively denote the set of nodes within distance $k$ and at a distance greater than $k$ from node $i$. Now, let $|N_{\leq k}|$
denote $\max\limits_{i}|\nltk(i)|$.
%Let $\ngtk(i)$ denote the set of nodes that are at a distance greater than $k$ away from $i$.

We need a notion of strong mixing in a network setting. Define the strong
mixing coefficients $\alpha(k)\doteq \sup_{X_i,X_j}\{|P(A\cap
B)-P(A)P(B)|:A\in \mathcal{F}(X_i), B\in\mathcal{F}(X_j),
\dist(i,j)\geq k\}$, where $\mathcal{F}(X)$ is the
sigma algebra generated by the random variable $X$. A similar proposal for strong mixing in random fields can be found in ~\cite{politis-subsampling-book}.
Let $\tau_k$ denote the tail sum $\sum\limits_{m>k}\satm\alpha(m)$. We are now ready to state the main result.

\begin{lem}
\label{lem-stein-main}
The Wasserstein distance $d_W(W,Z)$ between $W$ and the standard normal random
variable $Z$ is upper bounded as follows:
{\small
\begin{align}
d_W(W,Z) \leq & \min_{k\leq\nt}\left(\cone B^3\cn\left(\frac{\sltk}{\sigma\sub{\nt}}\right)^2+
\ctwo B\cn\alpha(k) +\right.\label{eq-stein-general}\\
& \left.B^2\sqrt{\cthree \left(\frac{\cn \tau_k}{\sigma\sub{\nt}}\right)^2+\cfour
\cn\left(\frac{\sltk}{\sigma\sub{\nt}}\right)^3+\cfive
\cn\frac{\tau_k}{\sigma\sub{\nt}}\left(\frac{\sltk}{\cn}\right)^2 }\right)\nonumber,
\end{align}
}
where $\cone,\ctwo,\cthree,\cfour,\cfive$ are constants.
\end{lem}
\begin{proof}[Proof sketch]
We will give a brief proof sketch here, and provide the full proof in the Appendix.
We want to bound $|E[f'(W)-Wf(W)]|$. We shall repeatedly break up $W$ into two parts: $W_i =
\sum\limits_{j\in\ngtk(i)}X_j$ being the contribution from all nodes with distance more than
$k$ from some node $i$, and the remainder from nodes ``close to''
$i$. In classical analysis of dependency graphs, $X_i$ and $W_i$ are independent;
in contrast, in our case we only have $\cov(X_i,W_i)=O(\alpha(k))$.
Here, $k$ is a parameter that shall be picked later to optimize the bound.
Since $W=\sum_{i=1}^\nt X_i$,
{\small
\begin{align*}
% \label{eq-stein-bound}
|E[f'(W)-Wf(W)]|%=|E[f'(W)-\sum\limits_i X_if(W)]|\nonumber\\
&\leq \underbrace{\left|E[f'(W)(1+\sum\limits_i X_i(W_i-W))]\right|}_{(A1)}\\
&+\underbrace{\left|E[\sum\limits_i X_i(W_i-W)f'(W)+\sum\limits_i X_i f(W)]\right|}_{(A2)}.
%&\leq E\left|f'(W)\left(1+\frac{\sum\limits_i X_i(W_i-W)}{\sigma\sub{\nt}}\right)\right|+\frac{1}{\sigma\sub{\nt}}E\left|\sum\limits_i X_i(W_i-W)f'(W)-\sum\limits_i X_i
\end{align*}
}
Using Taylor expansion the term $(A2)$ can be further bounded by
{\small
\begin{align*}
&(A2)
%&\leq E\left|\sum\limits_i X_i(W_i-W)f'(W)-\sum\limits_i X_i (f(W_i)-f(W))\right|+\left|E[\sum\limits_iX_if(W_i)]\right|\nonumber\\
%&\leq E\left|\sum\limits_i X_i(W_i-W)(f'(W)-f'(W_i))\right|+\frac{1}{2}E\left|\sum\limits_i X_i(W-W_i)^2f''(W_i^*)\right|+\left|E[\sum\limits_iX_if(W_i)]\right|\nonumber\\
%&\leq \frac{1}{2}E\left|\sum\limits_i X_i(W-W_i)^2f''(W_i^*)\right|+\left|E[\sum\limits_iX_if(W_i)]\right|\nonumber\\
\leq \frac{\|f''\|}{2}E\left|\sum\limits_i
X_i(W_i-W)^2\right|+\left|E[\sum\limits_iX_if(W_i)]\right|.%,\label{eq-term2}
%&\leq \frac{1}{\sigma\sub{\nt}}E\left|\sum\limits_i X_i(W_i-W)(f'(W)-f'(W_i))\right|+E\left|\right|\frac{1}{\sigma\sub{\nt}}\left|\sum\limits_iX_if(W_i)\right|.
\end{align*}
}
The first term of this result again can be bounded using the AM-GM inequality by $\cone B^3\frac{\nt \sltk^2}{\sigma\sub{\nt}^3}$, where $\cone$ is a constant. Recall that $\sltk$ upper bounds the size of the neighborhood of $k$ hops.
The second part of (A2) now is bounded by $\ctwo B\frac{\nt\alpha(k)}{\sigma\sub{\nt}}$, using the usual relationship between covariances and strong mixing coefficients. Thus the overall bound on $(A2)$ is as follows:
{\small\begin{align*}
(A2)\leq \cone B^3\frac{\nt \sltk^2}{\sigma\sub{\nt}^3}+\ctwo B\frac{\nt\alpha(k)}{\sigma\sub{\nt}}.
\end{align*}}

Now we need to bound $(A1)$. Denote $P_\nt=\sum_i X_i(W_i-W)$. Note that \emph{if $X_i$ and $W_i$ were independent}, we would have $E[P_{\nt}]=-E(W^2)=-1$, since $W$ is centered and scaled appropriately. For us however $E[P_\nt]$ does not equal $-1$; instead it becomes smaller as we increase $k$. We thus bound $(A1)$ as follows:
{\small\begin{align*}
(A1)\leq \|f'\|\sqrt{E[1+P_\nt]^2}\leq \|f'\|\sqrt{(1+E[P_\nt])^2+\mbox{var}(P_\nt)}.
\end{align*}}
Now note that $|1+E[P_\nt]|=|E[\sum_i X_iW_i]|$. Since $E[X_i]=0$,
  {\small\begin{align*}
  |E[\sum_i X_iW_i]|=\left|\sum\limits_i\sum\limits_{j\in \ngtk(i)} \cov(X_i,X_j)\right|\leq c''B^2/\sigma\sub{\nt}^2\sum_i\sum_{m>k}\alpha(m)\satm,
  \end{align*}}
  %$|E[\sum_i X_iW_i]|$ equals $\left|\sum\limits_i\sum\limits_{j\in \ngtk(i)} \cov(X_i,X_j)\right|$. Since
using the fact that $\ngtk(i)=\bigcup\limits_{m>k}\natm(i)$, and for all $j\in \natm(i)$, $\cov(X_i,X_j)=O(\alpha(m)/\sigma\sub{\nt}^2)$. We upper bound
 $|1+E[P_\nt]|$ by $c''B^2\nt\tau_k/\sigma\sub{\nt}^2$.
Using similar arguments (see Appendix) we upper bound $\var(P_\nt)$ by
 $8 B^4\nt\sltk^3/\sigma\sub{\nt}^4+16 B^4 \nt\sltk^2\tau_k/\sigma\sub{\nt}^4$.

Putting the pieces together and using $\cn=\nt/\sigma\sub{\nt}$ we see that
{\small
\begin{align*}
&d_W(W,Z)\leq (A2)+\|f'\|\sqrt{(1+E[P_\nt])^2+\mbox{var}(P_\nt)}\\
&\leq \cone B^3\cn\left(\frac{\sltk}{\sigma\sub{\nt}}\right)^2+ \ctwo B\cn\alpha(k) \\
&+ B^2\sqrt{\cthree \left(\frac{\cn \tau_k}{\sigma\sub{\nt}}\right)^2+\cfour
\cn\left(\frac{\sltk}{\sigma\sub{\nt}}\right)^3+\cfive
\cn\frac{\tau_k}{\sigma\sub{\nt}}\left(\frac{\sltk}{\sigma\sub{\nt}}\right)^2 }.
\end{align*}
}
The result is obtained by optimizing the upper bound over $k\leq T$.
\end{proof}
%As mentioned above, we do not assume any marginal
%independence, which is often made in arguments about normal
%convergence of sums of random variables with local dependence. We
%instead assume a condition on the strong mixing coefficients
%$\alpha(k)$ as in condition~\ref{cond-general-stein1}, i.e.
%$\sum\limits_{k=1}^\infty \alpha(k)k^{\rho-1}<\infty$. Under this condition
%we also have $\sigma^2_n/n<\infty$, i.e. $\sigma^2_n=O(n)$.

Next, we present a sufficient condition for the Wasserstein distance
to vanish asymptotically, implying convergence of $W$ to a standard normal.

%\begin{lemma}
%$W\rightarrow Z$ as $n\rightarrow\infty$ if the following conditions hold:
%\begin{align*}
%&\mbox{(1) } \sigma\sub{\nt}^2/n\rightarrow \sigma \mbox{ for some
%$\sigma>0$},\\
%&\mbox{(2) } \sum\limits_{m=1}^{\infty} \alpha(m) |N_m| < \infty,\\
%&\mbox{(3) There exists a sequence $k_n\rightarrow \infty$ such that
%$\sltkn = o(n^{1/4})$ and $\alpha(k_n) = o(n^{-1/2})$.}
%\end{align*}
%\end{lemma}
%\begin{proof}
%Choosing $k$ from the sequence $k_n$, each term of
%Eq.~\ref{eq-stein-general} can be seen to vanish asymptotically. By
%Stein's theorem~\cite{}, this proves that $W\rightarrow Z$.
%\end{proof}
\begin{lem}\label{lem:stein-nornal-conv}
$W\rightarrow \mathcal{N}(0,1)$ as $\nt\rightarrow\infty$ if the following conditions hold:
%Denote $\cn=\nt/\sigma\sub{\nt}$.
%, and let $d_n$ be a function of the size vector. All the asymptotic statements
%below are w.r.t $n\rightarrow \infty$.
\begin{enumerate}
\item $\cn\rightarrow \infty$.\label{cond-stein-1}
%\item $\sigma\sub{\nt}^2/d_n\rightarrow \sigma \mbox{ for some $\sigma>0$}$
\item \text{There exists a sequence $k(\nt)\rightarrow \infty$ such that the following are satisfied:}
\begin{enumerate}
\item $\cn \alpha(k(\nt))\rightarrow 0$
\item $\cn \frac{\tau\sub{k(\nt)}}{\sigma\sub{\nt}}\rightarrow 0$
\item $\cn \left(\frac{|N_{\leq k(\nt)}|}{\sigma\sub{\nt}}\right)^2\rightarrow 0$.
\end{enumerate}
\end{enumerate}
\end{lem}
\begin{proof}
The above conditions imply that $\alpha(k(\nt))\rightarrow 0$,
$\frac{\tau\sub{k(\nt)}}{\sigma\sub{\nt}}\rightarrow 0$, and $\left(\frac{|N_{\leq
k(\nt)}|}{\sigma\sub{\nt}}\right)^2\rightarrow 0$ (and thus $\frac{|N_{\leq
k(\nt)}|}{\sigma\sub{\nt}}\rightarrow 0$ as well) as $\nt\rightarrow \infty$.
Hence the product of two vanishing sequences,
$\cn \left(\frac{|N_{\leq k(\nt)}|}{\sigma\sub{\nt}}\right)^2 \times \frac{|N_{\leq
k(\nt)}|}{\sigma\sub{\nt}}$, also vanishes.
Similarly $\left(\cn\frac{\tau_k}{\sigma\sub{\nt}}\right)\left(\frac{|N_{\leq
k(\nt)}|}{\sigma\sub{\nt}}\right)^2 $ also vanishes as $T\rightarrow \infty$.
Thus, all terms on the right hand side of
Eq.~\eqref{eq-stein-general} vanish, thus proving $W\stackrel{d}{\rightarrow} \mathcal{N}(0,1)$.
\end{proof}

%% file: dist-conv.tex
\section{Weak Convergence of our Estimator}
\label{sec:dist-conv}
In this section we bring together the results from the previous two sections
to establish weak convergence of our estimator.
%We will start by recalling some of our old notation. Our graph evolution model is Markovian with a finite state space. We denote the time of entering some communication class by  $T_1$, and the event by $\mathcal{E}_{T_1}$. We also denote by $\mathcal{C}$ the event that the chain will eventually end up in $C$. Thus the event of entering class $C$ at time $T_1$ is denoted by $\mathcal{E}_{T_1}\bigcap\mathcal{C}$.
%
%
%Recall the following definitions:
%\begin{align*}
%\est{\h}(t)&:=\dfrac{1}{\nn}\sum\limits_{i=1}^\nn \Kb(\dch,\dchq)\dcnpluso\\
%\est{\f}(t)&:=\dfrac{1}{\nn}\sum\limits_{i=1}^\nn \Kb(\dch,\dchq)\dcno\\
%q_t&:=\est{\h}(t)-E[\est{\h}(t)|\ec]-\g(\est{\f}(t)-E[\est{\f}(t)|\ec])
%%\Bt&:=E[\est{\h}|\ec]/E[\est{\f}|\ec]-g
%\end{align*}
% Note that $\est{\h}$ equals $\sum_t \est{\h}(t)/(\nt-p)$ and $\est{\f}$ equals $\sum_t \est{\f}(t)/(T-p)$. We define $\Bt$ as $E[\est{\h}|\ec]/E[\est{\f}|\ec]-g$.
%Denote by $\sigma^2\sub{\nt}(T_1,C)$  $\var(\sum_t q_t|\et,\ec)$.
%Recall that using Lemma~\ref{lem:qm-conv}, we have $\sigma\sub{\nt}(T_1,C)/\sqrt{T}\rightarrow \sigma_c$ for some non-negative $\sigma_c$. Also, using Lemma~\ref{lem:ourexp}  $E[\est{\f}|\ec]\rightarrow R_c$ for $R_c>0$ as $T\rightarrow\infty$. We are now ready to state the normal convergence result for our estimator.
%\vspace{-.5em}

Recall that our estimator $\tilde{g}\sub{\nt}$ is defined in Equation~\ref{eq:hatf2}.
Recall also the definitions of $\est{\h}(t)$, $\est{\f}(t)$ and $q_t$ from
Equations~\ref{eq:htft} and~\ref{eq:qt}.  From Lemma~\ref{lem:est-g-dom}
we have $|\sqrt{T}(\tilde{g}\sub{\nt}-\est{g})|=O(\sqrt{T}\bti)$, where $\bti$ denotes the bandwidth for the pair-specific kernel function (see Equation~\ref{eq:innerkernel}).
Hence, with $\bti=T^{-(1/2+\epsilon)}$ for some $\epsilon>0$, we see
that $\sqrt{T}(\tilde{g}\sub{\nt}-\est{g})\stackrel{a.s.}{\rightarrow}0$.
We will show (in Proposition~\ref{prop-normal-conv}) that under suitable
conditions $\sqrt{\nt}(\est{\g}-\g)$ converges to a mean-zero normal
distribution. Hence, we also have the same normal distribution as the
limit of $\sqrt{\nt}(\tilde{\g}\sub{\nt}-\g)$ under the same conditions.

\begin{proposition}\label{prop-normal-conv}
Let Assumption~\ref{assumption:smooth} hold. If $\sigma_c>0$, and  $\bt=T^{-(1/2+\theta)}$ for some $\theta>0$, then:
{\small
%\vspace{-.5em}
\begin{align*}
\mbox{Conditioned on \ec,}\qquad \sqrt{T}(\est{\g}-\g)\stackrel{d}{\rightarrow} \mathcal{N}(0,\sigma_c^2/R_c^2)\qquad\mbox{As $\nt\rightarrow\infty$.}
\end{align*}}
%\vspace{-1em}
where $S_T$ is the state of the Markov chain at time $T$.
\end{proposition}
\begin{proof}[Proof]
%For bounded $q_t$, proposition~\ref{prop-normal-conv} (a) implies that $\sqrt{\nt}\sum_{t} q_t/\sigma_c\rightarrow \mathcal{N}(0,1)$ conditioned on $\et\bigcap\ec$.
%\begin{align}\label{eq-normal-conv1}
%\sqrt{\nt}\sum_{t} q_t/\sigma\rightarrow \mathcal{N}(0,1)
%\end{align}
From Equation~\ref{eq:est-g-bias} we see that $\sqrt{\nt}(\est{\g}-\g)$ equals $\left.(\sum_{t}q_t/\sqrt{\nt})\right/\est{\f}+\left.(E[\est{\f}|\ec]/\est{\f})\right/\sqrt{\nt}\Bt$.
Using the following lemma (Lemma~\ref{lem:normal-conv-part2}) we know that
the numerator of the first term converges to a $\mathcal{N}(0,\sigma_c^2)$
distribution. Using Lemmas~\ref{lem:ourvar} and~\ref{lem:ourexp} we have
$\est{\f}\stackrel{P}{\rightarrow} R_c$ for a positive constant $R_c$,
conditioned on $\ec$. Hence using Slutsky's lemma the first part converges
conditionally to $\mathcal{N}(0,\sigma_c^2/R_c^2)$.
Also, $E[\est{\f}|\ec]/\est{\f}$ converges to one in probability
conditioned on \ec. Finally, invoking Lemma~\ref{lem:ourexp} and
Lemma~\ref{lem:bias} we see that since $\Bt=O(\bt)$, for
$\bt\sim \nt^{-(1/2+\theta)}$, the second part is $o_P(1)$.
Now, Slutsky's lemma and the continuous mapping theorem yield
the statement of the proposition.
\end{proof}

\begin{lem}\label{lem:normal-conv-part2}
Under Assumption~\ref{assumption:smooth} and assuming $\sigma_c>0$,
{\small
\begin{align*}
\mbox{Conditioned on \ec,}\qquad \sum_t q_t/\sqrt{T}\stackrel{d}{\rightarrow}\mathcal{N}(0,\sigma^2_c) \qquad\mbox{As $\nt\rightarrow\infty$}.
\end{align*}}
%\vspace{-1em}
\end{lem}
%\vspace{-1em}
The proof of this result uses Lemma~\ref{lem:cond-exp} and is deferred to the Appendix.

\begin{lem}\label{lem:normal-conv-part1}
Define $p_t:=[\est{\h}(t)-\g\est{\f}(t)]-E[\est{\h}(t)-\g \est{\f}(t)|\et,\ec]$.
Under Assumption~\ref{assumption:smooth} and assuming $\sigma_c>0$, for any
finite $T_1$, we have:
%$$\sqrt{T}(\est{\g}-\g-\Bt)\stackrel{d}{\rightarrow} \mathcal{N}(0,\sigma_c^2/R_c^2)\qquad\mbox{conditioned on $\ec\bigcap\et$ as $T\rightarrow\infty$}$$
{\small
\begin{align*}
\sum_{t\geq T_1+\mc} p_t/\sqrt{T}\stackrel{d}{\rightarrow} \mathcal{N}(0,\sigma_c^2)\qquad\mbox{conditioned on $\et\bigcap\ec$ as $T\rightarrow\infty$}.
\end{align*}}
\end{lem}
%\vspace{-2em}
\begin{proof}[Proof Sketch]
%We will prove the above result in two parts.
 First we prove that, for a sequence $k(T)=c\log\nt$ for a properly chosen $c$,
the conditions in Lemma~\ref{lem:stein-nornal-conv} are satisfied for
$$W_\nt := \left.\left(\sum\limits_{t\geq T_1+\mc}p_t\right)\right/\sqrt{\var(\sum\limits_{t\geq T_1+\mc}p_t|\et,\ec)}.$$
We also show that for this value of $k$, the upper bound on the Wasserstein distance in Lemma~\ref{lem-stein-main} is $O(\log^2(\nt)/\nt)$.
The details are deferred to the Appendix.
 Now Lemma~\ref{lem:stein-orig} gives:
{\small
\begin{align*}
W_\nt\stackrel{d}{\rightarrow} \mathcal{N}(0,1)\qquad\mbox{conditioned on $\et\bigcap\ec$}.
\end{align*}}
%\vspace{-1em}
However, for finite values of $T_1$, $\var(\sum_{t\geq T_1+\mc}p_t|\et,\ec)/T\rightarrow \sigma^2_c$ (Lemma~\ref{lem:cond-var} and Equation~\ref{eq-var-endpart}).
Thus, the additional assumption of $\sigma_c>0$ proves the result.
\end{proof}%
%\vspace{-1em}
\begin{remark}
Proposition~\ref{prop-normal-conv} shows that, under some weak assumptions, $W_\nt$ converges to a standard normal distribution conditioned on \ec. Since there are a finite number of closed communication classes, unconditionally $W_\nt$ converges to a mixture of zero-mean Gaussians, the mixture proportions being determined by the probability of reaching the communication classes from the start state.
\end{remark}
%\vspace{-1em}
\begin{remark}
\label{remark-stein-cyclo-stationary}
We have established weak convergence for the case where $C$ is aperiodic.
However, as in Remark~\ref{remark:variance-cyclo}, we can consider $\mathcal{M}'$,
which is a Markov chain where each transition corresponds to $d$ transitions of
the original chain.
%Now, $\mathcal{M}'$ is irreducible
%and aperiodic (since $C$ was irreducible and had period $d$).
Again, any sum of the form $\sum_{t=1}^\nt q_t/\sqrt{T}$ can be written as
$1/\sqrt{d}\left(\sum\limits_{t=1}^{\lfloor T/d\rfloor}q'_t/\sqrt{T/d}+o_P(1)\right)$.
$q'_t$ now denotes the sum of the $d$ consecutive $q_t$'s.
%Since, $q_t$ is a function of two states $S_t$ and $S_{t+d}$ from the original Markov chain, we have the same alpha mixing argument from before. The only difference now is that the distance between $q'_t$ ($q_t$) and $q'_{t+kd}$ ($q_k$) is $k-3$, and not $k-1$. \txtred{Check}.
%Using equation~\ref{eq-cyclo-cov} we see that for
For $q'_i:=\sum\limits_{j=id+1}^{(i+1)d} q_j$, we have $\cov(q'_t,q'_{t+k})=O(\alpha(k-1))$ using Equation~\ref{eq-cyclo-cov}.
Thus the first sum again brings us to the irreducible aperiodic setting (with a slightly modified distance function), and hence normal convergence can be established.
\end{remark}
%\vspace{-1em}

%% file: related.tex
\section{Related Work}
\label{sec:related}
Existing work on link prediction in dynamic networks can be broadly divided
into two categories: link prediction based on generative models and link
prediction based on structural features.

A substantial amount of work has gone into the development of generative models
of graph structure based on the formalism of Markov random fields, loglinear
models or other graphical models~\citep{xing2010a,hanneke2006,xing2010b,sarkar05a,
holland-leinhardt,snijders97,vu_nips_2011}.  For example, ~\citet{hanneke2006}
present a dynamic loglinear model based on evolution statistics such as ``edge
stability,'' ``reciprocity'' and ``transitivity.'' \citet{xing2010a} propose
an extension of the mixed membership block model to allow a linear Gaussian
trend in the model parameters.  \citet{Zhou_timevarying} present a nonparametric
approach to estimating a time-varying Gaussian graphical model where the
covariance matrix changes smoothly over time. The discrete analog of this
is considered in~\citep{xing2010b}, where the goal is to learn the latent
structures of evolving graphs from a time series of node attributes.
The static model of~\citet{hoff:latent} is extended by~\citet{sarkar05a}
by allowing smooth transitions in latent space.  All of these models have
the virtue of a clean probabilistic formulation such that link prediction
can be cast in terms of Bayesian posterior inference.  Obtaining this
posterior is, however, often infeasible in large-scale graphs.  Moreover,
these models often make strong model assumptions, not only for the graph
structure but also for the network dynamics, which is often modeled as
linear.

Alternatives to generative models generally revolve around the definition
of various static features that aim to capture structural properties of
graphs.  These are extended to the dynamic setting via heuristics or via
autoregressive modeling.  For example, \citet{arimaHuang} propose a linear
autoregressive model for link prediction and investigate simple combinations
of static graph-based similarity measures (e.g., Katz, common neighbors)
with their autoregressive model to capture transitive similarities in
networks. A similar parametric approach can be found in~\citet{richard_nips_2012}, where a vector autoregressive model was used
for link prediction in dynamic graphs. The authors assume a low rank structure of the graph adjacency matrices and propose proximal methods for inference.

\citet{TylendaTimeaware} examine simple temporal extensions
of existing static measures.  As we have noted earlier, these methods
have the virtue of being applicable to large-scale graphs.  They also
tend to yield surprisingly good performance.  Our work falls into this
general category, while going beyond existing work by providing a formal
statistical treatment of link prediction as a nonparametric estimation
problem.

We conclude this section with a brief discussion on relevant research on nonparametric bootstrap estimators in strong mixing random fields and Markov processes. While these works are not relevant to the link prediction aspect of our work, they are similar because the estimation uses local resampling methods thereby retaining the dependency structure of the data. In the context of strong mixing random fields~\citet{politis:strong-mixing}
consider a blocks of blocks re-sampling method for estimating asymptotically accurate confidence intervals for parameters of the joint distribution of the random field. Nonparametric bootstrap algorithms have also been applied successfully to the area of computer vision.
~\citet{Levina_texturesynthesis} show that one such heuristic algorithm for texture synthesis can be formally framed as a resampling technique for stationary random fields, and prove consistency properties of it under broad conditions. In the context of stochastic processes with an autoregressive structure,~\citet{Paparoditis02thelocal} present the ``local bootstrap''  algorithm, which implicitly estimates the distribution of the one-step transition in the underlying Markov process and generates the bootstrap replicates using this estimated distribution. 

%% file: conc.tex
\section{Conclusions}
\label{sec:conc}
%\vspace{-.1in}
\vspace{-.5em}
In this paper we proposed a nonparametric model (\expnp) for link prediction in dynamic
networks, and showed that it performs as well as the state of the art
for several real-world graphs, and exhibits important advantages over
them in the presence of nonlinearities such as seasonality
	patterns. %We are currently working on longitudinal
	%econometric datasets where seasonality patterns are more marked.
	\expnp also allows us to incorporate features external to
	graph topology into the link prediction algorithm, and its
	asymptotic convergence to the true link probability is guaranteed under our fairly general model assumptions. In addition, we show how to make
\expnp computationally tractable via the use of locality sensitive hashing.
	Together, these make \expnp a useful
	tool for link prediction in dynamic networks.

%% file: appendix.tex
\section{Appendix}
\subsection{Statement and proofs of results from section 5}
%
%<<<<<<< .mine
\setcounter{section}{5}
\setcounter{thm}{2}
\begin{lem}
%Denote  $\eta$ value of the query datacube $\dchq$ for feature $s$ by $\eq$.
As $T\rightarrow \infty$, for some $R_c>0$ (a deterministic function of class $C$),
%\vspace{-.1in}
%{\small
\begin{align*}
%E[\est{h}(s,\dchq)] \rightarrow g(s,\dchq) R_c, \quad
E[\est{f}(s,\dchq)|\et,\ec] \rightarrow R_c & \mbox{,} & E[\est{f}(s,\dchq)|\ec] \rightarrow R_c.
\end{align*}
%}
\vspace{-2em}
\label{lem:ourexpA}
\end{lem}
\begin{proof}
Let $\epsilon$ denote the minimum distance between two datacubes that
are not identical; since the set of all possible datacubes is finite,
$\epsilon>0$. $E[\est{f}(s,\dchq)|\et,\ec]$ is an average
of terms $E[\Kb(\dch,\dchq)\dcno|\et,\ec]$, over $i\in \{1,\dots,\nn\}$
and $t\in \{p,\dots, T-1\}$.
Now, $$E[\Kb(\dch,\dchq)\dcno|\et,\ec] = E\left[e^{-\dcD(\dch,\dchq)/\bt} \dcno|\et,\ec\right].$$
%, as can be seen by summing Eq.~\eqref{eq:Y} over all
%pairs $(i, j)$ in a neighborhood with identical $\dcg$, and then
%taking expectations.
Writing the expectation in terms of a sum over
all possible datacubes, and noting that everything is bounded, gives
the following:
%{\small
\begin{align*}
&E\left[e^{-\dcD(\dch,\dchq)/\bt}\dcno|\et,\ec\right]\\
&=E[\dcno|\dch=\dchq,\et,\ec]P(\dch=\dchq|\et,\ec)+O(e^{-\epsilon/\bt}).
\end{align*}
%}
%\vspace{-1em}
Recalling that $E[\est{f}(s,\dchq)|\et,\ec]$ was an average of the above
terms,  we see that it equals:
{\small
\begin{align}
\frac{1}{n(T-p)}\sum\limits_{t,i} E[\dcno|\dch=\dchq,\et,\ec]\cdot P(\dch=\dchq|\et,\ec)+O(e^{-\epsilon/\bt})\label{eq:AVG_A}.
\end{align}
}
%\vspace{-1em}
We will now show that the above average converges to $\g(s, \dchq) R$
for some $R>0$. The second term in the RHS in eq~\eqref{eq:AVG_A} converges to zero, since $\bt\rightarrow 0$ as $T\rightarrow \infty$.
%For the first term, consider the expression $\sum\limits_{t,i} E[\dcno|\dch=\dchq]\cdot P(\dch=\dchq)/(n(T-p))$.
For the numerator of the first term we have,
$E[\dcno|\dch=\dchq,\et,\ec]\cdot P(\dch=\dchq|\et,\ec) = \sum\limits_{\eta}\eta
P(\dcno=\eta,\dch=\dchq|\et,\ec)$.
Both $\dch$ and $\dcno$ are fully
determined given the current state $S_t$ of the Markov chain.  Using
$I_S(X)$ to denote an indicator of $X$ in state $S$, we have
$P(\dcno=\eta,\dch=\dchq|\et,\ec)=
\sum\limits_{S}I_S(\dcno=\eta,\dch=\dchq)P(S_t=S|\et,\ec)$.
As a result of
this, the first term in the R.H.S of eq~\eqref{eq:AVG_A} becomes an average of the form
$\frac{1}{T}\sum\limits_t \sum\limits_S \xi(S) P(S_t=S|\et,\ec)$, where $\xi(S)=\frac{1}{n}\sum\limits_{i,\eta}\eta I_S(\dcno=\eta,\dch=\dchq)$.  Since
we have a finite state-space and $\xi(S)$ is bounded, we can rewrite the above expression as $\sum\limits_S \xi(S) \frac{\sum\limits_t  P(S_t=S|\et,\ec)}{T}$.

Now, recall that the query
datacube at $T$ is a function of the state $S_T$, which
belongs to a closed irreducible set
$C$ with probability $1$.
Due to stationarity (or cyclic stationarity with a
finite cycle length) the average $\sum\limits_t  P(S_t=S|\et,\ec)/T$ converges to some constant $R(S)$
(constant because it is a function of the finite state space). For the special case of $S=S_T$, we have the following:
(a) $S_T\in C$, so $R(S_T)>0$, and
(b) $S_T$ contains at least one pair of nodes with the feature vector $s$ (since we are attempting link prediction for such a pair),
 so there exists some $\eta>0$ for which $I_{S_T}(\eta,\dchq)=1$.
 Together, these imply that $\sum\limits_S \xi(S) \left(\sum\limits_t  P(S_t=S|\et,\ec)/T\right)$ converges to some $R_c>0$, where $R_c$ is a deterministic function of communication class $C$.

Noting that $E[\est{\f}(s,\dchq)|\ec]=E[E[\est{\f}(s,\dchq)|\et,\ec]|\ec]$, and the fact that $\est{\f}$ is bounded we invoke the Dominated Convergence Theorem
and see that $E[\est{\f}(s,\dchq)|\ec]\rightarrow R_c$ as well, thus completing the proof of the theorem.

\end{proof}
%\mypara{Statement and proof of lemma 5.4}
\begin{lem}
\label{lem:biasA}
Define $\Bt(s,Q,C)=(E[\est{\h}(s,Q)|\ec]-gE[\est{\f}(s,Q)|\ec])/E[\est{\f}(s,Q)|\ec]$. If assumption 1 holds, then, we have $\Bt=O(\bt)$. Since $\bt\rightarrow 0$ as $T\rightarrow \infty$, this implies $\Bt=o(1)$.
\end{lem}
\begin{proof}
For $t\in [p,T-2]; i\in [1,N]; s=\dcgq$,
the numerator of $\Bt$ is an average of the terms:
{\small
\begin{align*}
A_t:=E\left[\Kb(\dch,\dchq)\dcnpluso|\ec\right]-E\left[\Kb(\dch,\dchq)\dcno|\ec\right]\g(s,\dchq).
%E\left[\Kb(\dch,\dchq)\dcnpluso\right]-E\left[\Kb(\dch,\dchq)\dcno\right]\g(s,\dchq)\\\text{for
%} t\in p:T-2;\ \ i\in 1:N; s=\dcgq
\end{align*}
%$$E[\Kb(\n^t_i,\n_q)\np{t+1}{ij}]-E[\Kb(\n^t_i,\n_q)\nt{t+1}{ij}]\g(\n_q,\st_q)\ \ \text{for } t\in 1:T-2,\ \ i\in 1:N$$
}
Taking expectations w.r.t. \dch, and denoting
$\Kb(\dch,\dchq)$ by $\gamma$, the first term becomes:
%{\small
\begin{eqnarray}
E\left[\gamma \dcnpluso|\ec\right] & = &
E\left[\gamma E\left[\dcnpluso | \dch,\ec\right]|\ec\right]\nonumber.
%  =  E\left[\gamma \dcno\g(s, \dch)|\ec\right]\nonumber
% & = & E\left[\Kb(\dch,\dchq) \dcno\cdot \g(s, \dch)\right]\nonumber
\end{eqnarray}
%}
Now note that $E\left[\dcnpluso | \dch,\ec\right]=E[E[\dcnpluso|\dch,\et,\ec]|\ec]$. Conditioning on $\et$ makes $\dcnpluso$ conditionally independent of
\ec given $\dch$ if $t>T_1$. Also, for $t\geq T_1$, $E\left[\dcnpluso | \dch,\et,\ec\right] =\dcno\cdot \g(s, \dch)$, as
can be seen by summing Eq. 2.3 over all pairs $(i, j)$ in a
neighborhood with identical $\dcg$, and then taking expectations\footnote{Note that the conditioning on \et is crucial here.}.
This along with the fact that $\gamma\dcnpluso$ is bounded leads to:
%{\small
\begin{align*}
E[\dcnpluso|\dch,\et,\ec]&\leq \dcno g(s,\dch) \ind[T_1\leq t] + c\ind[T_1> t] \\
&\leq \dcno g(s,\dch)+c\ind[T_1> t].
\end{align*}
%}
 Thus the numerator of \Bt can be upper bounded as:
{\small
\begin{align*}
 |\sum_t A_t/T| &\leq \sum_t  |E[\gamma\dcno (g(s,\dch)-g(s,\dchq))|\ec]|/T+c'\sum_t P [T_1> t]/T.
\end{align*}
}
The second part is simply $O(E[T_1]/T)$ and $o(1)$.
%where the last equality follows from the fact that
%$E\left[\dcnpluso | \dch,\ec\right] = E\left[\dcnpluso | \dch\right]=\dcno\cdot \g(s, \dch)$, as
%can be seen by summing Eq.~\ref{eq:Y} over all pairs $(i, j)$ in a
%neighborhood with identical $\dcg$, and then taking expectations.
%Thus, Eq.~\ref{eq:B1} equals
%$E\left[\Kb(\dch,\dchq) \dcno\cdot \g(s, \dch)\right]$.
%where the last equality follows by summing Eq.~\ref{eq:Y} and taking
%expectations.
Thus, the numerator of $\Bt$ becomes an average %over $t,i$
of the terms of the following form:
$$E\left[\Kb(\dch,\dchq) \dcno\cdot \left(\g(s, \dch) - \g(s,
\dchq)\right)|\ec\right].$$

This expectation is over all possible configurations of the
neighborhoods $\dcN{t}$ and $\dcN{t+1}$.
Since our neighborhood sizes are bounded (because $n$ is bounded),  the expectation
is a sum over a finite number of terms.

We now use the smoothness assumption on $\g$. Using $\left|\g(s, \dch) - \g(s,
\dchq)\right|=O(\dcD(\dch,\dchq))$ and that $\dcno$ is finite for all $T$ and Lemma~\ref{lem:ourexpA},  we have:
{\small
\begin{align*}
\Bt= O \left(E[\dcD(\dch,\dchq)e^{-\dcD(\dch,\dchq)/\bt}|\ec]\right) =O(\bt).
\end{align*}
}
The last equation holds since for non-negative $x$, $xe^{-x/\bt}\leq \bt/e$.
%\vspace{-1em}
%We now let $b\rightarrow 0$. Since the expectation is over a finite
%sum, so we can take the limit inside the expectation.
%Since $\lim_{b\rightarrow 0} \Kb(\dch, \dchq)$ is zero unless
%$\dch=\dchq$ (Eq.~\ref{eq:bandwidth}),
%%. For any $i,t$ the expectation is over a finite number of terms, and so we can exchange limit and expectation.
%we find that
%$E\left[\Kb(\dch,\dchq) \dcno\cdot \left(\g(s, \dch) - \g(s,
%\dchq)\right)\right]\rightarrow 0$, for any $i$ and $t$.  So the numerator of $B$
%goes to zero asymptotically. Hence, the estimator $\est\g$ is
%unbiased.
\end{proof}

%\setcounter{thm}{3}
%\mypara{Statement and proof of Lemma 5.5}
\begin{lem}
\label{lem:diff-expA}
Consider an irreducible and aperiodic finite state Markov chain with probability transition matrix $P$, starting distribution $\pi_0$ and stationary distribution $\pi$.
Let $X_t$ be a deterministic function (with finite support) of the state at time $t$. The expectation of $X$ under the distribution at time $t$ is denoted by $E[X_t|\pi_0]$. Let $\mu$ denote the expectation of $X_\infty$ (i.e. under distribution $\pi$).
 There exists a constant $\lambda\in(0,1)$, and a constant $M$ such that, $\forall t>M$, $\max_{x\in \mathcal{S}} \sum\limits_{y\in \mathcal{S}}|P(x,y)-\pi(y)|=O(\lambda^t)$, and
 $|E[X_t|\pi_0]-\mu|=O(\lambda^t)$.
 \end{lem}

\begin{proof}
Using the same line of reasoning as~\cite{finitemarkov2005}, we first prove the above for $\max_x\sum_{y\in\mathcal{S}}|P^t(x,y)-\pi(y)|$. Here $|\mathcal{S}|$ denotes the state space and $P$
 the $|\mathcal{S}|\times |\mathcal{S}|$ probability transition matrix associated with the Markov chain.
 %Let $[|.|_ij]$ denote the matrix of absolute values $|P(i,j)-\pi(j)|$.
  Denote by $\Pi$ the matrix $\mathbf{1}\pi^T$, where $\mathbf{1}$ denotes the column vector of all ones. Note that since $P\Pi=\Pi$ and $\Pi P=\Pi$, we have $P^t-\Pi=(P-\Pi)^t$. For a finite state space irreducible and aperiodic Markov chain, $|P^t(x,y)-\Pi(x,y)|\rightarrow 0$, as $t\rightarrow\infty$. Hence for some positive $\delta<1$, we can find an $M$ s.t. $\forall t>M$, $\sum_y |P^t(x,y)-\Pi(x,y)|\leq \delta$, $\forall x\in \mathcal{S}$. Since $\max_x \sum_y |P^t(x,y)-\Pi(x,y)|=||P^t-\Pi||_\infty$, using matrix norm inequalities we have for $t=kM+\ell$, where $\ell<M$ and $t>M$,
$$|P^t-\Pi|_\infty\leq ||P^M-\Pi||_\infty^k||P^\ell-\Pi||_\infty=O(\delta^k),$$
since $\max_{\ell\leq M}||P^\ell-\Pi||_\infty$ is a constant.
However, $\delta^k=\delta^{k+1}/\delta=O(\lambda^t)$, where $\lambda=\delta^{1/M}<1$. Now for $t>M$ and $\lambda<1$, we have: $$\max_x |P^t(x,y)-\pi(y)|=O(\lambda^t).$$

First consider $\pi_0$ to be an atom at a state $x_0\in\mathcal{S}$. Since $|E(X_t|X_0)-\mu|\leq \sum_{x\in\mathcal{S}}|x||P(x_0,x)-\pi(x)|$, using that $X_t$ is bounded we have the main result. The result can be easily extended to the more general case where  $\pi_0$ is a convex combination of atoms at $x\in\mathcal{S}$.
\end{proof}

\setcounter{thm}{6}
\setcounter{equation}{6}
%\mypara{Statement and proof of Lemma~\ref{lem:cond-varA}}
\begin{lem}
\label{lem:cond-varA}
 For any finite integer $k$, we have
 \begin{align}
 \var(\sum_{t\geq T_1+\mc} q_t|\mathcal{E}_{T_1},T_1=k,\ec)/T&\rightarrow\sigma_c\qquad\mbox{for some $\sigma_c\geq 0$}\label{eq-var-endpartA}\\
 \var(\sum_t q_t|\mathcal{E}_{T_1},T_1=k,\ec)/T&\rightarrow\sigma_c\qquad\mbox{for some $\sigma_c\geq 0$}\label{eq-var-totA}.
 \end{align}
%$\var(U_\nt|\mathcal{E}_{T_1})=O(T_1^2/T)+\frac{1}{T}\sum\limits_{\ell=0}^{d-1}\sum\limits_{r=0}^{R_t-1}c_{T_1+rd+\ell}$.
For a finite state space Markov chain, we also have $E[\var(U_\nt|\mathcal{E}_{T_1},\ec)|\ec]\rightarrow \sigma_c$ for some $\sigma_c\geq 0$.
\end{lem}
%
%\begin{lem}
%\label{lem:cond-varA}
% For any finite integer $k$, we have $\var(U_\nt|\et,T_1=k,\ec)\rightarrow\sigma_c$, for some $\sigma_c\geq0$.
%%$\var(U_\nt|\et)=O(T_1^2/T)+\frac{1}{T}\sum\limits_{\ell=0}^{d-1}\sum\limits_{r=0}^{R_t-1}c_{T_1+rd+\ell}$.
%For a finite state space Markov chain, we also have $E[\var(U_\nt|\et,\ec)]\rightarrow \sigma_c$ for some $\sigma_c\geq 0$.
%\end{lem}
%\begin{lem}
%\label{lem:cond-varA}
%Let $R_t = \lfloor (T-t)/d \rfloor$. Let $T_1$ denote the time to hit $C$, and also let $\et$ denote the event that the chain hit $C$ at $T_1$. For any finite integer $k$, we have $\var(U_\nt|\et,T_1=k)\rightarrow\sigma$, for some $\sigma\geq0$.
%%$\var(U_\nt|\et)=O(T_1^2/T)+\frac{1}{T}\sum\limits_{\ell=0}^{d-1}\sum\limits_{r=0}^{R_t-1}c_{T_1+rd+\ell}$.
%For a finite state space Markov chain, we also have $E[\var(U_\nt|\et)]\rightarrow \sigma$ for some $\sigma\geq 0$.
%\end{lem}

\begin{proof}
Let $C$ have (finite) period $d$; the period is finite from the finiteness of
the Markov chain, and is typically very small (e.g., $d=1$ if
$0<g(.)<1$ everywhere).
Let $\mathcal{M}'$ be a Markov chain where
each transition corresponds to $d$ transitions of the original chain.
Now, $\mathcal{M}'$ is irreducible
and aperiodic (since $C$ was irreducible and had period $d$).
Thus, $\exists \mc,\lambda\in(0,1)$ s.t. $\forall t\geq\mc$, it is geometrically ergodic with rate $\lambda$ (Lemma~\ref{lem:diff-expA}), which implies in turn
that for $t\geq\mc$, $\mathcal{M}'$ is strongly mixing with exponential drop-off~\cite{Pham} for large $k$:
$\alpha(k)\sim e^{-\beta k}$ for some $\beta>0$.
Thus, distant states are almost independent, and we use this to bound
the covariances of the \qit, as follows. Also define $q_t=\sum_{i}\qit/n$.

For the first term, we have:
{\small
\begin{align*}
&(1/T)\var\left[\sum\limits_{t=1}^{T} q_t|\et,\ec\right]\\
% =(1/T)\sum\limits_{t,t'} \cov(q_t, q_{t'}|\et,\ec)\\
& = \underbrace{(1/T)\sum\limits_{t<T_1,t'<T_1} \cov(q_t, q_{t'}|\et,\ec)}_{(P_0)}+\underbrace{(1/T)\sum\limits_{t\geq T_1} \var(q_t|\et,\ec)}_{(P_1)}\\
&+\underbrace{(2/T)\sum\limits_{t<T_1,t'\geq T_1} \cov(q_t, q_{t'}|\et,\ec)}_{(P_2)}.
%& = \sum_{T_1=1}^T (1/T)\sum\limits_{i,j}\sum\limits_{t,t'} \cov(\qit, q_{t'})\\
%&= O(T_1^2/T) + (1/T)\sum\limits_{i,j}\sum\limits_{{\scriptsize\begin{array}{c} t\geq T_1,\\ t'\geq T_1\end{array}}} \cov(q_t, q_{t'})\\
%& = O(T_1^2/T) + (2/T)\sum\limits_{i,j}\sum\limits_{t\geq T_1}{\underbrace{\sum\limits_{t'\geq t} \cov(q_t q_{t'})}_{A_t}}
\end{align*}
}
First, note that $P_0=O(T_1^2/T)$. %Also, since $|\cov(X,Y)|\leq \sqrt{\var(X)\var(Y)}$, we have:
%\begin{align}\label{eq-var-p1}
%P_1\leq \sqrt{\frac{\var(\sum\limits_{t<T_1}q_t|\et,\ec)}{T}\frac{\var(\sum\limits_{t'\geq T_1}q_{t'}|\et,\ec)}{T}}=O\left(T_1\sqrt{P_2}/\sqrt{T} \right)
%\end{align}
We now focus on $P_1$.  Let $U:=\sum\limits_{T_1\leq t< T_1+\mc}q_t$, and $V:=\sum\limits_{t\geq T_1+\mc}q_t$. Thus,
 $$\var(\sum\limits_{t\geq T_1}q_t|\et,\ec)=\var(U|\et,\ec)+\var(V|\et,\ec)+\cov(U,V|\et,\ec).$$
$\var(U|\et,\ec)=O(\mc^2)$, as for $\var(V|\et,\ec)$, we have:
%First, note that
$$\var(V|\et,\ec)
%= (1/T)\sum\limits_{{\scriptsize\begin{array}{c} t\geq T_1,\\ t'\geq T_1\end{array}}} \cov(q_t, q_{t'})
=(2/T)\sum\limits_{t\geq T_1+\mc}{\underbrace{\sum\limits_{t'\geq t} \cov(q_t,q_{t'}|\et,\ec)}_{A_t}}.$$
Recall that for our Markov chain, $S_t$ involves $p+1$ graphs ($G_{t-p+1},\dots,G_{t+1}$). Since $p_t$ is a function of $S_t$, it also depends on $p+1$ graphs. Hence, the distance $\dist(t,t')$ between two sigma-algebras $\mathcal{F}_{\leq t}$ and $\mathcal{F}_{> t'}$ is defined as $\max(\lceil(t'-t-(p+1))/d\rceil,0)$ \txtred{CHECK}. Thus, the total number of states at distance $k$ is $O(1)$.
Let $R_t = \lfloor (T-t)/d \rfloor.$ Rather importantly, note that we will use basic conditional independence results from Markov chains. For example
$E[X_t X_{t+2d}|X_{t+d}]=E[X_t|X_{t+d}] E[X_{t+2d}|X_{t+d}]$. Unfortunately, conditioned on $\et\bigcap\ec$ this may not be true. However, if $t\geq T_1$, we can safely use the conditional independence, which is definitely true for $A_t$.

%Bounding $A_t$ becomes somewhat cumbersome in presence of cyclo-stationarity.
For notational convenience
we will denote by $\cov_{c}$ and $E_{c}$ covariance and expectation conditioned on $\et\bigcap\ec$.
Then,
{\small
\begin{align*}
A_t &=\sum\limits_{t\leq t' < t+(R_t-1)d} \cov_{c}(q_t, q_{t'}) + \sum\limits_{t+R_td\leq t'\leq T }\cov_{c}(q_t,q_{t'})\\
& = \sum\limits_{r=0}^{R_t-1}\sum\limits_{\ell=0}^{d-1} \cov_{c}(q_t, q_{t+rd+\ell})+\sum\limits_{t+R_td\leq t'\leq T }\cov_{c}(q_t,q_{t'})\\
& = \sum\limits_r \left(E_{c}[q_t \sjtr] - E_{c}[q_t]E_{c}[\sjtr]\right)+\left(E_{c}[q_t \sjtR] - E_{c}[q_t]E_{c}[\sjtR]\right)\\&\quad\mbox{(letting
$\sjtr=\sum\limits_{\ell=0}^{d-1} q_{t+rd+\ell}$ and $\sjtR=\sum\limits_{t'\geq t+R_td}^T q_{t+rd+\ell}$)}\\
& = \sum\limits_r \left(E_{c}[E_{c}[q_t\sjtr \mid S'_{t+rd}]] - E_{c}[q_t]E_{c}[\sjtr]\right)+\left(E_{c}[E_{c}[q_t\sjtR \mid S'_{t+R_td}]] - E_{c}[q_t]E_{c}[\sjtR]\right)\\
& = \sum\limits_r \left(E_{c}[E_{c}[q_t\mid S'_{t+rd}] E_{c}[\sjtr \mid S'_{t+rd}]] - E_{c}[q_t]E_{c}[E_{c}[\sjtr\mid S'_{t+rd}]]\right)\\
&+ \left(E_{c}[E_{c}[q_t\sjtR \mid S'_{t+R_td}]] - E_{c}[q_t]E_{c}[\sjtR]\right) \quad\mbox{By Markov property}\\
& = \sum\limits_r \left(E_{c}[E_{c}[q_t\mid S'_{t+rd}] p(S'_{t+rd})] - E_{c}[q_t]E_{c}[p(S'_{t+rd})]\right)\\
&+\left(E_{c}[E_{c}[q_t\mid S'_{t+R_td}] p(S'_{t+R_td})] - E_{c}[q_t]E_{c}[p(S'_{t+R_td})]\right)\quad\mbox{$E_{c}[\sjtr\mid S'_{t+rd}]$ is denoted as a function $p(.)$}\\
& = \sum\limits_r \left(E_{c}[E_{c}[q_t p(S'_{t+rd})\mid S'_{t+rd}]] - E_{c}[q_t]E_{c}[p(S'_{t+rd})]\right)\\
&+\left(E_{c}[E_{c}[q_t p(S'_{t+R_td})\mid S'_{t+R_td}]] - E_{c}[q_t]E_{c}[p(S'_{t+R_td})]\right)\\
& = \sum\limits_r \left(E_{c}[q_t p(S'_{t+rd})] - E_{c}[q_t]E_{c}[p(S'_{t+rd})]\right) + \left(E_{c}[q_t p(S'_{t+R_td})] - E_{c}[q_t]E_{c}[p(S'_{t+R_td})]\right)\\
&= B_t+\cov_{c}(q_t, p(S'_{t+R_td}))\quad\mbox{where $B_t=\sum\limits_r \cov_{c}(q_t, p(S'_{t+rd}))$}.
\end{align*}
}

Recall that we were originally interested in $\sum\limits_{t>T_1} A_t/T$. Let us first consider $1/T\sum\limits_t \cov(q_t, p(S'_{t+R_td})|\et,\ec)$.
By virtue of geometric ergodicity $\cov(q_t, p(S'_{t+R_td})|\et,\ec)= O\left(e^{-\beta R_t}\right)$, where $R_t = \lfloor (T-t)/d \rfloor$. Thus we have:
%{\small
\begin{align*}
\sum\limits_t |\cov(q_t, p(S'_{t+R_td})|\et,\ec)|=O\left(\sum\limits_t e^{-\beta \lfloor(T-t)/d\rfloor}\right)
%&\leq \const e^\beta \sum\limits_{t=T_1}^T e^{-\beta (T-t)/d}\\
%&\leq \const e^\beta \sum\limits_{t'=0}^{T-T_1} e^{-\beta t'/d}\\
= O\left(\frac{ e^\beta}{1-e^{-\beta/d}}\right).
\end{align*}
%}
Using elementary arguments from real analysis we see that $\sum\limits_t \cov(q_t, p(S'_{t+R_td})|\et,\ec)$ converges to some finite number.
Hence after dividing by $T$ it contributes a $o(1)$ term to the expression $\sum\limits_t A_t/T$.
For this reason we will now concentrate on $\sum\limits_{t>T_1}B_t/T$ term.
First note that the sequence $B_t$ is upper bounded by the following,
{\small
\begin{align*}
B_t & \leq \sum\limits_r |\cov(q_t, p(S'_{t+rd})|\et,\ec)| \qquad\mbox{Also $t>T_1$, and we have conditioned on $\et,\ec$}\\
& \leq O(\sum\limits_r e^{-\beta r})=O(1) \qquad\qquad\qquad \mbox{Since all $q_t$ are
bounded}.\\
\end{align*}
}
 We again see that $B_t$ also converges to some constant $c_{t}$, thus making $P_2$ asymptotically equivalent to:
$\frac{1}{T}\sum\limits_{\ell=0}^{d-1}\sum\limits_{r=0}^{R_t-1}c_{T_1+rd+\ell}$.
%\begin{align*}
%P_2=1/T\sum\limits_t B_t=\frac{1}{T}\sum\limits_{\ell=0}^{d-1}\sum\limits_{r=0}^{R_t-1}c_{T_1+rd+\ell}
%\end{align*}
However, for all $T_1\leq T$,
if the chain is cyclo-stationary, then after a finite time, for any $\ell\in\{0,\dots,d-1\}$, $c_{T_1+rd+\ell}$ approaches the same constant $c_\ell$, $\forall r$.
Therefore, for all $T_1\leq T$ we have $\lim_{R\rightarrow\infty} \left.\sum\limits_{r=0}^{R-1}c_{T_1+rd+\ell}\right/R = c_\ell$, where $c_\ell$ is a constant w.r.t $T$. This leads to:
%Hence, we have  as $T\rightarrow \infty$, ($P_2$) converges to $1/d \sum\limits_{\ell=0}^{d-1} c_\ell$.
%Now  eq~\eqref{eq-var-p1}.
$$\var(V|\et,\ec) \rightarrow 1/d \sum\limits_{\ell=0}^{d-1} c_\ell \qquad \mbox{as $T\rightarrow \infty$.}$$
Since the $P_1$ is a variance term, it is non-negative for all $T$, and hence $\sigma_c=1/d \sum\limits_{\ell=0}^{d-1} c_\ell$ must be non-negative as well, thus proving Equation~\ref{eq-var-endpartA}. Using the Cauchy Schwartz inequality, $$\cov(U,V|\et,\ec)/T=O(\sqrt{(\var(U|\et,\ec)/T)(\var(V|\et,\ec)/T)})=o(1).$$
Thus  $P_1\rightarrow\sigma_c$ as $T\rightarrow\infty$ for some non-negative constant $\sigma_c$.

Another use of the Cauchy Schwartz argument from before, %We also can upper bound the third part by $$\sqrt{\left.\var(\sum_{t<T_1}q_t|\et,\ec)\right/T\left.\var(\sum_{t\geq T_1}q_t|\et,\ec)\right/T}$$.
along with the convergence result on $P_1$ lets us upper bound $P_2$ by $O(T_1/\sqrt{T})$.

Thus, for finite $k$, putting all the bounds (i.e. on $P_0$, $P_1$, and $P_2$) together, we have $\var(\sum_t q_t|\et,\ec,T_1=k)/T\rightarrow\sigma_c$, for some $\sigma_c\geq 0$, proving Equation~\ref{eq-var-totA}. Also, since $T_1$ has finite first and second moments for a finite space Markov chain, we have $E[\var(\sum_t q_t|\et,\ec)|\ec]/T\rightarrow\sigma_c$.

%Combining the above with Equation~\ref{eq-var-p1}, we also have $|P_1|=O(T_1/\sqrt{T})$, since $P_2$ converges to a constant.
%Putting all the bounds (i.e. on $P_0$, $P_1$, and $P_2$) together we have
%\begin{align*}
%\var(U_\nt|\et,\ec)=O(T_1^2/T)+O(T_1/\sqrt{T})+\sigma
%\end{align*}
%
%Thus, for any finite $k$, $\var(U_\nt|\et,\ec,T_1=k)\rightarrow\sigma$ as $T\rightarrow \infty$. Also, the expectation of the conditional variance
%converges, since $T_1$ has finite first and second moments.
We remind the reader that using simple arguments for finite state space Markov chains, it can be shown that $T_1$'s tail probability is geometrically decaying, leading to the finiteness of the first and second moments.
%\begin{align*}
%\lim_{\nt\rightarrow\infty}E[\var(U_\nt)|\et,\ec,\ec]&=\lim_{\nt\rightarrow\infty}E(T_1^2)/T+o(1)+\sigma\\
%&=\sigma\qquad\mbox{Since $T_1$ has a geometric distribution, $E(T_1^2)$ is finite\txtred{CHECK Deepay}}
%\end{align*}

\end{proof}

\begin{lem}
\label{lem:cond-expA}
%Let $R_t = \lfloor (T-t)/d \rfloor$, and also let $E_{t}$ denote the event $\{S_t\in C\}$.
$\var(E[U_\nt|\et,\ec]|\ec)=o(1)$.
\end{lem}
\begin{proof}
%Let $q_t$ denote $\sum_i \qit/n$.
Recall that $U_\nt:=\sum_t q_t/\sqrt{T}$.
Let $\mu_c$ denotes the
 expectation of $q_t$ under the stationary distribution in communication class $C$ (it is a deterministic function of class $C$). Since $\var(E[q_t|\et,\ec]|\ec)=\var(E[q_t|\et,\ec]-\mu_c|\ec)$,
 we will simply upper bound $E[U_\nt-\mu_c|\et,\ec]$. Lemma~\ref{lem:diff-expA} shows that:
 $\exists \mc$, and $\lambda\in (0,1)$ such that, $\forall t>T_1+\mc$,  $|E[q_t|\et,\ec]-\mu_c|=O(\lambda^{t-T_1})$.
Thus, %$$ can be upper bounded as:
{\small
\begin{align}
\label{eq-dev-from-sta}
|E[U_\nt-\mu_c|\ec,\et]|\leq \frac{c(T_1+\mc)}{\sqrt{T}}+\frac{\sum_{t>T_1+M}\lambda^{t-T_1}}{\sqrt{T}}=O\left(\frac{T_1+M}{\sqrt{T}}\right)
\end{align}
}
Thus, $\var(E[U_\nt|\et,\ec])= O\left(E[(T_1+M)^2]/T\right)=o(1)$, since $T_1$ has finite second moment.
%
%Note that $\var(E[U_\nt|\et,\ec])$ can be decomposed as follows:
% \begin{align*}
%%&\var(E[U_\nt|\et,\ec])\\
%\underbrace{\var(\sum_{t\leq T_1+M}E[q_{t}|\et,\ec])/T}_{P_0}+\underbrace{2\cov(\sum_{t'\leq T_1+M}E[q_{t'}|\et,\ec],\sum_{t> T_1+M}E[q_{t}|\et,\ec])/T}_{(P_1)}
%+\underbrace{\var(\sum_{t> T_1+M}E[q_{t}|\et,\ec])/T}_{(P_2)}
%\end{align*}
%
%Also note that  $|\cov(W,X)-\cov(W,Y)|= O(E|W|E|X-Y|)$.
%
%We have, $|(P_0)|=O([T_1]^2/T)$, and we can bound $|(P_1)|$ as follows.
%
% \begin{align*}
%&| \underbrace{\cov(\sum_{t'\leq T_1+M} E[q_{t'}|\et,\ec],\sum_{t'> T_1+M}E[q_t|\et,\ec])/T}_{(P_1)}-\underbrace{\cov(\sum_{t'\leq T_1+M} E[q_{t'}|\et,\ec],(T-T_1-M)\mu)/T}_{(A)}|\\
%&= \underbrace{O(E\left|\sum_{t'\leq T_1+M} E[q_{t'}|\et,\ec]\right| E\left[\lambda^{M}\right]/T)}_{(B)}
% \end{align*}
% Since, $\cov(X,Y)\leq \sqrt{\var(X)\var(Y)}$, and $\var(\sum_{t'\leq T_1+M} E[q_{t'}|\et,\ec])=O(E[(T_1+M)^2])$, and $\var((T-T_1-M)\mu)=O(\var(T_1))$,
% we have  $|(A)|=O(E[(T_1+M)^2]/T)$ \txtred{Deepay, check}. Since $\lambda^{M}\leq 1$, we have $|(B)|=O(E[T_1+M]/T)$. Hence $P_0=O(E\left[(T_1+M)^2\right])$.
%
% Using Minkowski's inequality, we see that $|\var(X)-\var(Y)|=O(\var(X-Y)+ \sqrt{\var(X-Y)\var(Y)})$.
% Setting $X:=\sum_{t>T_1+M}E[q_t|\et,\ec]$ and $Y:=(T-T_1-M)\mu$, we have $\var(X-Y)=O(1)$, and $\var(Y)=O(\var(T_1))$. Combining these gives us,
% $|(P_2)|=O(\var(T_1)/T)$. \txtred{Deepay, check}
%
% Putting everything together we have:
% $\var(E[U_\nt|\et,\ec])=O(E[T_1^2]/T)$.
\end{proof}

\setcounter{section}{10}
\subsection{Statement and proofs of results from section 6}
\setcounter{section}{6}
\setcounter{thm}{1}

\begin{lem}
\label{lem-stein-mainA}
The Wasserstein distance $d_W(W,Z)$ between $W$ and the standard normal random
variable $Z$ is upper bounded as follows:
{\small
\begin{align*}
d_W(W,Z) \leq & \min_{k\leq\nt}\left(\cone B^3\cn\left(\frac{\sltk}{\sigma\sub{\nt}}\right)^2+
\ctwo B\cn\alpha(k) +\right.\nonumber\\
& \left.B^2\sqrt{\cthree \left(\frac{\cn \tau_k}{\sigma\sub{\nt}}\right)^2+\cfour
\cn\left(\frac{\sltk}{\sigma\sub{\nt}}\right)^3+\cfive
\cn\frac{\tau_k}{\sigma\sub{\nt}}\left(\frac{\sltk}{\cn}\right)^2 }\right)\nonumber,
\end{align*}
}
where $\cone,\ctwo,\cthree,\cfour,\cfive$ are constants.
\end{lem}%
%\begin{lem}
%\label{lem-stein-mainA}
%The Wasserstein distance $d_W(W,Z)$ between $W$ and the standard normal random
%variable $Z$ is upper bounded as follows:
%{\small
%\begin{align*}
%d_W(W,Z) \leq & \cone B^3\cn\left(\dfrac{\sltk}{\sigma\sub{\nt}}\right)^2+
%\ctwo B\cn\alpha(k) +\nonumber\\
%& B^2\sqrt{\cthree \left(\frac{\cn \tau_k}{\sigma\sub{\nt}}\right)^2+\cfour
%\cn\left(\frac{\sltk}{\sigma\sub{\nt}}\right)^3+\cfive
%\cn\left(\frac{\sltk}{\cn}\right)^2 \frac{\tau_k}{\sigma\sub{\nt}}},
%%\label{eq-stein-general}
%\end{align*}
%}
%where $\cone,\ctwo,\cthree,\cfour$ and $\cfive$ are constants.
%\end{lem}

%\begin{lem}
%\label{lem-stein-mainA}
%The Wasserstein distance $d_W(W,Z)$ between $W$ and the standard normal random
%variable $Z$ is upper bounded as follows:
%{\small
%\begin{align}
%d_W(W,Z) \leq & \min_{k\leq\nt}\left(\cone B^3\cn\left(\frac{\sltk}{\sigma\sub{\nt}}\right)^2+
%\ctwo B\cn\alpha(k) +\right.\label{eq-stein-generalA}\\
%& \left.B^2\sqrt{\cthree \left(\frac{\cn \tau_k}{\sigma\sub{\nt}}\right)^2+\cfour
%\cn\left(\frac{\sltk}{\sigma\sub{\nt}}\right)^3+\cfive
%\cn\left(\frac{\sltk}{\cn}\right)^2 \frac{\tau_k}{\sigma\sub{\nt}}}\right)\nonumber.
%\end{align}
%}
%where $\cone,\cdots$ and $\cfive$ are constants. $|N_{\leq k}|$ is defined as $\max\limits_{i}\bigcup\limits_{m\leq k}|\{j:\dist(i,j)=m\}|$.
%\end{lem}
\begin{proof}
We define the following sets:
{\small
\begin{align*}
N_m(i):=\{j: \dist(i,j)=m\}\quad\text{,}\quad \nltk(i)&:=\bigcup_{m\leq k}\natm(i)\quad\text{,}\quad\ngtk(i):=\bigcup_{m> k}\natm(i).
\end{align*}
}
We also define the following upper bounds on the sizes of these sets:
{\small
\begin{align*}
|N_m|:=\max_{i}|N_m(i)|\quad\text{,}\quad\sltk:=\max_{i}|\nltk(i)|\quad\text{,}\quad
\sgtk&:=\max_{i}|\ngtk(i)|.
\end{align*}
}
%Let $W_i$ denote the sum of $X_j$'s which are in the set $\ngtk$.
Before beginning, we recall two facts.

(1) {\em Bounded covariance via strong mixing:} For two random
variables $X$ and $Y$ that are more than distance $k$ away, we have
$$|E[XY]-E[X]E[Y]|\leq 4\|X\|_\infty\|Y\|_\infty\alpha(k).$$

(2) {\em Bounds on Wasserstein distance:}
For the set of functions $\mathcal{F} = \{f\mid \|f\|, \|f''\|\leq 2,
\|f'\|\leq \sqrt{2/\pi}\}$,
$$d_W(W,Z) \leq \sup_{f\in\mathcal{F}} |E[f'(W) -
Wf(W)]|,$$
where $d_W(.)$ is the Wasserstein distance and $Z$ has the standard
normal distribution.

In the following, we shall bound $|E[f'(W)-Wf(W)]|$.
We shall repeatedly break up $W$ into two parts: $W_i =
\sum\limits_{j\in\ngtk(i)}X_j$ being the contribution from all nodes within a
distance $k$ of some node $i$, and the remainder from nodes ``far away''
from $i$. Here, $k$ is a parameter that shall be picked later. We can bound $|E[f'(W)-Wf(W)]|$ as follows:

{\small
\begin{align}
&|E[f'(W)-Wf(W)]|=|E[f'(W)-\sum\limits_i X_if(W)]|\label{eq-stein-boundA}\\
&\leq \left|E[f'(W)(1+\sum\limits_i X_i(W_i-W))]\right|
+\left|E[\sum\limits_i X_i(W_i-W)f'(W)+\sum\limits_i X_i f(W)]\right| \nonumber.
%&\leq E\left|f'(W)\left(1+\frac{\sum\limits_i X_i(W_i-W)}{\sigma\sub{\nt}}\right)\right|+\frac{1}{\sigma\sub{\nt}}E\left|\sum\limits_i X_i(W_i-W)f'(W)-\sum\limits_i X_i
\end{align}
}
The second part in eq.~\ref{eq-stein-boundA} can be further bounded above as follows,
{\small
\begin{align}
&\left|E[\sum\limits_i X_i(W_i-W)f'(W)+\sum\limits_i X_i f(W)]\right|\label{eq-term2A}\\
&\leq E\left|\sum\limits_i X_i(W_i-W)f'(W)-\sum\limits_i X_i (f(W_i)-f(W))\right|+\left|E[\sum\limits_iX_if(W_i)]\right|\nonumber\\
%&\leq E\left|\sum\limits_i X_i(W_i-W)(f'(W)-f'(W_i))\right|+\frac{1}{2}E\left|\sum\limits_i X_i(W-W_i)^2f''(W_i^*)\right|+\left|E[\sum\limits_iX_if(W_i)]\right|\nonumber\\
&\leq \frac{1}{2}E\left|\sum\limits_i X_i(W-W_i)^2f''(W_i^*)\right|+\left|E[\sum\limits_iX_if(W_i)]\right|\nonumber\\
&\leq \frac{\|f''\|}{2}E\left|\sum\limits_i
X_i(W_i-W)^2\right|+\left|E[\sum\limits_iX_if(W_i)]\right|,\nonumber
%&\leq \frac{1}{\sigma\sub{\nt}}E\left|\sum\limits_i X_i(W_i-W)(f'(W)-f'(W_i))\right|+E\left|\right|\frac{1}{\sigma\sub{\nt}}\left|\sum\limits_iX_if(W_i)\right|\\
\end{align}
}
where the second inequality follows from Taylor expansion with $W_i^*$
being some value between $W$ and $W_i$.

First, note that:
{\small
\begin{align*}
\|f''\|E\left|\sum\limits_i X_i(W_i-W)^2\right|&=\|f''\|E\left|\sum\limits_i\sum\limits_{j1,j2\in \nltk(i)} X_iX_{j1}X_{j2}\right|\\
&\leq \|f''\|\sum\limits_i\sum\limits_{j1,j2\in \nltk(i)} E|X_iX_{j1}X_{j2}|\\
&\leq \|f''\|\sum\limits_i\sum\limits_{j1,j2\in \nltk(i)} \frac{E|X_i^3|+E|X^3_{j1}|+E|X_{j2}^3|}{3}\\
&\leq 2\cone B^3\frac{\nt \sltk^2}{\sigma\sub{\nt}^3}\quad\mbox{(The factor
$2$ is added for later ease of notation). %\textcolor{red}{Made it $\sltk^2$}
}
\end{align*}
}
As for the second term in eq.~\ref{eq-term2A} we have:
{\small
\begin{align*}
\left|E[\sum\limits_iX_if(W_i)]\right|&\leq \sum\limits_i
\abs{E[X_if(W_i)-E[X_i]E[f(W_i)]]} \quad\mbox{(because $E[X_i]=0$)}\\
&=\sum\limits_i|\cov(X_i,f(W_i))|\leq \frac{4\|f\|B\nt\alpha(k)}{\sigma\sub{\nt}} = \ctwo B\frac{\nt\alpha(k)}{\sigma\sub{\nt}}.
\end{align*}
}
Thus, we obtain a bound for both terms in eq.~\ref{eq-term2A}, and
hence a bound for the second term of eq.~\ref{eq-stein-boundA}.
We will now bound the first term in eq.~\ref{eq-stein-boundA}.
Let $P_\nt=\sum\limits_i X_i(W_i-W)$.  Denote by $\tau_k$ the tail sum
$\sum\limits_{m>k}\satm\alpha(m)$.  Recall that $E[X_i]=0$ and $E[W^2]=1$. Thus,
{\small
\begin{align*}
%& E\left|\sum\limits_i X_i(W-W_i)^2\right| =
\left|E[f'(W)(1+\sum\limits_i X_i(W_i-W))]\right| & \leq
E\left|f'(W)\left(1+P_\nt\right)\right|\leq \|f'\|\sqrt{E\left[1+P_\nt\right]^2}\nonumber\\
&\leq \|f'\|\sqrt{E[(1+E[P_\nt])+(P_\nt-E[P_\nt])]^2} \nonumber\\
&\leq \sqrt{2/\pi}
\sqrt{(1+E[P_\nt])^2+\mbox{var}(P_\nt)}.
\end{align*}
}
Now,
{\small
\begin{equation*}
\begin{array}{l}
\left|E[P_\nt]+1\right|=\left|E\left[\sum\limits_iX_iW_i\right]\right|=\left|\sum\limits_i E [X_i\sum\limits_{j\in \ngtk(i)}X_j]\right|\\
=\left|\sum\limits_i\sum\limits_{m>k}\sum\limits_{j\in\natm(i)}E[X_iX_j]\right|%=\left|E\left[\sum\limits_iX_iW_i\right]-\sum\limits_iE[X_i]E[W_i]\right|\\
=\left|\sum\limits_i\sum\limits_{m>k}\sum\limits_{j\in\natm(i)}(E[X_iX_j]-E[X_i]E[X_j])\right|\\
\leq \sum\limits_i\sum\limits_{m>k}\dfrac{c''B^2}{\sigma\sub{\nt}^2}\alpha(m) \satm \leq
c''B^2\dfrac{\nt\tau_k}{\sigma\sub{\nt}^2}.
\end{array}
\end{equation*}
}

%\txtred{Changed this:
%merged terms (A) and (B). This is because (B) had $i\neq j$ while our
%sets \nfatm didn't handle this. Specifically, the expansion of (B1)
%didn't enforce $i\neq j$.}
Next, we look at the $\mbox{var}(P_\nt)$ term:
{\small
\begin{align}
\mbox{var}(P_\nt)&=(E[P_\nt^2]-E[P_\nt]^2)\label{eq-var-PnA}\\
&=E\left[\left(\sum\limits_{\substack{i\\j\in \nltk(i)}}X_i X_j\right)^2\right]-E[P_\nt]^2
=\underbrace{E\left[\sum\limits_{\substack{i, j\\s\in \nltk(i)\\t\in
\nltk(j)}}X_i X_jX_sX_t\right]}_{(A)}-E[P_\nt]^2\nonumber.
\end{align}
}
The first term (i.e., term (A)) in eq.~\ref{eq-var-PnA} can be broken into two
parts, one such that the minimum distance between any node in
$\{i,s\}$ and any node in pair $\{j,t\}$ is $\leq k$ (denote this by
set $F_{\leq k}$), and one where its greater than $k$ (denote this by
set $F_{> k}$). Formally, we define the following terms:
{\small
\begin{align*}
\nfatm &= \{(i,j,s,t): s\in \nltk(i), t\in \nltk(j), \min_{a,b\in\{i,j,s,t\}} \dist(a,b)=m\}\\
\nfltk &= \bigcup_{m\leq k}\nfatm \quad\text{,}\quad\nfgtk = \bigcup_{m> k}\nfatm \quad\text{,}\quad\sfatm = \max_{i}|\nfatm(i)|\quad \text{,}\quad \sfltk = \max_{i}|\nfltk(i)|.
\end{align*}
}
%\txtred{I've changed this. NEEDS TO BE CHECKED, and if correct, formulas need to be updated.}
Consider the term $\sfltk$.
Given $i$, $s$ can be picked in at most \sltk ways. Now, either $j$ or
$t$ or both must be within distance $k$ of $i$ or $s$. Thus, given $i$
and $s$, $j$ (or $t$) can be picked in at most $2\sltk$ ways, and then
$t$ (or $j$) can be picked in another \sltk ways. Hence, $\sfltk \leq
4\nt \sltk^3$. By a similar argument, $\sfatm \leq 4\nt\sltk^2\satm.$

Now, we have:
{\small
\begin{align*}
(A)&=\sum\limits_{\nfltk}E[X_iX_jX_sX_t] + \sum\limits_{\nfgtk}E[X_iX_jX_sX_t]\nonumber\\
&= \sum\limits_{\nfltk}E[X_iX_jX_sX_t] + \sum\limits_{\nfgtk} E[X_iX_s]E[X_jX_t] +
\sum\limits_{\nfgtk} \left(E[X_iX_jX_sX_t] - E[X_iX_s][X_jX_t]\right)\nonumber\\
&\leq \underbrace{\sum\limits_{\nfltk}E[X_iX_jX_sX_t]}_{(B0)} +
\underbrace{\sum\limits_{\nfgtk} E[X_iX_s]E[X_jX_t]}_{(B1)} +
\underbrace{4\sum\limits_{m>k}\sum\limits_{\nfatm}\frac{B^4}{\sigma\sub{\nt}^4}\alpha(m)}_{(B2)}\nonumber.\\
%\label{eq-var-Pn-22}\\
(B0)& = \sum\limits_{\nfltk} E[X_iX_jX_sX_t]
\leq \sum\limits_{\nfltk} \frac{E[X_i^4]+E[X_j^4]+E[X_s^4]+E[X_t^4]}{4}%\nonumber\\
\leq \frac{B^4}{\sigma\sub{\nt}^4}\sum\limits_{\nfltk} 1\leq
4B^4\frac{\nt\sltk^3}{\sigma\sub{\nt}^4}.
\end{align*}
%\label{eq-var-Pn-21}\\
\begin{align*}
(B1)&=\sum\limits_{\nfgtk} E[X_iX_s]E[X_jX_t]= \sum\limits_{\nfgtk \bigcup \nfltk} E[X_iX_s]E[X_jX_t]-\sum\limits_{\nfltk} E[X_iX_s]E[X_jX_t]\nonumber\\
&\leq (\sum\limits_i E[X_i(W-W_i)])^2+\sum\limits_{\nfltk}
\frac{E[X_i^4]+E[X_s^4]+E[X_j^4]+E[X_t^4]}{4}\leq (E[P_\nt])^2+4B^4\frac{\nt\sltk^3}{\sigma\sub{\nt}^4}\nonumber.\\
%\label{eq-var-Pn-22-part1}\\
(B2) & \leq \frac{4B^4}{\sigma\sub{\nt}^4}\sum\limits_{m>k}\sfatm\alpha(m)\leq 16 B^4\frac{ \nt\sltk^2}{\sigma\sub{\nt}^4}\sum\limits_{m>k}\satm\alpha(m)
\leq 16 B^4\frac{ \nt\sltk^2}{\sigma\sub{\nt}^4}\tau_k\nonumber.%\label{eq-var-Pn-22-part2}
\end{align*}
}

The last equation simply uses a number of applications of the fact
that the geometric mean is less than the arithmetic mean, and Jensen's
inequality. Plugging these into Equation~\ref{eq-var-PnA}, we have:
{\small
\begin{align*}
\mbox{var}(P_\nt)&= (B0)+(B1)+(B2)-E[P_\nt]^2\\
&\leq 4B^4 \frac{\nt\sltk^{2}}{\sigma\sub{\nt}^4}+4
B^4\frac{\nt\sltk^3}{\sigma\sub{\nt}^4}+16 B^4\frac{ \nt\sltk^2}{\sigma\sub{\nt}^4}\tau_k \\
&\leq 8 B^4\frac{\nt\sltk^3}{\sigma\sub{\nt}^4}+16 B^4\frac{ \nt\sltk^2}{\sigma\sub{\nt}^4}\tau_k.
\end{align*}
}
Combining these steps, and recalling that $\cn=\nt/\sigma\sub{\nt}$, we finally obtain the following form for
Eq.~\ref{eq-stein-boundA}:
{\small \begin{align*}
&d_W(W, Z)\nonumber\\
&\leq \cone B^3\frac{\nt \sltk^2}{\sigma\sub{\nt}^3}+ \ctwo B\frac{\nt\alpha(k)}{\sigma\sub{\nt}} +
\|f'\|\sqrt{{c''}^2 B^4\frac{\nt^2\tau_k^2}{\sigma\sub{\nt}^4}+8
B^4\frac{\nt\sltk^3}{\sigma\sub{\nt}^4}+16 B^4\frac{ \nt\sltk^2}{\sigma\sub{\nt}^4}\tau_k}\nonumber\\
&\leq \cone B^3\cn\left(\frac{\sltk}{\sigma\sub{\nt}}\right)^2+ \ctwo B\cn\alpha(k) +
B^2\sqrt{\cthree \left(\frac{\cn \tau_k}{\sigma\sub{\nt}}\right)^2+\cfour
\cn\left(\frac{\sltk}{\sigma\sub{\nt}}\right)^3+\cfive
\cn\left(\frac{\sltk}{\sigma\sub{\nt}}\right)^2 \frac{\tau_k}{\sigma\sub{\nt}}}.
\end{align*}
}
\end{proof}
%\txtred{Made it $\sltk^2$}
\setcounter{section}{10}
\subsection{Statement and proofs of results from section 7}
\setcounter{section}{7}
\setcounter{thm}{1}
We will start by reminding the reader some of the definitions.
Define the following:
{\small
\begin{align*}
\est{\h}(t)&:=\frac{1}{\nn}\sum\limits_{i=1}^\nn \Kb(\dch,\dchq)\dcnpluso\\
\est{\f}(t)&:=\frac{1}{\nn}\sum\limits_{i=1}^\nn \Kb(\dch,\dchq)\dcno\\
q_t&:=\est{\h}(t)-E[\est{\h}(t)|\ec]-\g(\est{\f}(t)-E[\est{\f}(t)|\ec])\\
p_t&:=[\est{\h}(t)-\g\est{\f}(t)]-E[\est{\h}(t)-\g \est{\f}(t)|\et,\ec].
%\Bt&:=E[\est{\h}|\ec]/E[\est{\f}|\ec]-g
\end{align*}}
% Note that $\est{\h}$ equals $\sum_t \est{\h}(t)/(\nt-p)$ and $\est{\f}$ equals $\sum_t \est{\f}(t)/(T-p)$.
We define: $\sigma^2\sub{\nt}(T_1,C):=\var(\sum_t q_t|\et,\ec)$, and $\sigma^2\sub{\nt}(C):=\var(\sum_t q_t|\ec)$.
Also, $\sigma^2\sub{\nt}(T_1,C):=\var(\sum_t q_t|\et,\ec)$.

\begin{lem}\label{lem:normal-conv-part2A}
Under Assumption~1 and assuming $\sigma_c>0$,
%{\small
\begin{align*}
\mbox{Conditioned on \ec,}\qquad \sum_t q_t/\sqrt{T}\stackrel{d}{\rightarrow}\mathcal{N}(0,\sigma^2_c) \qquad\mbox{As $\nt\rightarrow\infty$}.
\end{align*}%}
%\vspace{-1em}
\end{lem}
%\begin{lem}\label{lem:normal-conv-part2A}
%Under assumptions~1 and $\sigma_c>0$,
%$$\mbox{Conditioned on \ec,}\qquad \sum_t q_t/\sqrt{T}\stackrel{d}{\rightarrow}\mathcal{N}(0,\sigma^2_c) \qquad\mbox{As $\nt\rightarrow\infty$}.$$
%\end{lem}

\begin{proof}
Using our distributional convergence results conditioned on $\et\bigcap\ec$, we have shown that
$$\sum_{t\geq T_1+\mc} p_t/\sqrt{T}\stackrel{d}{\rightarrow} \mathcal{N}(0,\sigma_c^2)\qquad\mbox{Conditioned on \et$\bigcap$\ec, when $T_1$ has a finite value}.$$

Denote by $V_t:=\est{\h}(t)-\g\est{\f}(t)$. We have,
{\small
\begin{align}
&|\sum_t q_t/\sqrt{T}-\sum_{t\geq T_1+\mc} p_t/\sqrt{T}|\leq |\sum_{t< T_1+\mc} q_t/\sqrt{T} |+\sum_{t\geq T_1+\mc} \left|E[V_t|\et,\ec]-E[V_t|\ec]\right|/\sqrt{T} \label{eq-bound-expdiffA}\\
& \leq c (T_1+\mc)/\sqrt{T}+c'\sum_{t\geq T_1+\mc} \lambda^{t-T_1}/\sqrt{T}= c'' (T_1+\mc)/\sqrt{T} \quad\mbox{Using Lemma~\ref{lem:diff-expA}.}\nonumber
\end{align}}
where $c$, $c'$ and $c''$ are positive constants.
Let $F_k(x)$ denote the c.d.f of  $\sum_{t\geq T_1+\mc} p_t/\sqrt{T}$, i.e. $F_k(x)=P(\sum_{t\geq k+\mc} p_t/\sqrt{T}\leq x|\et,\ec,T_1=k)$. Lemma~7.4 tells us that, for finite $k$ and $\forall x\in\mathcal{R}$, $F_k(x)\rightarrow \Phi_{0,\sigma_c^2}(x)$; $\Phi_{0,\sigma_c^2}(x)$ being the c.d.f of a normal distribution with mean zero, and standard deviation $\sigma_c$.
Now, using Equation~\ref{eq-bound-expdiffA} we have the following simple argument:
{\small
\begin{align*}
&P(\sum_t q_t/\sqrt{T}\leq x|\ec)\leq \sum_k P(\sum_{t\geq k+\mc} p_t/\sqrt{T}\leq x+c''(k+\mc)/\sqrt{T}|\et,\ec,T_1=k)P(T_1=k|\ec)\\
&\leq \sum_{k\leq K} F_k(x+c''(k+\mc)/\sqrt{T})P(T_1=k|\ec)+P(T_1> K)\qquad\mbox{For any finite $K$}\\
&\rightarrow \limsup\limits_{T\rightarrow\infty} P(\sum_t q_t/\sqrt{T}\leq x|\ec)\leq\Phi_{0,\sigma_c^2}(x)P(T_1\leq K)+P(T_1> K).
\end{align*}
}
In the last step, the exchange of limit and expectation is valid by virtue of the Dominated Convergence Theorem.
Now taking $K\rightarrow\infty$ (which minimizes the upper bound on the $\limsup$) and using the geometric bound on tail probability of $T_1$ in finite state space Markov chains, we have:
$$\limsup\limits_{T\rightarrow\infty} P(\sum_t q_t/\sqrt{T}\leq x|\ec)\leq\Phi_{0,\sigma_c^2}(x).$$

An identical argument on $P(\sum_t q_t/\sqrt{T}> x|\ec)$ gives the following equation.
$$\liminf\limits_{T\rightarrow\infty} P(\sum_t q_t/\sqrt{T}\leq x|\ec)\geq\Phi_{0,\sigma_c^2}(x).$$

Thus we show that
$\mbox{$\forall x\in \mathcal{R}$, as $T\rightarrow \infty$}\qquad P(\sum_t q_t/\sqrt{T}\leq x|\ec)\rightarrow \Phi_{0,\sigma_c^2}(x)$,
which in turn proves our result.
%Using argument similar to Lemma~\ref{lem:cond-expA} (in particular eq~\ref{eq-dev-from-sta}) we have
\end{proof}

%\begin{lem}\label{lem:normal-conv-part1A}
%Define $p_t:=[\est{\h}(t)-\g\est{\f}(t)]-E[\est{\h}(t)-\g \est{\f}(t)|\et,\ec]$.
%Under assumptions 1 and $\sigma_c>0$, for any finite $T_1$, we have:
%%$$\sqrt{T}(\est{\g}-\g-\Bt)\stackrel{d}{\rightarrow} \mathcal{N}(0,\sigma_c^2/R_c^2)\qquad\mbox{conditioned on $\ec\bigcap\et$ as $T\rightarrow\infty$}$$
%$$\sum_{t\geq T_1+\mc} p_t/\sqrt{T}\stackrel{d}{\rightarrow} \mathcal{N}(0,\sigma_c^2)\qquad\mbox{conditioned on $\et\bigcap\ec$ as $T\rightarrow\infty$}.$$
%\end{lem}

\begin{lem}\label{lem:normal-conv-part1A}
Define $p_t:=[\est{\h}(t)-\g\est{\f}(t)]-E[\est{\h}(t)-\g \est{\f}(t)|\et,\ec]$.
Under Assumption~1 and assuming $\sigma_c>0$, for any
finite $T_1$, we have:
%$$\sqrt{T}(\est{\g}-\g-\Bt)\stackrel{d}{\rightarrow} \mathcal{N}(0,\sigma_c^2/R_c^2)\qquad\mbox{conditioned on $\ec\bigcap\et$ as $T\rightarrow\infty$}$$
%{\small
\begin{align*}
\sum_{t\geq T_1+\mc} p_t/\sqrt{T}\stackrel{d}{\rightarrow} \mathcal{N}(0,\sigma_c^2)\qquad\mbox{conditioned on $\et\bigcap\ec$ as $T\rightarrow\infty$}.
\end{align*}%}
\end{lem}

\begin{proof}
We will prove the above result in two parts.
If we can show that the conditions in Lemma~6.3 are satisfied for $$W_\nt := \left.\left(\sum\limits_{t\geq T_1+\mc}p_t\right)\right/\sqrt{\var(\sum\limits_{t\geq T_1+\mc}p_t|\et,\ec)},$$
 then using Lemma~6.1 we will have:
$$W_\nt\stackrel{d}{\rightarrow} \mathcal{N}(0,1)\qquad\mbox{conditioned on $\et\bigcap\ec$}.$$
However, for any finite value of $T_1$, $\var(\sum_{t\geq T_1+\mc}p_t|\et,\ec)/T\rightarrow \sigma^2_c$ (see Lemma~\ref{lem:cond-varA}, eq.~\ref{eq-var-endpartA}).
Thus, with the additional assumption of $\sigma_c>0$, the result is proved.

Now we will show that, conditioned on \et$\bigcap$\ec, the conditions in Lemma~6.3 are satisfied for $W_\nt$, and thus the Wasserstein distance in Lemma~\ref{lem-stein-mainA} can be upper bounded by $O(T^{-1/2}\log^2 (\nt))$.

%\paragraph{Proposition~\ref{prop-normal-conv} (a):}
First note that $p_t$ is bounded and $E[p_t|\et,\ec]=0$. Thus $p_t$ corresponds to $Y_t$ in Lemma~\ref{lem-stein-mainA}.
Since $p_t$ is a function of $S_t$, it involves $p+1$ graphs ($G_{t-p+1},\dots,G_{t+1}$).
The distance $\dist(i,j)$ is defined as $\max(|i-j|-(p+1),0)$. Thus $\satm$ equals $2$ for $m> 0$, and $2(p+1)$ otherwise; hence $\satm=O(1)$.
Also, $|N_{\leq k}|=O(k)$.
Denote by $\sigma\sub{\nt}(T_1,C)$ the standard deviation of $\sum_{t\geq T_1} p_t$ conditioned on $\ec\bigcap\et$.
%If the conditions of Lemma~6.3 are satisfied then we have
%$\sum_{t} p_t/\sigma\sub{\nt}(T_1,C)\rightarrow \mathcal{N}(0,1)$. From Lemma~\ref{lem:qm-conv} we know that $\sigma\sub{\nt}(T_1,C)/\sqrt{\nt}\rightarrow \sigma_c$, for $\sigma_c\geq 0$. With the additional assumption of $\sigma_c>0$ we have $\left.\sum_{t} q_t\right/\sqrt{T}\sigma_c\rightarrow \mathcal{N}(0,1)$.
Let us now examine the conditions in Lemma~6.3.
\paragraph{Condition 1} \fbox{$\cn\rightarrow \infty$}
%\begin{myquote}
We have $\cn=\nt/\sigma\sub{\nt}(T_1,C)\rightarrow \sqrt{\nt}/\sigma_c\rightarrow\infty$, where the limits follow from Lemma~5.7 and the $\sigma_c>0$ assumption.
%$\sigma\sub{\nt}/\sqrt{\nt}$ converges to some non-negative constant. Hence $\cn=\sqrt{\nt}/(\sigma\sub{\nt}/\sqrt{\nt})$ clearly converges to infinity.
%\end{myquote}

\paragraph{Condition 2} Let $k(\nt)=\log \nt/\beta$. We will show that this satisfies conditions 2a, 2b, and 2c.

    \textbf{2a: } \fbox{$\cn \alpha(k(\nt))\rightarrow 0$}
%{\addtolength{\leftskip}{5mm}
\begin{myquote}
Plugging in the value of $k(\nt)$, and using Lemma~\ref{lem:cond-varA} we see that:
{\small
\begin{align*}
\cn\alpha(k(\nt))=O\left(\left.\nt e^{-\beta k(\nt)}\right/\sigma\sub{\nt}(T_1,C)\right)=
%O\left(\left.\sqrt{\nt \nt^{-(1/2+\delta)}\right/\sigma\sub{\nt}(T_1,C)\right)=
%O\left(\frac{\nt}{\sigma\sub{\nt}(T_1,C)}\nt^{-(1/2+\delta)}\right)
O(T^{-1/2}).%=o(1)
\end{align*}
}
%        $(\left.\sqrt{T}\right/\sigma\sub{\nt}(T_1,C))$ converges to a constant. Hence we only need to show that $\sqrt{T}\alpha(k(\nt))$ converges to zero.
%    \begin{align*}
%    \sqrt{T}\alpha(k(\nt))\leq \sqrt{T}\const\exp(-\beta k(\nt))\leq \const T^{1/2-c\beta}
%    \end{align*}
% If $c=(1/2+\delta)/\beta$ for some positive constant $\delta$, we have $\sqrt{T}\alpha(k(\nt))\leq \const T^{-\delta}$. Hence $\sqrt{T}\alpha(k(\nt))$ converges to zero, and so does $\cn \alpha(k(\nt))$.
 \end{myquote}
%}

    \textbf{2b: } \fbox{$\left.\cn\tau\sub{k(\nt)}\right/\sigma\sub{\nt}(T_1,C)\rightarrow 0$}

\begin{myquote}

    Using Lemma~\ref{lem:cond-varA} we see that:
    {\small
    \begin{align*}
    &\left.\cn \tau\sub{k(\nt)}\right/\sigma\sub{\nt}(T_1,C)=(\nt/\sigma\sub{\nt}(T_1,C)^2)\tau\sub{k(\nt)}=O(\tau\sub{k(\nt)})=O(\sum\limits_{m>k(\nt)}\satm\alpha(m))\\
    &=O(e^{-\beta k(\nt)}\sum_{t>0}e^{-\beta t}) \quad\mbox{Using $\satm=O(1)$ and $\alpha(k)=O(e^{-\beta k})$.}\\
    &=O(e^{-\beta k(\nt)})=O(\nt^{-1}) \quad\mbox{Using  $k(\nt)=\log \nt/\beta$.}
    \end{align*}
    }
    %
%    Since $\nt/\sigma\sub{\nt}(T_1,C)^2$ converges to a constant, we only need to show that $\tau\sub{k(\nt)}\rightarrow 0$.
%
%    Note that $\left.\cn \tau\sub{k(\nt)}\right/\sigma\sub{\nt}(T_1,C)=(\nt/\sigma\sub{\nt}^2)\tau\sub{k(\nt)}$. Since $\nt/\sigma\sub{\nt}^2$ converges to a constant, we only need to show that $\tau\sub{k(\nt)}\rightarrow 0$. Recall that $\tau\sub{k(\nt)}=\sum\limits_{m>k(\nt)}\satm\alpha(m)$. First note that $\tau\sub{k(\nt)}$ is a sum of positive elements, and for all $m$, we have $\satm\leq 2\nn^2=\mbox{constant}$. Using strong mixing conditions and that $n$ is a fixed number we have,
%  \begin{align*}
%  \tau\sub{k(\nt)} &\leq \const \exp(-\beta k(\nt))\sum_{t>0}\exp(-\beta t)\\
%  &\leq c'\exp(-\beta k(\nt)) \qquad\mbox{$c'$ is a positive constant.}\\
%  &\leq c' \nt ^{-c\beta}\qquad\qquad\qquad\mbox{Plugging in $k(\nt)=c\log \nt$.}
%  \end{align*}
%Since $c,\beta$ are positive constants, $\tau\sub{k(\nt)}$ converges to zero.
 \end{myquote}

    \textbf{2c: } \fbox{$\cn \left(\frac{|N_{\leq k(\nt)}|}{\sigma\sub{\nt}(T_1,C)}\right)^2\rightarrow 0$}
    \begin{myquote}
    Again, using Lemma~\ref{lem:cond-varA} gives us:
    {\small
    \begin{align*}
    &\cn \left(\frac{|N_{\leq k(\nt)}|}{\sigma\sub{\nt}(T_1,C)}\right)^2=T/\sigma\sub{\nt}(T_1,C)^2 \left.|N_{\leq k(\nt)}|^2\right/\sqrt{T}=O(\left.|N_{\leq k(\nt)}|^2\right/\sqrt{T})\\
    &=O(\left.k(\nt)^2\right/\sqrt{T})=O((\log \nt)^2/T^{1/2}) \quad\mbox{Using  $k(\nt)=\log \nt/\beta$.}\\
    %& = o(1)
    \end{align*}
    }
%    Since $T/\sigma\sub{\nt}(T_1,C)^2$ converges to a constant, we need to only show that $\left.|N_{\leq k(\nt)}|^2\right/\sqrt{T}$ converges to zero. Note that $|N_{\leq k(\nt)}|\leq 2\nn^2 k(\nt)$. We now plug in $k(\nt)=c\log \nt$, and note that $n$ is a constant. This leads to
%    $\frac{|N_{\leq k(\nt)}|^2}{\sqrt{\nt}}\leq c'' (\log \nt/ T^{1/4})^2$, for some constant $c''$. Since $\log T/T^{1/4}$ converges to zero as $T\rightarrow\infty$, this condition is also satisfied.
    \end{myquote}

%Using this conditions, we also see that %the Wasserstein distance in Lemma~\ref{lem-stein-mainA} can be upper bounded by
%%$\cn\left(\frac{\sltk}{\sigma\sub{\nt}}\right)^2=O\left(\frac{\log^2\nt}{T^{1/2}}\right)$, $\cn\alpha(k) = O(T^{-1/2})$,
%$\left(\frac{\cn \tau_k}{\sigma\sub{\nt}}\right)^2=O(1/T^2)$, and $\cn\left(\frac{\sltk}{\sigma\sub{\nt}}\right)^3=O\left(\frac{\log^3(\nt)}{\nt}\right)$,
%and finally $\cn\frac{\tau_k}{\sigma\sub{\nt}}\left(\frac{\sltk}{\cn}\right)^2=O\left(\frac{\log(\nt)}{\nt}\right)^2$.
Now the upper bound on Wasserstein distance (Lemma~\ref{lem-stein-mainA}) becomes $O(\log(\nt)^{2}/\nt)$ by using $k=\log(\nt)/\nt$ and the expressions derived before as part of the second condition.
\end{proof}%
%\bibliography{purna}
%\bibliographystyle{plain}
%\end{document}